\def\isarxiv{1} 

\ifdefined\isarxiv
\documentclass[11pt]{article}

\usepackage[numbers]{natbib}

\else
\documentclass[letterpaper]{article} 
\usepackage[submission]{aaai24}  

\fi

\ifdefined\isarxiv

\usepackage{amsmath}
\usepackage{amsthm}
\usepackage{amssymb}
\usepackage{algorithm}
\usepackage{subfig}
\usepackage{algpseudocode}
\usepackage{graphicx}
\usepackage{grffile}
\usepackage{wrapfig,epsfig}
\usepackage{url}
\usepackage{xcolor}
\usepackage{epstopdf}

\usepackage{bbm}
\usepackage{dsfont}

\else 

\usepackage{amsmath}
\usepackage{amsthm}
\usepackage{amssymb}
\usepackage{xcolor}
\usepackage{algpseudocode}

\usepackage{times}  
\usepackage{helvet}  
\usepackage{courier}  
\usepackage[hyphens]{url}  
\usepackage{graphicx} 
\usepackage{natbib}  
\usepackage{caption} 
\usepackage{algorithm}

\usepackage{newfloat}
\usepackage{listings}
\DeclareCaptionStyle{ruled}{labelfont=normalfont,labelsep=colon,strut=off} 
\floatstyle{ruled}
\newfloat{listing}{tb}{lst}{}
\floatname{listing}{Listing}
\fi

\allowdisplaybreaks

\ifdefined\isarxiv

\usepackage{tikz}
\usepackage{hyperref}  
\hypersetup{colorlinks=true,citecolor=blue,linkcolor=blue} 
\usetikzlibrary{arrows}
\usepackage[margin=1in]{geometry}

\else

\fi

\newtheorem{theorem}{Theorem}
\newtheorem{lemma}[theorem]{Lemma}
\newtheorem{definition}[theorem]{Definition}

\newtheorem{fact}[theorem]{Fact}
\newtheorem{remark}[theorem]{Remark}

\newcommand{\wh}{\widehat}

\newcommand{\R}{\mathbb{R}}

\renewcommand{\d}{\mathrm{d}}

\renewcommand{\hat}{\wh}

\DeclareMathOperator{\diag}{diag}

\makeatletter
\newcommand*{\RN}[1]{\expandafter\@slowromancap\romannumeral #1@}
\makeatother

\usepackage{lineno}

\ifdefined\isarxiv

\else
\newcommand{\texorpdfstring}[1]{{#1}} 
\fi

\begin{document}

\ifdefined\isarxiv

\date{}

\title{How to Protect Copyright Data in Optimization of Large Language Models?}
\author{
Timothy Chu\thanks{\texttt{hungryTOAN@gmail.com}. Google.}
\and 
Zhao Song\thanks{\texttt{zsong@adobe.com}. Adobe Research.}
\and 
Chiwun Yang\thanks{\texttt{christiannyang37@gmail.com}. 
Sun Yat-sen University.}
}

\else

\title{How to Protect Copyright Data in Optimization of Large Language Models?}
\maketitle 
\fi

\ifdefined\isarxiv
\begin{titlepage}
  \maketitle
\begin{abstract}
Large language models (LLMs) and generative AI have played a transformative role in computer research and applications. Controversy has arisen as to whether these models output copyrighted data, which can occur if the data the models are trained on is copyrighted. LLMs are built on the transformer neural network architecture, which in turn relies on a mathematical computation called Attention that uses the softmax function.

In this paper, we show that large language model training and optimization can be seen as a softmax regression problem. We then establish a method of efficiently performing softmax regression, in a way that prevents the regression function from generating copyright data. This establishes a theoretical method of training large language models in a way that avoids generating copyright data.
\end{abstract}

  \thispagestyle{empty}
\end{titlepage}

{\hypersetup{linkcolor=black}
\tableofcontents
}
\newpage

\else

\begin{abstract}
Large language models (LLMs) and generative AI have played a transformative role in computer research and applications. Controversy has arisen as to whether these models output copyrighted data, which can occur if the data the models are trained on is copyrighted. LLMs are built on the transformer neural network architecture, which in turn relies on a mathematical computation called Attention that uses the softmax function.

In this paper, we show that large language model training and optimization can be seen as a softmax regression problem. We then establish a method of efficiently performing softmax regression, in a way that prevents the regression function from generating copyright data. This establishes a theoretical method of training large language models in a way that avoids generating copyright data.
\end{abstract}

\fi

\section{Introduction}

Large language models have changed the world, with the rise of generative AI models such as ChatGPT, GPT-4, Llama, BERT, BARD, PaLM, and OPT \cite{cha22, bce23, dcl+18, tms+23, tli+23, bar23, cnd+22, adf+23, zrg+22}. These models are able to process natural language effectively, handling a wide range of tasks including story generation, code creation, machine translation, and elementary mathematical problem solving \cite{bmr+20, sdf+20, wsc+16, wws+22}. One core component in the large language model is the \textbf{transformer} architecture \cite{vsp+17}, which is built on a computational step known as \textit{attention}. Transformers have been used in a wide variety of tasks outside of large language models, including generative image systems such as DALL-E \cite{dal21} and DALL-E2 \cite{dal22}. Recent research has integrated the transformer architecture with scalable diffusion-based image generation models \cite{bnx+23, cwr+22, wff+23, hwc+20, dbk+20}. 

Once challenge in generative AI is guaranteeing that outputs are protected from copyright infringement and intellectual property issues \cite{hg15, hri16, sag18, gil19, vkb23}. Generative models trained on large corpuses of data can inadvertently generate outputs that are direct copies, or close variants, of copyrighted text or images that the model is trained on. This has led to controversy in using generative artificial intelligence, and past researchers have considered models and theoretical frameworks for evaluating whether generative models are copying data, and how to evaluate and avoid copyright issues that arise~\cite{vkb23}.

Our paper has two main contributions:
\begin{enumerate}
    \item We provide an approach for solving general regression problems in a way that avoids generating copyright data. We term this approach \textit{copyright regression}.
    \item We show how to protect copyright data in the optimization and training of transformer-based architectures (including most large language models), by solving copyright regression for the softmax function.
\end{enumerate}

Solving the copyright regression problem for the softmax function is the key technical contribution of our paper. To establish the copyright regression framework, we provide a new optimization objective for a general regression problem where some outputs are copyrighted. Such a case can arise in when regression outputs are images or sentences, which occurs in transformer-based architectures for language generation and image generation (where we rely on the fact that transformer training can be viewed as a softmax regression problem~\cite{lsxyz23, dls23}). To solve copyright regression for the softmax function, we show that the objective function of the softmax copyright regression is convex, and that its Hessian is bounded. Showing this convexity is non-trivial, and requires intricate bounding of key matrix and vector quantities that arise in the softmax copyright regression problem. Establishing convexity and the bounded Hessian property of the objective function in softmax copyright regression allows us to use gradient-based methods to efficiently solve this problem, with guaranteed bounds on convergence and good stability properties. We formally define copyright regression in Section~\ref{sec:softmax_reg}, and provide our formal proof guarantees in Section~\ref{sec:guarantees}.

\section{Related Work}

This section briefly reviews the related research work on privacy and security of AI, theoretical large language model work, and optimization of neural networks. These topics have a close connection to our work.

\noindent \textbf{Privacy and Security.} Generative AI has achieved impressive results in various domains, including images, text, and code. However, preventing copyright infringement is a challenge that needs to be addressed \cite{hg15, hri16, sag18, gil19}. \cite{sag18} discusses whether data mining and machine learning on copyrighted text qualify as "fair use" under U.S. law. \cite{gil19} investigates copyright infringement in AI-generated artwork and argues that using copyrighted works during the training phase of AI programs does not result in infringement liability. To mitigate potential harms of large language models, in \cite{kgw+23}, a watermarking framework is introduced that facilitates the embedding of signals within generated text. This framework aims to enhance the detection of output from Language Model (LLM) systems, thereby mitigating potential misuse or abuse. Building upon this foundation, subsequent research \cite{hxl+22, hxq+22} has contributed to the development of more robust and less intrusive watermark embedding algorithms. These advancements seek to improve the stability and minimize any adverse effects associated with the process of embedding watermarks. Such endeavors are important in ensuring the integrity and responsible utilization of LLM technology. \cite{vkb23} proposes a framework that provides stronger protection against sampling protected content, by defining near access-freeness (NAF) and developing generative model learning algorithms. Experiments demonstrate promising results with some impact on output quality for both language and image generative models. Recently, \cite{gsy23_dp} focuses on this issue of sampling protected content, and proposes a provable method for privately computing the attention matrix using differential privacy. \cite{xza+23} trains language models (LMs) with federated learning (FL) and differential privacy (DP) in the Google Keyboard (Gboard).

\noindent \textbf{Theoretical LLM.} Since the explosion of large language models, theoretical research on \textbf{transformer} has been one major component of improving language model performance \cite{kkl20, clp+20, tda+20, nab+22, dls23, pmx+23, ag23, szs+23, sht23, jrl23, as23, bsz23, zhl+23, mgn+23, llh+23, rsm+23, ija+23, gsy23_coin, zpg+23, dlms23, gsyz23_quantum, wyw+23, lwd+23}. \cite{rgg+20} proposes AdapterDrop, a method that removes adapters from lower transformer layers during training and inference to reduce computational overhead, while still maintaining task performance. \cite{tbm+20} shows that random alignment matrices perform competitively and learning attention weights from token-token interactions is not highly significant. So they propose Synthesizer, a model that learns synthetic attention weights without token-token interactions and performs well in various tasks. \cite{cdw+20} proposes Scatterbrain, a way to balance model quality and efficiency in approximating long sequences. Recent work \cite{ag23} explores the emergence of new skills in language models through scaling up their parameters and training data. This demonstrates through mathematical analysis that the Scaling Laws provide a strong inductive bias, enabling efficient learning in pre-trained models. they term this phenomenon "slingshot generalization," as it seems to violate traditional generalization theory.

\noindent \textbf{Optimization and Convergence of Deep Neural Networks.} Prior research \cite{ll18, dzp+18, als19a, als19b, adh+19a, adh+19b, sy19, cgh+19, zmg19, cg19, zg19, os20, jt19, lss+20, hls+21, zpd+20, bps+20,  zks+20, szz21, als+22, mosw22, zha22, gms23, lsz23, qsy23} on the optimization and convergence of deep neural networks has been crucial in understanding their exceptional performance across various tasks. These studies have also contributed to enhancing the safety and efficiency of AI systems. In \cite{gms23} they define a neural function using an exponential activation function and apply the gradient descent algorithm to find optimal weights. In \cite{lsz23}, they focus on the exponential regression problem inspired by the attention mechanism in large language models. They address the non-convex nature of standard exponential regression by considering a regularization version that is convex. They propose an algorithm that leverages input sparsity to achieve efficient computation. The algorithm has a logarithmic number of iterations and requires nearly linear time per iteration, making use of the sparsity of the input matrix. 

\section{Preliminary}\label{sec:short_preli}

In this section, we present the preliminary concepts and definitions that form the foundation of our paper. We begin by introducing the notations we utilize in Section~\ref{subsec:notation}. In Section~\ref{subsec:problem_def} we provide the problem definition that we aim to solve.

\subsection{Notations}\label{subsec:notation}

Now we utilize the following notations and definitions: The $\ell_p$ norm of a vector $x$ is denoted as $\| x \|_p$, for examples, $\| x \|_1 := \sum^n_{i=1} | x_i |$, $\| x \|_2 := ( \sum^n_{i=1} x_i^2 )^{1/2}$ and $\| x \|_\infty := \max_{i \in [n]} | x_i |$. For a vector $x \in \R^n$, $\exp(x) \in \R^n$ denotes a vector where whose i-th entry is $\exp(x_i)$ for all $i \in [n]$. For $n > k$, for any matrix $A \in \R^{n \times k}$, we denote the spectral norm of $A$ by $\| A \|$, i.e., $\| A \| := \sup_{x \in \R^k} \| Ax \|_2 / \| x \|_2$. We denote $\sigma_{\min}(A)$ as the minimum singular value of $A$. For two vectors $x, y \in \R^n$, we denote $\langle x, y \rangle = \sum^n_{i=1}$ for $i \in [n]$. Given two vectors $x, y \in \R^n$, we denote $x \circ y$ as a vector whose i-th entry is $x_i y_i$ for all $i \in [n]$. We use $e_i \in \R^n$ to denote a vector whose i-th entry is $1$ and all the other entries are $0$. Let $x \in \R^n$ be a vector. For a vector $x \in \mathbb{R}^n$, $\diag(x) \in \mathbb{R}^{n \times n}$ is defined as a diagonal matrix with its diagonal entries given by $\diag(x)_{i,i} = x_i$ for $i = 1, ..., n$, and all off-diagonal entries are $0$. A symmetric matrix $A \in \mathbb{R}^{n \times n}$ is said to be positive definite (PD) when $A \succ 0$, for all non-zero vectors $x \in \mathbb{R}^n$, we have $x^\top A x > 0$. Similarly, a symmetric matrix $A \in \mathbb{R}^{n \times n}$ is said to be positive semidefinite (PSD) when $A \succeq 0$, for all vectors $x \in \mathbb{R}^n$, we have $x^\top A x \geq 0$.

\subsection{Problem Definition}\label{subsec:problem_def}

To achieve a successful copyright infringement claim in the United States and many other jurisdictions, the plaintiff must provide evidence that demonstrates two key elements. Firstly, they must establish that the defendant had access to the plaintiff's copyrighted work. Secondly, they must show that there are substantial similarities between the defendant's work and the original elements of the plaintiff's work \cite{uni22}.

While access to high-quality copyrighted data is essential for enhancing the performance of AI models, it also introduces legal risks. Therefore, when considering the safety and legality of AI systems, it is imperative to ensure that the ideal language model can effectively learn from all data without producing output that resembles copyrighted material present in its training set. By adhering to these considerations, we can maintain both the integrity of intellectual property rights and the lawful operation of AI technologies.

For convenience, we denote training dataset $\mathcal{D}$, copyright data $\mathcal{C} \subset \mathcal{D}$ and other data $\mathcal{O} = \mathcal{D} - \mathcal{C}$. Our objective is to ensure a model $f$, satisfies: for any input $x$, given a metric $L$, the model's output $f(x)$ will not exhibit substantial similarity to any copyrighted content present in its training set. We enforce this by defining a strict gap $\tau$ such that the metric $L(f(x), C)$, where $C \in \mathcal{C}$, is greater than or equal to $\tau$ plus the metric $L(f(x), O)$, where $O \in \mathcal{O}$. That is
\begin{align*}
    L(f(x), C) \geq \tau + L(f(x), O).
\end{align*}
The choice of metric $L$ depends on the specific task, such as Cross Entropy for text generation, mean absolute error or mean square error for regression problems, and Kullback-Leibler divergence or image similarity for image generation, etc. 

To ensure compliance with copyright laws, we apply $\tau$ to the average metric $L$ calculated over both $\mathcal{C}$ and $\mathcal{O}$, thus implementing a formal and conservative definition. And we convert dataset $\mathcal{D}$ to a input matrix $A \in \R^{n \times d}$ and a target vector $b \in \R^n$, where $n$ is the size of dataset, $d$ is the dimension of input data. We now provide the definition of problem below.
\begin{definition}[$\tau$-Copyright-Protected]\label{def:probelm:informal}
Given matrix $A \in \R^{n \times d}$ and vector $b \in \R^n$ that $A = \begin{bmatrix}
    A_1 \\
    A_2
\end{bmatrix}$, and $b = \begin{bmatrix}
    b_1 \\
    b_2
\end{bmatrix}$, where $A_1 \in \R^{n_1 \times d}$, $A_2 \in \R^{n_2 \times d}$, $b_1 \in \R^{n_1}$, $b_2 \in \R^{n_2}$ and $n = n_1 + n_2$. $A_1$, $b_1$ are the data has copyright issue and $A_2$, $b_2$ are the data does not have copyright issue. Denote the train objective $L$. Denote $\tau > 0$ a scalar.

If there is a trained model $f_\theta$ with parameter $\theta$ that satisfies
\begin{align*}
    \frac{L(f_\theta(A_1), b_1)}{n_1} \geq \tau + \frac{L(f_\theta(A_2), b_2)}{n_2}
\end{align*}
then we say this model $f_\theta$ is $\tau$-Copyright-Protected.
\end{definition}

\section{Methodology: Copyright Regression}\label{sec:softmax_reg}

A prominent existing approach, as outlined in the work by \cite{vkb23}, introduces an algorithm that involves training an additional generative model, denoted as $p$, using non-copyrighted data. This algorithm employs rejection sampling to effectively manage the probability of the model generating copyrighted data. However, it is important to note that this method does have certain limitations. Specifically, it incurs higher computational costs during the decoding process and necessitates the retraining of an additional model. Now we introduce that our method, a simple modification to the standard training objective of generative language models to ensure that their outputs do not infringe upon copyright laws.

In accordance with the findings presented in \cite{dls23}, our approach involves decomposing the mechanism of \textbf{Attention} \cite{vsp+17}, into a regression problem termed Softmax Regression. This decomposition enables a deeper examination of the learning process underlying attention training. By adopting this method, we gain valuable insights into the intricacies of attention and its associated learning mechanisms.

We propose a modification to the standard training objective of generative language models based on the principles of Softmax Regression. The objective is to train the model to generate desired outputs, denoted as $f(A) = b$. However, in the case of copyrighted data, represented by $A_1 \in \mathbb{R}^{n_1 \times d}$ and $b_1 \in \mathbb{R}^{n_1}$, we aim to prevent the model from learning to generate these specific outputs. To address this concern, we introduce an additional term $L(f(A_1), b_1)^{-1}$ to the training objective to discourage the model from generating outputs matching the copyrighted data. To control the level of this protection, we introduce a scalar coefficient $\gamma_c > 0$. Consequently, the modified training objective becomes $L(f(A), b) + \gamma_c L(f(A_1), b_1)^{-1}$. This modification serves to strike a balance between achieving the desired outputs and avoiding the generation of copyrighted data. The addition of the inverse term in the training objective helps mitigate the model's tendency to generate prohibited outputs, while the coefficient $\gamma_c$ allows for fine-tuning the level of protection. Compare to \cite{vkb23}, our approach does not necessitate training additional models and impact the generation speed of the model during decoding. It offers a simple and practical method that can be plug-and-play applied on all training objectives and algorithms in attention-based models, to prevent the output of model from outputting copyrighted data.

In Section~\ref{subsec:softmax_regression}, we present the definition of Softmax Regression. In Section~\ref{subsec:copyright_regression}, we present the definition of Copyright Regression. In Section~\ref{subsec:regularization}, we present the regularization of parameters for better optimization.

\subsection{Softmax Regression}\label{subsec:softmax_regression}

In \cite{dls23}, Softmax Regression applies a softmax function, denoted as $f$, to the product of the input matrix $A$ and the parameter vector $x$. The training objective is then defined as minimizing the squared Euclidean distance between $f(x)$ and the target vector $b$, represented as $\langle f(x) - b, f(x) - b \rangle$. By optimizing this objective, Softmax Regression aims to gain insights into the learning process of the attention mechanism.

We define Softmax Regression as follows
\begin{definition}[Softmax Regression in \cite{dls23}]\label{def:f:informal}
Given a matrix $A \in \R^{n \times d}$, we define
\begin{align*}
    f(x) := \langle \exp(A x) , {\bf 1}_n \rangle^{-1} \exp(A x)
\end{align*}
\end{definition}

For the convenience of calculation, we define a intermediate operator $c(x)$ as follows
\begin{definition}\label{def:c:informal}
Given a matrix $A \in \R^{n \times d}$ and a vector $b \in \R^n$, let $f(x)$ be defined as Definition~\ref{def:f:informal}, we define
\begin{align*}
    c(x) := f(x) - b
\end{align*}
\end{definition}

We define the training objective of Softmax Regression as follows
\begin{definition}[Training Objective of Softmax Regression in \cite{dls23}]\label{def:ell:informal}
Given matrix $A \in \R^{n \times d}$ and vector $b \in \R^n$, let $c(x)$ be defined as Definition~\ref{def:c:informal}, we define
\begin{align*}
    \ell(x) =  \langle c(x), c(x) \rangle
\end{align*}
\end{definition}

\subsection{Copyright Regression}\label{subsec:copyright_regression}

Given a matrix $A \in \R^{n \times d}$ and a vector $b \in \R^n$ that $A = \begin{bmatrix}
    A_1 \\
    A_2
\end{bmatrix}$, and $b = \begin{bmatrix}
    b_1 \\
    b_2
\end{bmatrix}$, where $A_1 \in \R^{n_1 \times d}$, $A_2 \in \R^{n_2 \times d}$, $b_1 \in \R^{n_1}$, $b_2 \in \R^{n_2}$ and $n = n_1 + n_2$. $A_1$, $b_1$ are the data has copyright issue and $A_2$, $b_2$ are the data does not have copyright issue. Now to distinguish between train objective of $A_1$, $b_1$ and $A_2$, $b_2$, we follow what we did in Section~\ref{subsec:softmax_regression}. We first provide the definition of Softmax Regression function on Copyright Data as follows
\begin{definition}[Softmax Regression function on Copyrighted Data]\label{def:f_1:informal}
    Given all data matrix $A \in \R^{n \times d}$ and copyrighted data matrix $A_1 \in \R^{n_1 \times d}$, we define
    \begin{align*}
        f_1(x) := \langle \exp(A_{i, *} x) , {\bf 1}_n \rangle^{-1} \exp(A x)
    \end{align*}
    where $i \in [1, n_1]$ denote a integer.
\end{definition}

Also, we provide the definition of intermediate operator $c(x)$ as follows
\begin{definition}\label{def:c_1:informal}
    Given all data matrix $A \in \R^{n \times d}$ and copyrighted data matrix $A_1 \in \R^{n_1 \times d}$ and vector $b_1 \in \R^n$, let $f_1(x)$ be defined as Definition~\ref{def:f_1:informal}, we define
    \begin{align*}
        c_1(x) := f_1(x) - b_1
    \end{align*}
\end{definition}

Now we have officially provided our definition of Copyright Regression below, which can prevent language models from infringing copyright with controllable performance damage and without occupying more resources.
\begin{definition}\label{def:L_copyright:informal}

We denote $\ell(x)$ as Definition~\ref{def:ell:informal}. The function $c_1(x)$ is defined as Definition~\ref{def:c_1:informal}, and we denote $\ell_1(x) = \langle c_1(x), c_1(x) \rangle$ and $\ell_2(x) := \ell(x) - \ell_1(x)$. Let $\gamma_c >0$ denote a parameter that control loss related to copyright data.

We consider the following copyright loss
\begin{align*}
    L_{\mathrm{copyright}}(x) := 0.5 \ell_1(x) + \gamma_c \cdot \ell_1(x)^{-1} + 0.5 \ell_2(x)
\end{align*}
\end{definition}

Additionally, by adjusting the value of $\gamma_c$, one can easily control the learning of copyrighted data within the model. This flexibility allows for a more effective and data-sensitive approach to training language models.

\subsection{Regularization}\label{subsec:regularization}

To make sure the stability during training, we add a regularization term on $L_{\mathrm{copright}}(x)$. We define $L_{\mathrm{reg}}$ as follows
\begin{definition}\label{def:L_reg:informal}
    Given a matrix $A \in \R^{n \times d}$. Given a vector $w \in \R^n$, we define $W = \diag(w)$. We define $L_{reg} : \R^d \to \R$ as follows
    \begin{align*}
        L_{\mathrm{reg}} := 0.5 \| W Ax \|_2^2
    \end{align*}
\end{definition}

After adding regularization term, we define our final objective $L$ as follows
\begin{definition}\label{def:L:informal}
    We denote $L_{\mathrm{copyright}}(x)$ as Definition~\ref{def:L_copyright:informal}, let $L_{reg}$ be defined as Definition~\ref{def:L_reg:informal}, then we define
    \begin{align*}
        L := L_{\mathrm{copyright}}(x) + L_{\mathrm{reg}}
    \end{align*}
    Minimizing $L$ is the \textbf{softmax regression on copyrighted data} problem.
\end{definition}

\section{Optimization Properties of Objective Function \texorpdfstring{$L$}{}}\label{sec:optimization_L}

The main contribution of this section involves addressing the convexity of the objective function $L$, which allows for more efficient and reliable optimization of $L$. This achievement not only enables us to optimize the objective more effectively but also validates the feasibility of utilizing Copyright Regression for achieving convergence in LLM (Language Model) training. For instance, we can leverage popular optimization algorithms such as gradient descent, Newton's method, and their variants to solve the optimization problem efficiently (see Section 8 in \cite{dls23}). 

In Section~\ref{subsec:gradient_hessian}, we compute the gradient and hessian of our train objective. In Section~\ref{subsec:psd}, we show our result that the Hessian of our train objective is Positive Definite. In Section~\ref{subsec:lipschitz}, we show our result that the Hessian of our train objective is Lipschitz. Thus, we can say our train objective $L$ is convex.

\subsection{Gradient and Hessian of \texorpdfstring{$L$}{}}\label{subsec:gradient_hessian}

In order to calculate the convergence and optimization of $L$, we first compute the $\nabla L$ and $\nabla^2 L$. We show our result as follows

\begin{lemma}[Gradient of $L$, informal version of Lemma~\ref{lem:grad:formal}]\label{lem:grad:informal}
    Given matrix $A \in \R^{n \times d}$ that $A = \begin{bmatrix}
        A_1 \\
        A_2
    \end{bmatrix}$, where $A_1, A_2 \in \R^{n_2 \times d}$ and $n = n_1 + n_2$. Also, we are given a vector $b \in \R^n$ with $b = \begin{bmatrix}
        b_1 \\
        b_2
    \end{bmatrix}$, where $b_1, b_2 \in \R^{n_2}$. 
    
    We denote $\ell_1(x)$ and $\ell_2(x)$ as Definition~\ref{def:L_copyright:informal}, denote $L$ as Definition~\ref{def:L:informal}, denote $f(x)$ as Definition~\ref{def:f:informal}, denote $c(x)$ as Definition~\ref{def:c:informal}. Give a vector $w \in \R^n$, we define $W = \diag(w)$.

    We have
    \begin{align*}
        \frac{\d L}{\d x}
        = & ~ A_{*,i}^\top ( - f(x) c(x)^\top f(x) + \diag(f(x)) c(x) ) + 2 \gamma_c \ell_1(x)^{-2} \cdot {A_1}_{*,i}^\top ( f_1(x) c_1(x)^\top f_1(x) \\
        & ~ - \diag(f_1(x)) c_1(x) ) + A^\top W^2 Ax
    \end{align*}
    where $i \in [1, n]$ denote a integer.
\end{lemma}

Please see Appendix~\ref{app:hessian_L} for the proof of Lemma~\ref{lem:grad:informal}.

For convenient, we define $B(x)$ and $B_c(x)$ ($B(x)$ function on copyrighted data)
\begin{definition}[Definition 6.1 in \cite{dls23}]\label{def:B:informal}
    Given matrix $A \in \R^{n \times d}$ and vector $b \in \R^n$ that $A = \begin{bmatrix}
        A_1 \\
        A_2
    \end{bmatrix}$, and $b = \begin{bmatrix}
        b_1 \\
        b_2
    \end{bmatrix}$, where $A_1 \in \R^{n_1 \times d}$, $A_2 \in \R^{n_2 \times d}$, $b_1 \in \R^{n_1}$, $b_2 \in \R^{n_2}$ and $n = n_1 + n_2$. $A_1$, $b_1$ are the data has copyright issue and $A_2$, $b_2$ are the data does not have copyright issue. 

    Denote $f(x)$ as Definition~\ref{def:f:informal}, denote $c(x)$ as Definition~\ref{def:c:informal}, denote $f_1(x)$ as Definition~\ref{def:f_1:informal}, denote $c_1(x)$ as Definition~\ref{def:c_1:informal}. We define $B(x)$ as follows
    \begin{align*}
        B(x) 
        = & ~\langle 3f(x) - 2b, f(x)\rangle \cdot f(x) f(x)^\top + \langle f(x) - b, f(x) \rangle \cdot \diag(f(x)) \\
        & ~ + \diag((2f(x) - b) \circ f(x)) + (b \circ f(x)) \cdot f(x)^\top + f(x) \cdot (b \circ f(x))^\top
    \end{align*}
    and then we also define $B_c(x)$ as follows
    \begin{align*}
        B_c(x)
        = & ~\langle 3f_1(x) - 2b_1, f_1(x)\rangle \cdot f_1(x) f_1(x)^\top  + \langle f_1(x) - b_1, f_1(x) \rangle \cdot \diag(f_1(x)) \\
        & ~ + \diag((2f_1(x) - b_1) \circ f_1(x)) + (b_1 \circ f_1(x)) \cdot f_1(x)^\top + f_1(x) \cdot (b_1 \circ f_1(x))^\top
    \end{align*}
\end{definition}

With $B(x)$ and $B_c(x)$, we can abbreviate our compute result of Hessian of $L$ as follows
\begin{lemma}[Hessian of $L$, informal version of Lemma~\ref{lem:hessian:formal}]\label{lem:hessian:informal}
    Given matrix $A \in \R^{n \times d}$ that $A = \begin{bmatrix}
        A_1 \\
        A_2
    \end{bmatrix}$, where $A_1, A_2 \in \R^{n_2 \times d}$ and $n = n_1 + n_2$. Also, we are given a vector $b \in \R^n$ with $b = \begin{bmatrix}
        b_1 \\
        b_2
    \end{bmatrix}$, where $b_1, b_2 \in \R^{n_2}$. 
    
    Denote $\ell_1(x)$ and $\ell_2(x)$ as Definition~\ref{def:L_copyright:informal}, denote $L$ as Definition~\ref{def:L:informal}, denote $f(x)$ as Definition~\ref{def:f:informal}, denote $c(x)$ as Definition~\ref{def:c:informal}, denote $B(x)$ and $B_c(x)$ be defined as Definition~\ref{def:B:informal}. Given a vector $w \in \R^n$, we define $W = \diag(w)$.

    We have
    \begin{align*}
        \frac{\d^2 L}{\d x_i \d x_i}
        = & ~ A_{*,i}^\top B(x) A_{*,i}^\top +  A^\top W^2 A + 2 \gamma_c \ell_1(x)^{-2} (16 \cdot \ell_1(x)^{-1} \cdot ({A_1}_{*,i}^\top ( - f_1(x) c_1(x)^\top f_1(x) \\
        & ~ + \diag(f_1(x)) c_1(x) ))^2 - {A_1}_{*,i}^\top B_1(x) {A_1}_{*,i}^\top)
    \end{align*}
    where $i \in [0, n]$ denote a integer.
    
    And we also have 
    \begin{align*}
        \frac{\d^2 L}{\d x_i \d x_j}
        = & ~ A_{*,i}^\top B(x) A_{*,j}^\top + A^\top W^2 A + 2 \gamma_c \ell_1(x)^{-2} (16 \cdot \ell_1(x)^{-1} \cdot  {A_1}_{*,i}^\top ( - f_1(x) c_1(x)^\top f_1(x) \\ 
        & ~ + \diag(f_1(x)) c_1(x) ) \cdot {A_1}_{*,j}^\top ( - f_1(x) c_1(x)^\top f_1(x) + \diag(f_1(x)) c_1(x) ) - {A_1}_{*,i}^\top B_c(x) {A_1}_{*,j}^\top)
    \end{align*}
    where $i, j \in [1, n]$ denote two integers, $i \neq j$.
\end{lemma}

Please see Appendix~\ref{app:hessian_L} for the proof of Lemma~\ref{lem:hessian:informal}.

\subsection{Hessian of \texorpdfstring{$L$}{} is Positive Definite}\label{subsec:psd}

After computing the Hessian of $L$, we now show our result that can confirm it is positive definite, which implies that $\nabla^2 L \succ 0$. Therefore, we have strong evidence that $L$ satisfies the condition of convexity, which is a desirable property for optimization purposes.

\begin{lemma}[Hessian is positive definite, informal version of Lemma~\ref{lem:psd:informal}]\label{lem:psd:informal}
    Given matrix $A \in \R^{n \times d}$ and vector $b \in \R^n$. Denote $\gamma \in (0, 1)$ a scalar. Given a vector $w$, denote $W = \diag(w) \in \R^{n \times n}$. We define $w_{i, i}^2$ as the i-th diagonal entry of matrix $W^2 \in \R^{n \times n}$. Let $l > 0$ denote a scalar.
    
    If for all $i \in [n]$, $w_i^2 \geq 8 + 200 \gamma_c \gamma^{-3} + l / \sigma_{\min}(A)^2$, we have
    \begin{align*}
        \nabla^2 L \succeq l \cdot I_d
    \end{align*}
\end{lemma}

Please see Appendix~\ref{app:psd} for the proof of Lemma~\ref{lem:psd:informal}.

\subsection{Hessian of \texorpdfstring{$L$}{} is Lipschitz}\label{subsec:lipschitz}

We now show our result that confirm the Hessian of $L$ is Lipschitz, which is a desirable property in optimization. This indicates that the second derivatives of $L$ change smoothly within a defined range. By leveraging this Lipschitz property, we can employ gradient-based methods with guaranteed convergence rates and improved stability. Overall, this finding validates the feasibility of utilizing Copyright Regression for achieving convergence in LLM (Language Model) training.

\begin{lemma}[Informal version of Lemma~\ref{lem:lipschitz:formal}]\label{lem:lipschitz:informal}
    Denote $R \geq 4$ denote a scalar. Given a matrix $A \in \R^{n \times d}$ and a vector $b \in \R^n$, $\| A \| \leq R$, $\| b \|_2 \leq 1$. Given $x, y \in \R^d$ be two vector parameter for Copyright Regression with conditions $\| x \|_2 \leq R$, $\| y \|_2 \leq R$ and $\| A ( x - y ) \|_\infty \leq 0.01$. Let $L$ be defined as Definition~\ref{def:L:informal}, let $\gamma \in (0,1)$, let $\beta \in (0, 0.1)$. Denote $H(x) := \nabla^2 L(x)$.

    Then,
    \begin{align*}
        \| H(x) - H(y) \|
        \leq ( 13344 \gamma_c + 2 ) \gamma^{-4} \beta^{-2} n^{1.5} \exp(40R^2) \| x - y \|_2
    \end{align*}
\end{lemma}

Please see Appendix~\ref{app:lipschitz} for the proof of Lemma~\ref{lem:lipschitz:informal}.

\section{Optimization and Copyright Protection Guarantees}\label{sec:guarantees}

We have already established the convexity of the training objective $L$ in Section~\ref{sec:optimization_L}, providing a strong foundation to confidently pursue the global optimal value of $L$ through optimization techniques. Now we present the main results of this paper: 1) the minimization guarantee of $L$, 2) the copyright protection efficiency of Copyright Regression. 

Firstly, in Section~\ref{subsec:main_result}, our objective is to minimize $L$ to its optimal value, ensuring that we achieve the most favorable outcome in terms of our training process. The minimization guarantee of $L$ confirms our main result on optimization of Copyright Regression, it also demonstrates the ease of use of Copyright Regression, which can be optimized on any attention-based model. At the same time, denote $x^*$ as the optimal solution of training objective $L$, analyzing $L(x^*)$'s performance on copyright data can help us to understand how the trained Copyright Regression can avoid copyright infringement.

Secondly, in Section~\ref{subsec:copyright_protected}, we aim to demonstrate that the optimal $L$ provides robust protection for its outputs, safeguarding them from potential copyright infringement. By delineating this boundary, we can quantitatively assess the extent to which Copyright Regression preserves the integrity and exclusivity of copyrighted content. This analysis will provide valuable insights into the effectiveness of our approach and its ability to strike a balance between data protection and the need for authorized access.

\subsection{Minimizing Loss Guarantee}\label{subsec:main_result}

We provide our minimum training objective theorem below.
\begin{theorem}[Minimizing training objective $L$, informal version of Theorem~\ref{the:main_result:formal}]\label{the:main_result:informal}
    Suppose we have matrix $A \in \R^{n \times d}$ and $A_1 \in \R^{n_1 \times d}$, $n_1 \leq n$, vector $b, w \in \R^n$. Let $L$ be defined as Definition~\ref{def:L:informal}, denote $x^*$ as the optimal solution of $L$ where $g(x^*) = \mathbf{0}_d$ and $\| x^* \| \leq R$. Denote $R \geq 10$ be a positive scalar. Denote $M = n^{1.5}\exp(40R^2)$, Let $x_0$ be denoted as an initial point where $M \| x_0 - x^* \|_2 \leq 0.1l$, where $l > 0$ denoted a scalar.

    For any accuracy $\epsilon \in (0, 0.1)$ and any failure probability $\delta \in (0, 0.1)$, there exists a randomized algorithm, with probability $1 - \delta$, it runs $T = \log(\| x_0 - x^* \|_2 / \epsilon)$ iteration and outputs a vector $\widetilde{x} \in \R^d$ such that
    \begin{align*}
        \| \widetilde{x} - x^* \| \leq \epsilon
    \end{align*}
    and the time cost of each iteration is
    \begin{align*}
        O( ( \mathrm{nnz}(A) + d^w ) \cdot \mathrm{poly}( \log( n / \delta ) ) )
    \end{align*}
    Here $w$ is the exponent of matrix multiplication. Currently $w \approx 2.373$.
\end{theorem}

Please see Appendix~\ref{app:main_result} for the proof of Theorem~\ref{the:main_result:informal}.

\subsection{\texorpdfstring{$L$}{} is \texorpdfstring{$\tau_c$}{}-Copyright-Protected}\label{subsec:copyright_protected}

Now we provide a boundary that illustrates the efficacy of Copyright Regression in safeguarding copyrighted data, while also addressing the criteria outlined in Definition~\ref{def:probelm:informal}, which serves as our definition of copyright protection in this paper. 

We set $\ell(x)$ in Definition~\ref{def:ell:informal} as a $\ell_2$ metric for measuring parameter $x$ on learning data $A$. Now we present our result to confirm that training using our Copyright Regression method can ensure that the model's outputs do not infringe copyright. Specifically, we can assert that the trained model $L$ is protected against copyright infringement with a threshold of $\tau_c$ based on Theorem~\ref{the:tau_c_protect:informal} below.

\begin{theorem}[Informal version of Theorem~\ref{the:tau_c_protect:formal}]\label{the:tau_c_protect:informal}
    Let $x^*$ be denoted the optimal parameter on Copyright Regression. We define $\ell(x)$ as Definition~\ref{def:ell:informal}, denote $\ell(x)$ as the original train objective of Softmax Regression. Denote $\epsilon_2 \in (0, 0.1)$ a scalar. Denote $\tau_c := \sqrt{2\gamma_c} / n_1 - \epsilon_2 / n_2$, we have
    \begin{align*}
        \frac{\ell_1(x^*)}{n_1} \geq \tau_c + \frac{\ell_2(x^*)}{n_2}
    \end{align*}
    so $x^*$ in Copyright Regression is $\tau_c$-Copyright-Protected.
    
\end{theorem}

Please see Appendix~\ref{app:copyright_protected} for the proof of Theorem~\ref{the:tau_c_protect:informal}.

Now we have provided evidence of the copyright protection achieved through training under the Copyright Regression objective. This method has been rigorously proven and offers complete control over copyright infringement. However, concerns may arise regarding the potential impact of the Copyright Regression approach on the model's overall performance, particularly when copyright data includes high-quality novels and images that contribute significantly to the model's performance. In fact, language models cannot remember all its train data. Its training loss has a considered range instead of equaling to 0. Base on this, we only need to let model's performance on copyrighted data be different from model's performance on other data, even this difference is very small, then we can ascertain whether the model has access to these copyright data during output generation and intentionally avoids outputting them. The difference, namely $\tau$, can be easily control by adjust the value of $\gamma_c$ and $n_1 / n$, we will continue to explain that why we say this in Section~\ref{sec:experiment}.

\begin{figure*}
\centering
\includegraphics[scale=0.48]{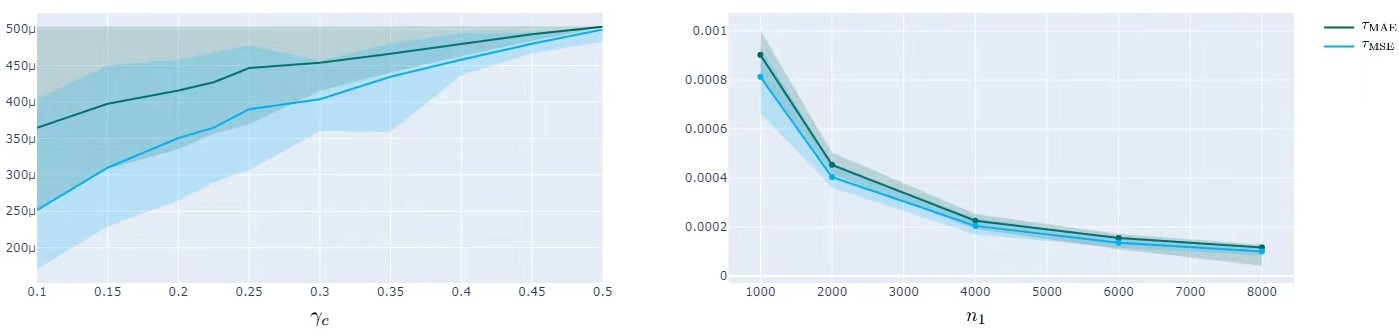}
\caption{Copyright Regression experiment result}
\label{fig:f1}
\end{figure*}

\section{Experiment}\label{sec:experiment}

In order to evaluate and demonstrate the effectiveness of our proposed Copyright Regression approach, we conducted extensive experiments using Softmax Regression. By varying the values of $n_1$ (representing the number of data instances with copyright issues) and $\gamma_c$ (the coefficient used to control the Copyright Regression), we compared the results against a baseline model. The experimental findings clearly indicate the efficacy of our method in providing effective copyright protection.

In Section~\ref{subsec:setup}, we provided the details of our experiment. In Section~\ref{subsec:result_analysis}, we provided experimental results and analyzed the effectiveness of Copyright Regression.

\subsection{Setup}\label{subsec:setup}

\textbf{hyper-parameters.} The hyper-parameters used in each experiment run were set to $n = 10000$ and $d = 512$. To assess the influence of copyright data with different proportions during training, we varied the value of $n_1$ to be $n_1 \in \{1000, 2000, 4000, 6000, 8000\}$. Additionally, to evaluate the impact of different values of $\gamma_c$ on copyright protection, we consider $\gamma_c$ values of $\{0.1, 0.15, 0.2, 0.225, 0.25, 0.3, $ $0.35, 0.4, 0.45, 0.5\}$.

\noindent \textbf{Dataset.} We employed random selection of data from a Gaussian distribution. Specifically, we randomly selected an input matrix $A \in \mathbb{R}^{n \times d}$ from a normal distribution $\mathcal{N}(0, \mathbf{I}_d)$. For the target vector $b \in \mathbb{R}^n$, we let $b = \langle \exp(u) , {\bf I}_n \rangle^{-1} \exp(u)$, where $u \in \mathcal{N}(0, I_n)$.

\noindent \textbf{Metrics.} We use two evaluation metrics, including $\mathrm{MAE}(\hat{y}, y) = \frac{1}{n} \sum_{i = 1}^n (y - \hat{y})$ and $\mathrm{MSE}(\hat{y}, y) = \frac{1}{n} \sum_{i = 1}^n | y - \hat{y} |$. We define $\tau_{\mathrm{MSE}} := \mathrm{MSE}(f(A_1), b_1) - \mathrm{MSE}(f(A_2), b_2)$ and $\tau_{\mathrm{MAE}} := \mathrm{MAE}(f(A_1), b_1) - \mathrm{MAE}(f(A_2), b_2)$ as Definition~\ref{def:probelm:informal}, where $f$ denote a model function, $A_1$, $b_1$ are copyright data and $A_2$, $b_2$ are other data.

\noindent \textbf{Baseline.} To evaluate the effectiveness of our approach, we conduct a comparative analysis against a baseline method referred to as \textbf{Random}. In the \textbf{Random} baseline, a parameter vector $x \in \mathbb{R}^d$ is randomly selected from a normal distribution $\mathcal{N}(0, \mathbf{I}_d)$.

\begin{figure}
    \centering
    \includegraphics[scale=0.6]{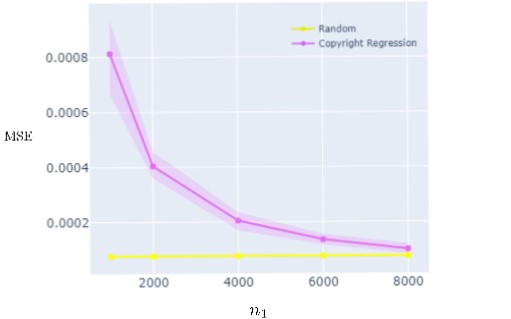}
    \caption{Comparison with \textbf{Random} and Copyright Regression on copyright data}
    \label{fig:f2}
\end{figure}

\subsection{Results and Analysis}\label{subsec:result_analysis}

\textbf{Impact of $\gamma_c$.} The left image of Figure~\ref{fig:f1} depicts the relationship between the variables $\gamma_c$ and the difference metrics $\tau_{\mathrm{MSE}}$ and $\tau_{\mathrm{MAE}}$. In this experiment, we set the value of $n_1 = 2000$. Remarkably, the observed trend aligns closely with the result we derived in Section~\ref{subsec:copyright_protected}. Our derived result, stated as $\tau_{\mathrm{MSE}} = \frac{\ell_1(x)}{n_1} - \frac{\ell_2(x)}{n_2} \geq \frac{\sqrt{2 \gamma_c}}{n_1} - \frac{\epsilon_2}{n_2}$, affirms that our Copyright Regression approach effectively encourages the model to avoid copyright infringement while still maintaining a controllable level of performance degradation.

\noindent \textbf{Impact of the proportion of copyright data.} $n_1$ impacts on model performance is illustrated in the right image of Figure~\ref{fig:f1}. This image showcases the relationship between $n_1$ and the difference metrics $\tau_{\mathrm{MSE}}$ and $\tau_{\mathrm{MAE}}$. Notably, the findings indicate that as the ratio $n_1 / n$ increases, the disparity in model performance between copyright and non-copyright data diminishes. This observation provides valuable insight, suggesting that the addition of data with a distribution similar to that of copyright-protected data can enhance the model's ability to effectively capture the characteristics of copyright data while ensuring that the model's output remains free from copyright infringement.

\noindent \textbf{Comparison with baseline.} Figure~\ref{fig:f2} shows comparison between Copyright Regression and baseline \textbf{Random} on copyright data with metric $\mathrm{MSE}$. While $n_1$ increasing and $\mathrm{MAE}(f(A_1), b_1)$ decreasing, our method shows its strong protection on copyright data even when $n_1 = 8000$, $\mathrm{MAE}(f(A_1), b_1)$ still greater than \textbf{Random}'s $\mathrm{MSE}$ on copyright data. This finding provides compelling evidence that our Copyright Regression approach effectively prevents the occurrence of the "infinite monkey" phenomenon, ensuring that the model's outputs consistently avoid copyright infringement. By maintaining a reliable level of performance on copyright data, our method demonstrates its ability to strike a crucial balance between performance and copyright protection.

\section{Conclusion}\label{sec:conclusion}
Our work shows that the training of transformers can be viewed as a softmax regression problem. We provide a notion of copyright regression, which encourages regression functions to avoid outputting copyrighted data. Then, we combine the two to perform copyright regression on the softmax function, which allows us to train transformers in a way that avoids outputting copyright data.

The main idea to solve copyright regression on the softmax function, was to show that the copyright regression problem is convex and that the Hessian is Lipschitz. This guarantees that gradient descent methods will have guaranteed convergence to the optimal solution with good stability properties. We provide experiments showing that our algorithm performs well in preventing copyright issues on data drawn from a Gaussian distribution, one of the fundamental distributions in machine learning and a test bed for many algorithms, where some data is randomly assigned to be copyrighted.

\ifdefined\isarxiv
\else
\bibliography{ref}

\fi

\newpage
\onecolumn
\appendix

\section*{Appendix}

\textbf{Road map.} In Appendix~\ref{app:preli}, we provide the preliminaries used in our proofs. In Appendix~\ref{app:cr_defs}, we reaffirm our definitions of Copyright Regression. In Appendix~\ref{app:grad_hess}, we provide our computation results for several functions. In Appendix~\ref{app:psd}, we prove that $\nabla^2 L \succeq 0$ and thus $L$ is convex. In Appendix~\ref{app:lipschitz}, we provide our result that Hessian of $L$ is Lipschitz. In Appendix~\ref{app:lipschitz_tool}, we provide a set of tools that can assist in the computation of the Lipschitz property. In Appendix~\ref{app:helpful_bounds}, we present a collection of bound that are valuable for facilitating computations in our proofs. In Appendix~\ref{app:main_result}, we provide guarantee of minimizing our final training objective. In Appendix~\ref{app:copyright_protected},  we show our result that Copyright Regression avoids model outputting copyright data. In Appendix~\ref{app:approx_method}, we provide an approximate version of the newton method for convex optimization.

\section{Preliminary}\label{app:preli}

In this appendix, we present the preliminaries utilized in our proofs. In Appendix~\ref{subapp:notations} introduces the notations employed throughout the document. In Appendix~\ref{subapp:algebras} provides fundamental facts regarding exact computation. In Appendix~\ref{subapp:vec_norm} presents useful tools for determining bounds on norms based on vectors. In Appendix~\ref{subapp:mat_norm} provides useful tools for determining bounds on norms related to matrices. In Appendix~\ref{subapp:basic_psd} furnishes basic inequalities for positive semidefinite (psd) matrices. In Appendix~\ref{subapp:calculus} supplies essential lemmas for exact computation.

\subsection{Notations}\label{subapp:notations}

Now we utilize the following notations and definitions: The $\ell_p$ norm of a vector $x$ is denoted as $\| x \|_p$, for examples, $\| x \|_1 := \sum^n_{i=1} | x_i |$, $\| x \|_2 := ( \sum^n_{i=1} x_i^2 )^{1/2}$ and $\| x \|_\infty := \max_{i \in [n]} | x_i |$. For a vector $x \in \R^n$, $\exp(x) \in \R^n$ denotes a vector where whose i-th entry is $\exp(x_i)$ for all $i \in [n]$. For $n > k$, for any matrix $A \in \R^{n \times k}$, we denote the spectral norm of $A$ by $\| A \|$, i.e., $\| A \| := \sup_{x \in \R^k} \| Ax \|_2 / \| x \|_2$. We denote $\sigma_{\min}(A)$ as the minimum singular value of $A$. For two vectors $x, y \in \R^n$, we denote $\langle x, y \rangle = \sum^n_{i=1}$ for $i \in [n]$. Given two vectors $x, y \in \R^n$, we denote $x \circ y$ as a vector whose i-th entry is $x_i y_i$ for all $i \in [n]$. We use $e_i \in \R^n$ to denote a vector whose i-th entry is $1$ and all the other entries are $0$. Let $x \in \R^n$ be a vector. For a vector $x \in \mathbb{R}^n$, $\diag(x) \in \mathbb{R}^{n \times n}$ is defined as a diagonal matrix with its diagonal entries given by $\diag(x)_{i,i} = x_i$ for $i = 1, ..., n$, and all off-diagonal entries are $0$. A symmetric matrix $A \in \mathbb{R}^{n \times n}$ is said to be positive definite (PD) when $A \succ 0$, for all non-zero vectors $x \in \mathbb{R}^n$, we have $x^\top A x > 0$. Similarly, a symmetric matrix $A \in \mathbb{R}^{n \times n}$ is said to be positive semidefinite (PSD) when $A \succeq 0$, for all vectors $x \in \mathbb{R}^n$, we have $x^\top A x \geq 0$.

\subsection{Basic Algebras}\label{subapp:algebras}

\begin{fact}\label{fact:vector}
    For vectors $u, v, w \in \R^n$, we have
    \begin{itemize}
        \item $\langle u, u \rangle - \langle v, v \rangle = ( u - v )^\top ( u + v ) = ( u + v )^\top ( u - v )$
        \item $\langle u, v \rangle = u^\top v = v^\top u$
        \item $u + v = u - w + w - v$
    \end{itemize}
\end{fact}

\begin{fact}\label{fact:abs}
    For scalars $a, b \in \R$, we have
    \begin{itemize}
        \item $| a + b | \leq |a| + |b|$
        \item $|a b| = |a| \cdot |b|$
    \end{itemize}
\end{fact}

\subsection{Basic Vector Norm Bounds}\label{subapp:vec_norm}

\begin{fact}\label{fact:vec_norm_bound}
    For vectors $u, v \in \R^n$, we have
    \begin{itemize}
        \item $\langle u, v \rangle \leq \|u\|_2 \cdot \|v\|_2$ (Cauchy-Schwarz inequality)
        \item $\| u \|_1 \geq \| u \|_2 \geq \| u \|_\infty$
        \item $\| u + v \|_2 \leq \|u\|_2 + \|v\|_2$
        \item $\| u \circ v \|_2 \leq \| u \|_\infty \cdot \| v \|_2 \leq \| u \|_2 \cdot \| v \|_2$
        \item $\| u - v \|_2 \leq \| u \|_2 + \| v \|_2$
        \item $\| \diag(u) \| \leq \| u \|_\infty \leq \| u \|_2$
        \item $\| u \|_\infty \leq \| u \|_2$
        \item $\|u\|_2 = \|u^\top\|_2$
        \item $\|u^\top - v^\top\|_2 = \|u - v\|_2$
        \item Let $\alpha$ denote a scalar, we have $\|\alpha u\|_2 = | \alpha | \cdot \|u\|_2$
        \item Let $u_1, u_2, \dots, u_n, v_1, v_2, \dots, v_n \in \R^n$ denote $n$ vectors, we have
        \begin{align*}
            \| \sum_{i = 1}^n u_i - \sum_{i = 1}^n v_i \|_2 \leq \sum_{i = 1}^n \| u_i - v_i \|_2
        \end{align*}
    \end{itemize}
\end{fact}

\subsection{Basic Matrix Norm Bounds}\label{subapp:mat_norm}

\begin{fact}\label{fact:mat_norm_bound}
    For matrices $U, V, W \in \R^n$, we have
    \begin{itemize}
        \item $\|U + V\| \leq \|U\| + \|V\|$
        \item $\|U + V\| = \|U - W + W + V\|$
        \item Let $\alpha \in \R$ denote a scalar, then we have $\|\alpha U\| \leq |\alpha|\cdot\|U\|$
        \item Let $u, v \in \R^n$ denote two vectors, then we have $\|uv^\top\| \leq \|u\|_2 \cdot \|v\|_2$
    \end{itemize}
\end{fact}

\subsection{Basic Positive Semidefinite}\label{subapp:basic_psd}

\begin{fact}\label{fact:psd}
    Let $u, v \in \R^n$, we have
    \begin{itemize}
        \item $uu^\top \preceq \| u \|_2^2 \cdot I_n$
        \item $\diag(u) \preceq \| u \|_2 \cdot I_n$
        \item $( u \circ v ) \circ u^\top \preceq \| v \|_\infty \cdot u u^\top$
        \item $u \circ ( u \circ v )^\top \preceq \| v \|_\infty \cdot u u^\top$
        \item $u u^\top \succeq 0$
        \item Let $U \in \R^n$ denote a matrix, we have $U \succeq \sigma_{\min}(U)$
    \end{itemize}
\end{fact}

\subsection{Basic Calculus}\label{subapp:calculus}

\begin{lemma}\label{lem:diff_chain}
We have
\begin{itemize}
\item 
\begin{align*}
    \frac{\d^2 f(x)^{-1} }{\d t \d t} =  2 f(x)^{-3} \cdot ( \frac{\d f(x)}{ \d t } )^2 - f(x)^{-2} \cdot \frac{\d^2 f(x)}{\d^2 t}
\end{align*}
\item \begin{align*}
    \frac{\d^2 f(x)^{-1} }{\d t_1 \d t_2} = 2 f(x)^{-3} \cdot \frac{\d f(x)}{ \d t_1 } \cdot \frac{\d f(x)}{ \d t_2 } - f(x)^{-2} \cdot \frac{\d^2 f(x)}{\d t_1 \d t_2} 
\end{align*}
\end{itemize}
\end{lemma}
\begin{proof}
We have
\begin{align*}
& ~ \frac{\d^2 f(x)^{-1} }{\d t \d t} \\
 = & ~ \frac{\d}{\d t} (\frac{\d f(x)^{-1}}{\d t}  ) \\
 = & ~  \frac{\d}{\d t} ( - f(x)^{-2} \frac{\d f(x) }{ \d t} ) \\
 = & ~ 2 f(x)^{-3} \cdot \frac{\d f(x)}{ \d t } \cdot \frac{\d f(x)}{ \d t } - f(x)^{-2} \cdot \frac{\d^2 f(x)}{\d^2 t} \\
 = & ~ 2 f(x)^{-3} \cdot ( \frac{\d f(x)}{ \d t } )^2 - f(x)^{-2} \cdot \frac{\d^2 f(x)}{\d^2 t}
\end{align*}
where the first equality follows from the expansion of hessian, the second, third equalities follow from differential chain rule, the fourth equality follows from simply algebra.

Similarly, we have 
\begin{align*}
 & ~ \frac{\d^2 f(x)^{-1} }{\d t_1 \d t_2 } \\
 = & ~ \frac{\d}{\d t_1} (\frac{\d f(x)^{-1}}{\d t_2}  ) \\
 = & ~  \frac{\d}{\d t_1} ( - f(x)^{-2} \frac{\d f(x) }{ \d t_2} ) \\
 = & ~ 2 f(x)^{-3} \cdot \frac{\d f(x)}{ \d t_1 } \cdot \frac{\d f(x)}{ \d t_2 } - f(x)^{-2} \cdot \frac{\d^2 f(x)}{\d t_1 \d t_2} 
\end{align*}
where the first equality follows from the expansion of hessian, the second, third equalities follow from differential chain rule.
\end{proof}

\begin{lemma}\label{lem:partializ_bound_tool}
    If the given conditions are satisfied
    \begin{itemize}
        \item Let $x, y \in \R^d$
        \item For $u(x), v(x) \in \R^{n}$ 
    \end{itemize}
    We have
    \begin{itemize}
        \item Part 1.
        \begin{align*}
            & ~ | u(x)^\top v(x) - u(y)^\top v(y) | \\
            \leq & ~ \|u(x)\|_2 \cdot \| v(x) - v(y) \|_2 + \| u(x) - u(y) \|_2 \cdot \|v(y)\|_2
        \end{align*}
        \item Part 2.
        \begin{align*}
            & ~ \| u(x) v(x)^\top - u(y) v(y)^\top \| \\
            \leq & ~ \|u(x)\|_2 \cdot \| v(x) - v(y) \|_2 + \| u(x) - u(y) \|_2 \cdot \|v(y)\|_2
        \end{align*}
        \item Part 3. Let $\alpha(x) \in \R$ denote a scalar, then we have
        \begin{align*}
            & ~ \| \alpha(x) u(x) v(x)^\top - \alpha(y) u(y) v(y)^\top \|  \\
            \leq & ~ | \alpha(x) | \cdot \| u(x) \|_2 \cdot \| v(x) - v(y) \|_2 \\
            & ~ + | \alpha(x) | \cdot \|u(x) - u(y)\|_2 \cdot \| v(y) \|_2 \\
            & ~ + \| u(y) \|_2 \cdot | \alpha(x) - \alpha(y) | \cdot \| v(y) \|_2
        \end{align*}
        \item Part 4. Let $\alpha(x), \beta(x) \in \R$ denote two scalars, then we have
        \begin{align*}
            & ~ | \alpha(x)\beta(x) - \alpha(y)\beta(y) |
            \\
            \leq & ~ | \alpha(x) | \cdot | \beta(x) - \beta(y) | + | \alpha(x) - \alpha(y) | \cdot | \beta(y) |
        \end{align*}
        \item Part 5.
        \begin{align*}
            & ~ \| u(x) \circ v(x) - u(y) \circ v(y) \|_2 \\
            \leq & ~ \| u(x) \|_2 \cdot \| v(x) - v(y) \|_2 + \| u(x) - u(y) \|_2 \cdot \| v(y)\|_2
        \end{align*}
    \end{itemize}
\end{lemma}
\begin{proof}

{\bf Proof of Part 1.}
    \begin{align*}
        & ~ | u(x)^\top v(x) - u(y)^\top v(y) | \\
        = & ~ | u(x)^\top v(x) - u(x)^\top v(y) + u(x)^\top v(y) - u(y)^\top v(y) | \\
        = & ~ | u(x)^\top ( v(x) - v(y) ) + ( u(x)^\top - u(y)^\top ) v(y) | \notag \\
        \leq & ~ | u(x)^\top ( v(x) - v(y) ) | + | ( u(x)^\top - u(y)^\top ) v(y) | \\
        \leq & ~ \|u(x)^\top\|_2 \cdot \| v(x) - v(y) \|_2 + | ( u(x)^\top - u(y)^\top ) v(y) | \\
        = & ~ \|u(x)\|_2 \cdot \| v(x) - v(y) \|_2 + | ( u(x)^\top - u(y)^\top ) v(y) | \\
        \leq & ~ \|u(x)\|_2 \cdot \| v(x) - v(y) \|_2 + \| u(x)^\top - u(y)^\top \|_2 \cdot \|v(y)\|_2 \\
        = & ~ \|u(x)\|_2 \cdot \| v(x) - v(y) \|_2 + \| u(x) - u(y) \|_2 \cdot \|v(y)\|_2
    \end{align*}
    where the first equality follows from Fact~\ref{fact:vector}, the second equality follows from simple algebra, the third equality follows from Fact~\ref{fact:abs}, the fourth, fifth, sixth, seventh equalities follow from Fact~\ref{fact:vec_norm_bound}.

{\bf Proof of Part 2.}
    \begin{align*}
        & ~ \| u(x) v(x)^\top - u(y) v(y)^\top \| \\
        = & ~  \| u(x) v(x)^\top - u(x) v(y)^\top + u(x) v(y)^\top - u(y) v(y)^\top\| \\
        \leq & ~ \| u(x) v(x)^\top - u(x) v(y)^\top \| + \| u(x) v(y)^\top - u(y) v(y)^\top\| \\
        = & ~ \| u(x) ( v(x)^\top - v(y)^\top ) \| + \| ( u(x) - u(y) ) v(y)^\top\| \\
        \leq & ~ \| u(x) \|_2 \cdot \| v(x)^\top - v(y)^\top \|_2 + \| ( u(x) - u(y) ) v(y)^\top\| \\
        \leq & ~ \| u(x) \|_2 \cdot \| v(x)^\top - v(y)^\top \|_2 + \| u(x) - u(y) \|_2 \cdot \|v(y)^\top\|_2 \\
        = & ~ \|u(x)\|_2 \cdot \| v(x) - v(y) \|_2 + \| u(x) - u(y) \|_2 \cdot \|v(y)\|_2
    \end{align*}
    where the first, second equalities follow from Fact~\ref{fact:mat_norm_bound}, the third equality follows from simple algebra, the fourth, fifth equalities follow from Fact~\ref{fact:mat_norm_bound}, the sixth equality follows from Fact~\ref{fact:vec_norm_bound}.

{\bf Proof of Part 3.}
    \begin{align*}
        & ~ \| \alpha(x) u(x) v(x)^\top - \alpha(y) u(y) v(y)^\top \| \\
        \leq & ~ \| \alpha(x) u(x) \|_2 \cdot \| v(x) - v(y) \|_2 \\
        & ~ + \| \alpha(x) u(x) - \alpha(y) u(y) \|_2 \cdot \| v(y) \|_2 \\
    \end{align*}
    where the first equality follows from Part 2 of Lemma~\ref{lem:partializ_bound_tool}.
    
    For the first term, we have
    \begin{align*}
        & ~ \| \alpha(x) u(x) \|_2 \cdot \| v(x) - v(y) \|_2 \\
        = & ~ | \alpha(x) | \cdot \| u(x) \|_2 \cdot \| v(x) - v(y) \|_2
    \end{align*}
    where the first equality follows from Fact~\ref{fact:vec_norm_bound}.
    
    For the second term, we have
    \begin{align*}
        & ~ \| \alpha(x) u(x) - \alpha(y) u(y) \|_2 \cdot \| v(y) \|_2 \\
        = & ~ \| \alpha(x) u(x) - \alpha(x) u(y) + \alpha(x) u(y) - \alpha(y) \alpha(y) \|_2 \cdot \| v(y) \|_2 \\
        = & ~ \| \alpha(x) ( u(x) - u(y) ) + ( \alpha(x) - \alpha(y) ) u(y) \|_2 \cdot \| v(y) \|_2 \\
        \leq & ~ ( \| \alpha(x) ( u(x) - u(y) ) \|_2 + \| ( \alpha(x) - \alpha(y) ) u(y) \|_2 ) \cdot \| v(y) \|_2 \\
        = & ~ ( | \alpha(x) | \cdot \|u(x) - u(y)\|_2 + \| ( \alpha(x) - \alpha(y) ) u(y) \|_2 ) \cdot \| v(y) \|_2 \\
        \leq & ~ ( | \alpha(x) | \cdot \|u(x) - u(y)\|_2 + \| u(y) \|_2 \cdot | \alpha(x) - \alpha(y) | ) \cdot \| v(y) \|_2 \\
         = & ~ | \alpha(x) | \cdot \|u(x) - u(y)\|_2 \cdot \| v(y) \|_2 \\
         & ~ +\| u(y) \|_2 \cdot | \alpha(x) - \alpha(y) | \cdot \| v(y) \|_2
    \end{align*}
    where the first equality follows from Fact~\ref{fact:vec_norm_bound}, the second equality follows from simple algebra, the third, fourth, fifth equalities follow from Fact~\ref{fact:vec_norm_bound}, the sixth equality follows from simple algebra.

    Then we combine two terms, we have
    \begin{align*}
        & ~ \| \alpha(x) u(x) v(x)^\top - \alpha(y) u(y) v(y)^\top \| \\
        \leq & ~ | \alpha(x) | \cdot \| u(x) \|_2 \cdot \| v(x) - v(y) \|_2 \\
        & ~ + | \alpha(x) | \cdot \|u(x) - u(y)\|_2 \cdot \| v(y) \|_2 \\
        & ~ + \| u(y) \|_2 \cdot | \alpha(x) - \alpha(y) | \cdot \| v(y) \|_2
    \end{align*}

{\bf Proof of Part 4.}
    \begin{align*}
        & ~ | \alpha(x)\beta(x) - \alpha(y)\beta(y) | \\
            = & ~ | \alpha(x) \beta(x) + \alpha(x) \beta(y) - \alpha(x) \beta(y) - \alpha(y) \beta(y) | \\
            = & ~ | \alpha(x) ( \beta(x) - \beta(y) ) + ( \alpha(x) - \alpha(y) ) \beta(y) | \\
            \leq & ~ | \alpha(x) ( \beta(x) - \beta(y) ) | + | ( \alpha(x) - \alpha(y) ) \beta(y) | \\
            = & ~ | \alpha(x) | \cdot | \beta(x) - \beta(y) | + | ( \alpha(x) - \alpha(y) ) \beta(y) | \\
            = & ~ | \alpha(x) | \cdot | \beta(x) - \beta(y) | + | \alpha(x) - \alpha(y) | \cdot | \beta(y) |
            \end{align*}
    where the first equality follows from Fact~\ref{fact:vector}, the second equality follows from simple algebra, the third, fourth, fifth equalities follow from Fact~\ref{fact:abs}.

{\bf Proof of Part 5.}
    \begin{align*}
        & ~ \| u(x) \circ v(x) - u(y) \circ v(y) \|_2 \\
        = & ~ \| u(x) \circ v(x) - u(x) \circ v(y) + u(x) \circ v(y) - u(y) \circ v(y)\|_2 \\
        = & ~ \| u(x) \circ ( v(x) - v(y) ) + ( u(x) - u(y) ) \circ v(y)\|_2 \\
        \leq & ~ \| u(x) \circ ( v(x) - v(y) ) \|_2 + \| ( u(x) - u(y) ) \circ v(y)\|_2 \\
        \leq & ~ \| u(x) \|_2 \cdot \| v(x) - v(y) \|_2 + \| ( u(x) - u(y) ) \circ v(y)\|_2 \\
        \leq & ~ \| u(x) \|_2 \cdot \| v(x) - v(y) \|_2 + \| u(x) - u(y) \|_2 \cdot \| v(y)\|_2
    \end{align*}
    where the first equality follows from Fact~\ref{fact:vector}, the second equality follows from simple algebra, the third, fourth, fifth equality follow from Fact~\ref{fact:vec_norm_bound}.
\end{proof}

\section{Copyright Regression}\label{app:cr_defs}

In this appendix, we reaffirm our definitions of Copyright Regression. In Appendix~\ref{subapp:def}, we provide our formal definitions of Softmax Regression and Copyright Regression. In Appendix~\ref{subapp:reg}, we provide our formal definitions of regularization term $L_{\mathrm{reg}}$ and train objective $L$.

\subsection{Definitions}\label{subapp:def}

\begin{definition}[Softmax Regression in \cite{dls23}]\label{def:f:formal}
Given a matrix $A \in \R^{n \times d}$, we define
\begin{align*}
    f(x) := \langle \exp(A x) , {\bf 1}_n \rangle^{-1} \exp(A x)
\end{align*}
\end{definition}

For the convenience of calculation, we define a intermediate operator $c(x)$ as follows
\begin{definition}\label{def:c:formal}
Given a matrix $A \in \R^{n \times d}$ and vector $b \in \R^n$, let $f(x)$ be denoted as Definition~\ref{def:f:informal}, we define
\begin{align*}
    c(x) := f(x) - b
\end{align*}
\end{definition}

We define train objective of Softmax Regression as follows
\begin{definition}[Train Objective of Softmax Regression in \cite{dls23}]\label{def:ell:formal}
Given a matrix $A \in \R^{n \times d}$ and vector $b \in \R^n$, let $c(x)$ be denoted as Definition~\ref{def:c:formal}, we define
\begin{align*}
    \ell(x) =  \langle c(x), c(x) \rangle
\end{align*}
\end{definition}

\begin{definition}[Softmax Regression on Copyrighted Data]\label{def:f_1:formal}
    Given all data matrix $A \in \R^{n \times d}$ and copyrighted data matrix $A_1 \in \R^{n_1 \times d}$, we define
    \begin{align*}
        f_1(x) := \langle \exp(A_{i, *} x) , {\bf 1}_n \rangle^{-1} \exp(A x)
    \end{align*}
    where $i \in [1, n_1]$ denote a integer.
\end{definition}

Also, we provide the definition of intermediate operator $c(x)$ as follows
\begin{definition}\label{def:c_1:formal}
    Given all data matrix $A \in \R^{n \times d}$ and copyrighted data matrix $A_1 \in \R^{n_1 \times d}$ and vector $b_1 \in \R^n$, let $f_1(x)$ be denoted as Definition~\ref{def:f_1:formal}, we define
    \begin{align*}
        c_1(x) := f_1(x) - b_1
    \end{align*}
\end{definition}

Now we have officially provided our definition of Copyright Regression below, which can prevent language models from infringing copyright with controllable performance damage and without occupying more resources.
\begin{definition}\label{def:L_copyright:formal}

We define $\ell(x)$ as Definition~\ref{def:ell:formal}. Denote $c_1(x)$ as Definition~\ref{def:c_1:formal}, and we define $\ell_1(x) = \langle c_1(x), c_1(x) \rangle$, define $\ell_2(x) := \ell(x) - \ell_1(x)$. Let $\gamma_c >0$ denote a parameter that control loss related to copyright data.

We consider the following copyright loss
\begin{align*}
    L_{\mathrm{copyright}}(x) := 0.5 \ell_1(x) + \gamma_c \cdot \ell_1(x)^{-1} + 0.5 \ell_2(x)
\end{align*}

\end{definition}

\subsection{Regularization}\label{subapp:reg}

\begin{definition}\label{def:L_reg:formal}
    Given a matrix $A \in \R^{n \times d}$. Given a vector $w \in \R^n$, we denote $W = \diag(w)$. We define $L_{\mathrm{reg}} : \R^d \to \R$ as follows
    \begin{align*}
        L_{\mathrm{reg}} := 0.5 \| W Ax \|_2^2
    \end{align*}
\end{definition}

After adding regularization term, we define our final train objective $L$ as follows
\begin{definition}\label{def:L:formal}
    Let $L_{\mathrm{copyright}}(x)$ be denoted as Definition~\ref{def:L_copyright:formal}, let $L_{\mathrm{reg}}$ be denoted as Definition~\ref{def:L_reg:formal}, then we define
    \begin{align*}
        L := L_{\mathrm{copyright}}(x) + L_{\mathrm{reg}}
    \end{align*}
\end{definition}

\section{Gradients and Hessians}\label{app:grad_hess}

In this appendix, we provide our computation results of several functions. In Appendix~\ref{subapp:grad_ell}, we provide our result and proof of gradient of $\ell(x)$. In Appendix~\ref{subapp:grad_ell-1}, we provide our result and proof of gradient of $\ell(x)^{-1}$. In Appendix~\ref{subapp:hess_ell}, we provide our result and proof of Hessian of $\ell(x)$. In Appendix~\ref{subapp:hess_ell-1}, we provide our result and proof of Hessian of $\ell(x)^{-1}$. In Appendix~\ref{subapp:grad_hess_L_reg}, we provide our result and proof of gradient and Hessian of $L_{\mathrm{reg}}$. In Appendix~\ref{subapp:grad_hess_L}, we provide our result and proof of gradient and Hessian of $L$.

\subsection{Gradient of \texorpdfstring{$\ell(x)$}{}}\label{subapp:grad_ell}

\begin{lemma}\label{lem:grad_ell}
If the given conditions are satisfied
\begin{itemize}
    \item Denote $c(x)$ as Definition~\ref{def:c:formal}
    \item Denote $f(x)$ as Definition~\ref{def:f:formal}
    \item Denote $\ell(x)$ as Definition~\ref{def:ell:formal} 
\end{itemize}
then we have
\begin{itemize}
\item  (see Part 7 of Lemma 5.6 in \cite{dls23})
\begin{align*}
\frac{\d 0.5 \ell(x) }{\d x_i } = A_{*,i}^\top ( - f(x) c(x)^\top f(x) + \diag(f(x)) c(x) )
\end{align*} 
\end{itemize}
\end{lemma} 
\begin{remark}
In \cite{dls23}, they write a typo in the equation, they forgot to add a negative sign. They write 
\begin{align*}
\frac{\d 0.5 \ell(x) }{\d x_i } = A_{*,i}^\top (  f(x) c(x)^\top f(x) + \diag(f(x)) c(x) )
\end{align*}
\end{remark}

\subsection{Gradient of \texorpdfstring{$\ell(x)^{-1}$}{}}\label{subapp:grad_ell-1}

\begin{lemma}\label{lem:grad_ell-1}
If the given conditions are satisfied
\begin{itemize}
    \item Denote $c(x)$ as Definition~\ref{def:c:formal}
    \item Denote $f(x)$ as Definition~\ref{def:f:formal}
    \item Denote $\ell(x)$ as Definition~\ref{def:ell:formal} 
\end{itemize}
then we have
\begin{itemize}
\item 
\begin{align*}
    \frac{\d 0.5 \ell(x)^{-1} }{ \d x_i} = \ell(x)^{-2} \cdot A_{*,i}^\top ( f(x) c(x)^\top f(x) - \diag(f(x)) c(x) )
\end{align*}
\end{itemize}
\end{lemma}
\begin{proof}
We have
\begin{align*}
     \frac{\d 0.5 \ell(x)^{-1} }{ \d x_i} 
     = & ~ - \cdot \ell(x)^{-2} \cdot \frac{\d 0.5 \ell(x) }{\d x_i} \\
     = & ~ - \ell(x)^{-2} \cdot A_{*,i}^\top ( - f(x) c(x)^\top f(x) + \diag(f(x)) c(x) ) \\
     = & ~ \ell(x)^{-2} \cdot A_{*,i}^\top ( f(x) c(x)^\top f(x) - \diag(f(x)) c(x) )
\end{align*}
where the first equality follows from differential chain rule, the second equality follow from Lemma~\ref{lem:grad_ell}, the third equality follows from simple algebra.

\end{proof}

\subsection{Hessian of \texorpdfstring{$\ell(x)$}{}}\label{subapp:hess_ell}

\begin{lemma}[Hessian of $0.5 \ell(x)$, Lemma 5.13 in \cite{dls23}]\label{lem:hessian_ell}
    We define
    \begin{itemize}
        \item $B_1(x) \in \R^{n_1 \times n_1}$ such that
        \begin{align*}
            A_{*,i}^\top B_1(x) A_{*,j}^\top := (-\langle f(x), A_{*, j} \rangle f(x) + f(x) \circ A_{*, j})^\top \cdot (-\langle f(x), A_{*, i} \rangle f(x) + f(x) \circ A_{*, i})
        \end{align*}
        \item $B_2(x) \in \R^{n_1 \times n_1}$ such that

        \begin{align*}
            A_{*,i}^\top B_2(x) A_{*,j}^\top 
            := & ~ c(x)^\top \cdot (2 \langle f(x), A_{*, i} \rangle \langle f(x), A_{*, j} \rangle f(x) - \langle f(x), A_{*, i} \circ A_{*, j} \rangle f(x) \\
            & ~ - \langle f(x), A_{*, i} \rangle f(x) \circ A_{*, j} - \langle f(x), A_{*, j} \rangle f(x) \circ A_{*, i} + A_{*, i} \circ f(x) \circ A_{*, j})
        \end{align*} 
    \end{itemize}
    Then we have 
    \begin{itemize}
        \item Part 1.
        \begin{align*}
            \frac{\d^2 0.5 \ell(x) }{ \d x_i \d x_i} = A_{*,i}^\top B_1(x) A_{*,i}^\top + A_{*,i}^\top B_2(x) A_{*,i}^\top
        \end{align*}
        \item Part 2.
        \begin{align*}
            \frac{\d^2 0.5 \ell(x) }{ \d x_i \d x_j} = A_{*,i}^\top B_1(x) A_{*,j}^\top + A_{*,i}^\top B_2(x) A_{*,j}^\top
        \end{align*}
    \end{itemize}
\end{lemma}

\begin{lemma}[Rewriting $B_1(x)$ and $B_2(x)$, see Part 3 of Lemma 5.15 in \cite{dls23}]\label{lem:rewrite_B}
    If the given conditions are satisfied
    \begin{itemize}
        \item Given a matrix $A \in \R^{n \times d}$.
        \item Denote $f(x)$ as Definition~\ref{def:f:formal}.
        \item Let $B(x) = B_1(x) + B_2(x)$
    \end{itemize}
    then, for $B(x) \in \R^{n \times n}$, we have
    \begin{align*}
        B(x) 
        = & ~\langle 3f(x) - 2b, f(x)\rangle \cdot f(x) f(x)^\top \\
        & ~ + \langle f(x) - b, f(x) \rangle \cdot \diag(f(x)) \\
        & ~ + \diag((2f(x) - b) \circ f(x)) \\
        & ~ + (b \circ f(x)) \cdot f(x)^\top + f(x) \cdot (b \circ f(x))^\top
    \end{align*}
    so we can rewrite hessian of $\ell(x)$ as follows
    \begin{itemize}
        \item Part 1.
        \begin{align*}
            \frac{\d^2 0.5 \ell(x) }{ \d x_i \d x_i} = A_{*,i}^\top B(x) A_{*,i}^\top
        \end{align*}
        \item Part 2.
        \begin{align*}
            \frac{\d^2 0.5 \ell(x) }{ \d x_i \d x_j} = A_{*,i}^\top B(x) A_{*,j}^\top
        \end{align*}
    \end{itemize}
\end{lemma}

For convenient, we define $B(x)$
\begin{definition}\label{def:B:formal}
    If the given conditions are satisfied
    \begin{itemize}
        \item Given vectors $b \in \R^n$ and $b_1 \in \R^n_1$
        \item Denote $f(x)$ as Definition~\ref{def:f:formal}
        \item Denote $f_1(x)$ as Definition~\ref{def:f_1:formal}
        \item Denote $c(x)$ as Definition~\ref{def:c:formal}
        \item Denote $c_1(x)$ as Definition~\ref{def:c:formal}
    \end{itemize}
    We define $B(x)$ and $B_c(x)$ as follow
    \begin{itemize}
        \item Part 1.
        \begin{align*}
            B(x) 
            = & ~\langle 3f(x) - 2b, f(x)\rangle \cdot f(x) f(x)^\top \\
            & ~ + \langle f(x) - b, f(x) \rangle \cdot \diag(f(x)) \\
            & ~ + \diag((2f(x) - b) \circ f(x)) \\
            & ~ + (b \circ f(x)) \cdot f(x)^\top + f(x) \cdot (b \circ f(x))^\top
        \end{align*}
        \item Part 2.
        \begin{align*}
            B_c(x)
            = & ~\langle 3f_1(x) - 2b_1, f_1(x)\rangle \cdot f_1(x) f_1(x)^\top \\
            & ~ + \langle f_1(x) - b_1, f_1(x) \rangle \cdot \diag(f_1(x)) \\
            & ~ + \diag((2f_1(x) - b_1) \circ f_1(x)) \\
            & ~ + (b_1 \circ f_1(x)) \cdot f_1(x)^\top + f_1(x) \cdot (b_1 \circ f_1(x))^\top
        \end{align*}
    \end{itemize}
\end{definition}

\subsection{Hessian of \texorpdfstring{$\ell(x)^{-1}$}{}}\label{subapp:hess_ell-1}

\begin{lemma}\label{lem:hessian_ell-1}
If the given condition is satisfied
\begin{itemize}
    \item Let $\ell(x)$ be denoted as Definition~\ref{def:ell:formal}.
\end{itemize}
then we have
\begin{itemize}
    \item Part 1.
    \begin{align*}
        \frac{\d^2 0.5 \ell(x)^{-1} }{ \d x_i \d x_i}
        = & ~ \ell(x)^{-2} (16 \cdot \ell(x)^{-1} \cdot (A_{*,i}^\top ( - f(x) c(x)^\top f(x) + \diag(f(x)) c(x) ))^2 \\
        & ~ - A_{*,i}^\top B(x) A_{*,i}^\top)
    \end{align*}
    \item Part 2.
    \begin{align*}
        \frac{\d^2 0.5 \ell(x)^{-1} }{ \d x_i \d x_j}
        = & ~ \ell(x)^{-2} (16 \cdot \ell(x)^{-1} \cdot  A_{*,i}^\top ( - f(x) c(x)^\top f(x) + \diag(f(x)) c(x) ) \\
        & ~ \cdot A_{*,j}^\top ( - f(x) c(x)^\top f(x) + \diag(f(x)) c(x) ) - A_{*,i}^\top B(x) A_{*,j}^\top) \\
    \end{align*}
\end{itemize}
\end{lemma}

\begin{proof}

\textbf{Proof of Part 1.} 
\begin{align*}
    \frac{\d^2 0.5 \ell(x)^{-1} }{ \d x_i \d x_i}
    = & ~ 16 \cdot \ell(x)^{-3} \cdot (\frac{\d 0.5 \ell(x) }{\d x_i})^2 - \ell(x)^{-2} \cdot \frac{\d^2 0.5 \ell(x)}{ \d x_i \d x_i} \\
    = & ~ \ell(x)^{-2} (16 \cdot \ell(x)^{-1} \cdot (\frac{\d 0.5 \ell(x) }{\d x_i})^2 - \frac{\d^2 0.5 \ell(x)}{ \d x_i \d x_i}) \\
    = & ~ \ell(x)^{-2} (16 \cdot \ell(x)^{-1} \cdot (A_{*,i}^\top ( - f(x) c(x)^\top f(x) + \diag(f(x)) c(x) ))^2 - \frac{\d^2 0.5 \ell(x)}{ \d x_i \d x_i}) \\
    = & ~ \ell(x)^{-2} (16 \cdot \ell(x)^{-1} \cdot (A_{*,i}^\top ( - f(x) c(x)^\top f(x) + \diag(f(x)) c(x) ))^2 - A_{*,i}^\top B(x) A_{*,i}^\top )
\end{align*}
where the first equality follows from Lemma~\ref{lem:diff_chain}, the second equality follows from simple algebra, the third equality follows from Lemma~\ref{lem:grad_ell}, the fourth equality follows from Lemma~\ref{lem:hessian_ell} and Lemma~\ref{lem:rewrite_B}.

\textbf{Proof of Part 2.}
\begin{align*}
    \frac{\d^2 0.5 \ell(x)^{-1} }{ \d x_i \d x_j}
    = & ~ 16 \cdot \ell(x)^{-3} \cdot \frac{\d 0.5 \ell(x)}{\d x_j} \cdot \frac{\d 0.5 \ell(x) }{\d x_i} - \ell(x)^{-2} \cdot \frac{\d^2 0.5 \ell(x)}{ \d x_i \d x_j} \\
    = & ~ \ell(x)^{-2} (16 \cdot \ell(x)^{-1} \cdot \frac{\d 0.5 \ell(x)}{\d x_i} \cdot \frac{\d 0.5 \ell(x) }{\d x_j} - \frac{\d^2 0.5 \ell(x)}{ \d x_i \d x_j}) \\
    = & ~ \ell(x)^{-2} (16 \cdot \ell(x)^{-1} \cdot A_{*,i}^\top ( - f(x) c(x)^\top f(x) + \diag(f(x)) c(x) ) \\
    & ~ \cdot A_{*,j}^\top ( - f(x) c(x)^\top f(x) + \diag(f(x)) c(x) ) - \frac{\d^2 0.5 \ell(x)}{ \d x_i \d x_j}) \\
    = & ~ \ell(x)^{-2} (16 \cdot \ell(x)^{-1} \cdot  A_{*,i}^\top ( - f(x) c(x)^\top f(x) + \diag(f(x)) c(x) ) \\
    & ~ \cdot A_{*,j}^\top ( - f(x) c(x)^\top f(x) + \diag(f(x)) c(x) ) - A_{*,i}^\top B(x) A_{*,j}^\top) 
\end{align*}
where the first equality follows from Lemma~\ref{lem:diff_chain}, the second equality follows from simple algebra, the third equality follows from Lemma~\ref{lem:grad_ell}, the fourth equality follows from Lemma~\ref{lem:hessian_ell} and Lemma~\ref{lem:rewrite_B}.
\end{proof}

\subsection{Gradient and Hessian of \texorpdfstring{$L_{\mathrm{reg}}$}{}}\label{subapp:grad_hess_L_reg}

\begin{lemma}\label{lem:grad_hess_reg}[Folklore, see \cite{lsz23} as an example]
    For a given vector $w \in \R^n$, let $W = \diag(w)$. Let $L_{\mathrm{reg}} : \R^d \to \R$ be denoted as Definition~\ref{def:L_reg:formal}.

    Then, we have
    \begin{itemize}
        \item The gradient of $L$ is
        \begin{align*}
            \frac{\d L_{\mathrm{reg}}}{\d x} = A^\top W^2 Ax
        \end{align*}
        \item The Hessian of $L$ is
        \begin{align*}
            \frac{\d^2 L_{\mathrm{reg}}}{\d x^2} = A^\top W^2 A
        \end{align*}
    \end{itemize}
\end{lemma}

\subsection{Gradient and Hessian of \texorpdfstring{$L$}{}}\label{app:hessian_L}\label{subapp:grad_hess_L}

\begin{lemma}[Formal vision of Lemma~\ref{lem:grad:informal}]\label{lem:grad:formal}
    If the given conditions are satisfied
    \begin{itemize}
        \item Given two matrices $A_1, A_2 \in \R^{n \times d}$, where $A_1$ is the part of data has copyright
        \item Let $\ell_1(x)$ and $\ell_2(x)$ be denoted as Definition~\ref{def:L_copyright:formal}.
        \item Let $L$ be denoted as Definition~\ref{def:L:formal}
        \item Let $B(x)$ be denoted ad Definition~\ref{def:B:formal}
    \end{itemize}
    we have
    \begin{align*}
        \frac{\d L}{\d x} = & ~ A_{*,i}^\top ( - f(x) c(x)^\top f(x) + \diag(f(x)) c(x) ) \\
        & ~ + 2 \gamma_c \ell_1(x)^{-2} \cdot {A_1}_{*,i}^\top ( f_1(x) c_1(x)^\top f_1(x) - \diag(f_1(x)) c_1(x) ) + A^\top W^2 Ax
    \end{align*}
\end{lemma}

\begin{proof}
    We have
    \begin{align*}
        \frac{\d L}{\d x} 
        = & ~ \frac{\d L}{\d x} \\
        = & ~ \frac{\d ( L_{\mathrm{copyright}} + L_{\mathrm{reg}} )}{\d x} \\
        = & ~ \frac{\d ( 0.5 \ell_1(x) + \gamma_c \ell_1(x)^{-1} + 0.5 \ell_2(x) + L_{\mathrm{reg}} )}{\d x} \\
        = & ~ \frac{\d ( 0.5 \ell(x) + \gamma_c \ell_1(x)^{-1} + L_{\mathrm{reg}} )}{\d x} \\
        = & ~ \frac{\d \gamma_c \ell_1(x)^{-1}}{\d x} + \frac{\d  L_{reg}}{\d x} + \frac{\d 0.5 \ell(x)}{\d x} \\
        = & ~ \gamma_c \frac{\d \ell_1(x)^{-1}}{\d x} + \frac{\d  L_{reg}}{\d x} + \frac{\d 0.5 \ell(x)}{\d x} \\
        = & ~ 2 \gamma_c \ell_1(x)^{-2} \cdot {A_1}_{*,i}^\top ( f_1(x) c_1(x)^\top f_1(x) - \diag(f_1(x)) c_1(x) ) + \frac{\d  L_{reg}}{\d x} + \frac{\d 0.5 \ell(x)}{\d x} \\
        = & ~ 2 \gamma_c \ell_1(x)^{-2} \cdot {A_1}_{*,i}^\top ( f_1(x) c_1(x)^\top f_1(x) - \diag(f_1(x)) c_1(x) ) + A^\top W^2 A x + \frac{\d 0.5 \ell(x)}{\d x} \\
        = & ~ A_{*,i}^\top ( - f(x) c(x)^\top f(x) + \diag(f(x)) c(x) ) \\
        & ~ + 2 \gamma_c \ell_1(x)^{-2} \cdot {A_1}_{*,i}^\top ( f_1(x) c_1(x)^\top f_1(x) - \diag(f_1(x)) c_1(x) ) + A^\top W^2 Ax
    \end{align*}
    where the first equality follows from Definition~\ref{def:L:formal}, the second equality follows from Definition~\ref{def:L_copyright:formal}, the third equality follows from $\ell(x) = \ell_1(x) + \ell_2(x)$, the fourth, fifth equalities follow from simple differential rules, the sixth equality follows from Lemma~\ref{lem:grad_ell-1}, the seventh equality follows from Part 1 of Lemma~\ref{lem:grad_hess_reg}, the eighth follows from Lemma~\ref{lem:grad_ell}.
\end{proof}

\begin{lemma}[Formal version of Lemma~\ref{lem:hessian:informal}]\label{lem:hessian:formal}
    If the given conditions are satisfied
    \begin{itemize}
        \item Given two matrices $A_1, A_2 \in \R^{n \times d}$, where $A_1$ is the part of data has copyright
        \item Let $\ell_1(x)$ and $\ell_2(x)$ be denoted as Definition~\ref{def:L_copyright:formal}.
        \item Let $L$ be denoted as Definition~\ref{def:L:formal}
        \item Let $B(x)$ and $B_c(x)$ be denoted ad Definition~\ref{def:B:formal}
    \end{itemize}
    we have
    \begin{itemize}
        \item Part 1.
        \begin{align*}
            \frac{\d^2 L}{\d x_i \d x_i}
            = & ~ A_{*,i}^\top B(x) A_{*,i}^\top +  A^\top W^2 A + 2\gamma_c \ell_1(x)^{-2} (16 \cdot \ell_1(x)^{-1} \cdot ({A_1}_{*,i}^\top ( - f_1(x) c_1(x)^\top f_1(x) \\
            & ~ + \diag(f_1(x)) c_1(x) ))^2 - {A_1}_{*,i}^\top B_1(x) {A_1}_{*,i}^\top)
        \end{align*}
        \item Part 2.
        \begin{align*}
            \frac{\d^2 L}{\d x_i \d x_j}
            = & ~ A_{*,i}^\top B(x) A_{*,j}^\top + A^\top W^2 A + 2\gamma_c \ell_1(x)^{-2} (16 \cdot \ell_1(x)^{-1} \cdot  {A_1}_{*,i}^\top ( - f_1(x) c_1(x)^\top f_1(x) \\ 
            & ~ + \diag(f_1(x)) c_1(x) ) \cdot {A_1}_{*,j}^\top ( - f_1(x) c_1(x)^\top f_1(x) + \diag(f_1(x)) c_1(x) ) - {A_1}_{*,i}^\top B_c(x) {A_1}_{*,j}^\top)
        \end{align*}
    \end{itemize}
\end{lemma}

\begin{proof}
    \textbf{Proof of Part 1.}
    \begin{align*}
        \frac{\d^2 L}{\d x_i \d x_i}
            = & ~ \frac{\d^2 ( L_{\mathrm{copyright}} + L_{\mathrm{reg}} )}{\d x_i \d x_i} \\
            = & ~ \frac{\d^2 ( 0.5 \ell_1(x) + \gamma_c \ell_1(x)^{-1} + 0.5 \ell_2(x) + L_{\mathrm{reg}} )}{\d x_i \d x_i} \\
            = & ~ \frac{\d^2 ( 0.5 \ell(x) + \gamma_c \ell_1(x)^{-1} + L_{\mathrm{reg}} )}{\d x_i \d x_i} \\
            = & ~ \frac{\d^2 \gamma_c \ell_1(x)^{-1}}{\d x_i \d x_i} + \frac{\d^2  L_{reg}}{\d x_i \d x_i} + \frac{\d^2 0.5 \ell(x)}{\d x_i \d x_i} \\
            = & ~ \gamma_c \frac{\d^2 \ell_1(x)^{-1}}{\d x_i \d x_i} + \frac{\d^2  L_{reg}}{\d x_i \d x_i} + \frac{\d^2 0.5 \ell(x)}{\d x_i \d x_i} \\
            = & ~ \gamma_c \frac{\d^2 \ell_1(x)^{-1}}{\d x_i \d x_i} + \frac{\d^2  L_{reg}}{\d x_i \d x_i} + A_{*,i}^\top B(x) A_{*,i}^\top \\
            = & ~ \gamma_c \frac{\d^2 \ell_1(x)^{-1}}{\d x_i \d x_i} + A^\top W^2 A + A_{*,i}^\top B(x) A_{*,i}^\top \\
            = & ~ A_{*,i}^\top B(x) A_{*,i}^\top +  A^\top W^2 A + 2\gamma_c \ell_1(x)^{-2} (16 \cdot \ell_1(x)^{-1} \cdot ({A_1}_{*,i}^\top ( - f_1(x) c_1(x)^\top f_1(x) \\
            & ~ + \diag(f_1(x)) c_1(x) ))^2 - {A_1}_{*,i}^\top B_1(x) {A_1}_{*,i}^\top)
    \end{align*}
    where the first equality follows from Definition~\ref{def:L:formal}, the second equality follows from Definition~\ref{def:L_copyright:formal}, the third equality follows from $\ell(x) = \ell_1(x) + \ell_2(x)$, the fourth, fifth equalities follow from simple differential rules, the sixth equality follows from Part 1 of Lemma~\ref{lem:rewrite_B}, the seventh equality follows from Part 2 of Lemma~\ref{lem:grad_hess_reg}, the eighth equality follows from Part 1 of Lemma~\ref{lem:hessian_ell-1}.

    \textbf{Proof of Part 2.}
    \begin{align*}
        \frac{\d^2 L}{\d x_i \d x_j}
            = & ~ \frac{\d^2 ( L_{\mathrm{copyright}} + L_{\mathrm{reg}} )}{\d x_i \d x_j} \\
            = & ~ \frac{\d^2 ( 0.5 \ell_1(x) + \gamma_c \ell_1(x)^{-1} + 0.5 \ell_2(x) + L_{\mathrm{reg}} )}{\d x_i \d x_j} \\
            = & ~ \frac{\d^2 ( 0.5 \ell(x) + \gamma_c \ell_1(x)^{-1} + L_{\mathrm{reg}} )}{\d x_i \d x_j} \\
            = & ~ \frac{\d^2 \gamma_c \ell_1(x)^{-1}}{\d x_i \d x_i} + \frac{\d^2  L_{reg}}{\d x_i \d x_i} + \frac{\d^2 0.5 \ell(x)}{\d x_i \d x_j} \\
            = & ~ \gamma_c \frac{\d^2 \ell_1(x)^{-1}}{\d x_i \d x_j} + \frac{\d^2  L_{reg}}{\d x_i \d x_j} + \frac{\d^2 0.5 \ell(x)}{\d x_i \d x_j} \\
            = & ~ \gamma_c \frac{\d^2 \ell_1(x)^{-1}}{\d x_i \d x_j} + \frac{\d^2  L_{reg}}{\d x_i \d x_j} + A_{*,i}^\top B(x) A_{*,i}^\top \\
            = & ~ \gamma_c \frac{\d^2 \ell_1(x)^{-1}}{\d x_i \d x_j} + A^\top W^2 A + A_{*,i}^\top B(x) A_{*,i}^\top \\
            = & ~ A_{*,i}^\top B(x) A_{*,j}^\top + A^\top W^2 A + 2\gamma_c \ell_1(x)^{-2} (16 \cdot \ell_1(x)^{-1} \cdot  {A_1}_{*,i}^\top ( - f_1(x) c_1(x)^\top f_1(x) \\ 
            & ~ + \diag(f_1(x)) c_1(x) ) \cdot {A_1}_{*,j}^\top ( - f_1(x) c_1(x)^\top f_1(x) + \diag(f_1(x)) c_1(x) ) - {A_1}_{*,i}^\top B_c(x) {A_1}_{*,j}^\top)
    \end{align*}
    where the first equality follows from Definition~\ref{def:L:formal}, the second equality follows from Definition~\ref{def:L_copyright:formal}, the third equality follows from $\ell(x) = \ell_1(x) + \ell_2(x)$, the fourth, fifth equalities follow from simple differential rules, the sixth equality follows from Part 2 of Lemma~\ref{lem:rewrite_B}, the seventh equality follows from Part 2 of Lemma~\ref{lem:grad_hess_reg}, the eighth equality follows from Part 2 of Lemma~\ref{lem:hessian_ell-1}.
\end{proof}

\begin{lemma}\label{lem:Hessian_L}
    If the given conditions are satisfied
    \begin{itemize}
        \item Given two matrices $A_1, A_2 \in \R^{n \times d}$, where $A_1$ is the part of data has copyright
        \item Let $\ell_1(x)$ and $\ell_2(x)$ be denoted as Definition~\ref{def:L_copyright:formal}.
        \item Let $L$ be denoted as Definition~\ref{def:L:formal}
        \item Let $B(x)$ be denoted ad Definition~\ref{def:B:formal}
        \item Denote $H(x) := \frac{\d^2 L}{\d x^2}$
        \item Denote $H_1(x) := \frac{\d^2 \gamma_c \cdot \ell_1(x)^{-1}}{\d x^2}$
        \item Denote $H_2(x) := \frac{\d^2 ( 0.5 \ell_1(x) + L_{reg}(x) )}{\d x^2}$
        \item Denote $H_3(x) := \frac{\d^2 0.5 \ell_2(x)}{\d x^2}$
    \end{itemize}
    \begin{align*}
        H(x) = H_1(x) + H_2(x) + H_3(x)
    \end{align*}
\end{lemma}

\begin{proof}
    We have
    \begin{align*}
        H(x) = & ~ \frac{\d^2 L}{\d x^2} \\
        = & ~ \frac{\d^2 (L_{copyright}(x) + L_{reg})}{\d x^2} \\
        = & ~ \frac{\d^2 (0.5 \ell_1(x) + \gamma_c \cdot \ell_1(x)^{-1} + 0.5 \ell_2(x) + L_{reg})}{\d x^2} \\
        = & ~ \frac{\d^2 \gamma_c \cdot \ell_1(x)^{-1}}{\d x^2} + \frac{\d^2 (0.5 \ell_1(x) + L_{reg})}{\d x^2} + \frac{\d^2 0.5 \ell_2(x)}{\d x^2} \\
        = & ~ H_1(x) + \frac{\d^2 (0.5 \ell_1(x) + 0.5 \ell_2(x) + L_{reg})}{\d x^2} + \frac{\d^2 0.5 \ell_2(x)}{\d x^2} \\
        = & ~ H_1(x) + H_2(x) + \frac{\d^2 0.5 \ell_2(x)}{\d x^2} \\
        = & ~ H_1(x) + H_2(x) + H_3(x)
    \end{align*}
    where the first equality follows from the Definition of $H(x)$, the second equality follows from Definition~\ref{def:L:formal}, the third equality follows from simple differential rule, the fourth equality follows from the Definition of $H_1(x)$, the fifth equality follows from the Definition of $H_2(x)$, the sixth equality follows from the Definition of $H_3(x)$.
\end{proof}

\section{Lower Bound on Hessian}\label{app:psd}

In this appendix, we prove that $\nabla^2 L \succeq 0$ and thus $L$ is convex. In Appendix~\ref{subapp:psd_main_result}, we provide our main result that $\nabla^2 L \succeq 0$. In Appendix~\ref{subapp:def_P}, we divide $H_1(x)$ to 5 part to help us to compute lower bound on $H_1(x)$. In Appendix~\ref{subapp:lower_bound_P}, we provide our result and proof of lower bound on $P(x)$. In Appendix~\ref{subapp:psd_helpful_lemma}, we provide some helpful lemma to simplify our proofs. In Appendix~\ref{subapp:lower_bound_P1}, we provide our result and proof of lower bound on $P(x)_1$. In Appendix~\ref{subapp:lower_bound_P2.5}, we provide our result and proof of lower bound on $P(x)_{2.5}$. In Appendix~\ref{subapp:lower_bound_P4}, we provide our result and proof of lower bound on $P(x)_4$. In Appendix~\ref{subapp:lower_bound_P5}, we provide our result and proof of lower bound on $P(x)_5$.

\subsection{Main Result}\label{subapp:psd_main_result}

\begin{lemma}[Formal version of Lemma~\ref{lem:psd:informal}]\label{lem:psd:formal}
    If the given conditions are satisfied
    \begin{itemize}
        \item Given two matrices $A_1, A_2 \in \R^{n \times d}$, where $A_1$ is the part of data has copyright
        \item Let $\ell_1(x)$ and $\ell_2(x)$ be denoted as Definition~\ref{def:L_copyright:formal}
        \item Let $L$ be denoted as Definition~\ref{def:L:formal}
        \item Denote $P(x)$ as Definition~\ref{def:P}
        \item Denote $B(x)$ as Definition~\ref{def:B:formal}
        \item $\gamma \in (0, 1)$
        \item Given a vector $w$, let $W = \diag(w) \in \R^{n \times n}$. Let $W^2 \in \R^{n \times n}$ denote the matrix that i-th diagonal entry is $w_{i, i}^2$
        \item Denote $H(x) := \frac{\d^2 L}{\d x^2}$
        \item Denote $H_1(x) := \frac{\d^2 \gamma_c \cdot \ell_1(x)^{-1}}{\d x^2}$
        \item Denote $H_2(x) := \frac{\d^2 ( 0.5 \ell_1(x) + L_{reg}(x) )}{\d x^2}$
        \item Denote $H_3(x) := \frac{\d^2 0.5 \ell_2(x)}{\d x^2}$
        \item Denote $B_c(x)$ that $\frac{\d^2 0.5 \ell_1(x)}{\d x^2} = A^\top B_c(x) A$
        \item Denote $B_{nc}(x)$ that $\frac{\d^2 0.5 \ell_2(x)}{\d x^2} = A^\top B_{nc}(x) A$
        \item Let $l > 0$ denote a scalar
    \end{itemize}
    if for all $i \in [n]$, $w_i^2 \geq 8 + 200 \gamma_c \gamma^{-3} + l / \sigma_{\min}(A)^2$, we have
    \begin{align*}
        H(x) \succeq l \cdot I_d
    \end{align*}
\end{lemma}

\begin{proof}
    We have
    \begin{align*}
        H(x)
        = & ~ H_1(x) + H_2(x) + H_3(x) \\
        = & ~ A^\top P(x) A + H_2(x) + H_3(x) \\
        = & ~ A^\top P(x) A + A^\top ( B_c(x) + W^2 ) + H_3(x) \\
        = & ~ A^\top P(x) A + A^\top ( B_c(x) + W^2 ) + A^\top B_{nc}(x) A \\
        = & ~ A^\top ( P(x) + B_c(x) + W^2 + B_{nc}(x) ) A
    \end{align*}
    where the first equality follows from Lemma~\ref{lem:Hessian_L}, the second equality follows from Definition~\ref{def:P}, the third equality follows from Lemma~\ref{lem:grad_hess_reg} and the Definition of $B_c(x)$, the fourth equality follows from the Definition of $B_{nc}(x)$, the last equality follows from simple algebra.
    
    Let
    \begin{align*}
        D := P(x) + B_c(x) + W^2 + B_{nc}(x)
    \end{align*}
    then, $\frac{\d^2 L}{\d x^2}$ can be rewrite as
    \begin{align}\label{eq:D}
        H(x) = A^\top D A
    \end{align}
    now we have the boundary of $D$ as follows
    \begin{align*}
        D
        = & ~ P(x) + B_c(x) + W^2 + B_{nc}(x) \\
        \succeq & ~ - 200 \gamma_c \gamma^{-3} \cdot I_n + B_c(x) + W^2 + B_{nc}(x) \\
        \succeq & ~ - 200 \gamma_c \gamma^{-3} \cdot I_n - 4 I_n + W^2 + B_{nc}(x) \\
        \succeq & ~ - 200 \gamma_c \gamma^{-3} \cdot I_n - 4 I_n + W^2 - 4 I_n \\
        \succeq & ~ - 200 \gamma_c \gamma^{-3} \cdot I_n - 4 I_n + w_{\min}^2 - 4 I_n \\
    \end{align*}
    where the first equality follows from the Definition of $D$, the second equality follows from 
    Lemma~\ref{lem:lower_bound_P}, the third and fourth equalities follow from Lemma~\ref{lem:B_lower_bound}, the fifth equality follows from Fact~\ref{fact:psd}.

    When $w_{\min}^2 \geq 8 + 200 \gamma_c \gamma^{-3} + l / \sigma_{\min}(A)^2$, we have
    \begin{align}\label{eq:lower_bound_D}
        D
        \succeq \frac{l}{\sigma_{\min}(A)^2}I_n
    \end{align}
    then
    \begin{align*}
        H(x)
        = & ~ A^\top D A \\
        \succeq & ~ \sigma_{\min}(D) \cdot \sigma_{\min}(A)^2 I_d \\
        \succeq & ~ l \cdot I_d
    \end{align*}
    where the first equality follows from Eq~\eqref{eq:D}, the second equality follows from Fact~\ref{fact:psd}, the last equality follows from Eq~\eqref{eq:lower_bound_D}.
\end{proof}

\subsection{Definition of matrix functions \texorpdfstring{$P_i(x)$}{}}\label{subapp:def_P}
\begin{definition}\label{def:P}
If the given conditions are satisfied
\begin{itemize}
    \item Let $\ell(x)$ be denoted as Definition~\ref{def:ell:formal} 
    \item Denote $f(x)$ as Definition~\ref{def:f:formal}
    \item Denote $c(x)$ as Definition~\ref{def:c:formal}
    \item Denote $B(x)$ as Definition~\ref{def:B:formal}
    \item Let $H_1(x) := \frac{\d^2 \gamma_c \ell_1(x)^{-1}}{\d x^2}$
    \item Denote $\gamma_c > 0$ a scalar
\end{itemize}
We define $P_1(x), P_2(x), P_3(x), P_4(x), P_5(x) \in \R^{n \times n}$ as follows
\begin{itemize}
    \item $P_1(x) = \underbrace{ \ell(x)^{-3} }_{\mathrm{scalar}} \cdot \underbrace{ \langle f(x), c(x) \rangle^2 }_{ \mathrm{scalar} } \cdot \underbrace{ f(x) }_{n \times 1} \cdot \underbrace{ f(x)^\top }_{1 \times n}$
    \item $P_2(x) = \underbrace{ \ell(x)^{-3} }_{ \mathrm{scalar} } \cdot \underbrace{ \langle f(x), c(x) \rangle }_{ \mathrm{scalar} } \cdot \underbrace{ ( f(x) \circ c(x) ) }_{ n \times 1 } \cdot \underbrace{ f(x)^\top }_{1 \times n}$
    \item $P_3(x) = \underbrace{ \ell(x)^{-3} }_{ \mathrm{scalar} } \cdot \underbrace{ \langle f(x), c(x) \rangle }_{ \mathrm{scalar} } \cdot \underbrace{ f(x) }_{n \times 1} \cdot \underbrace{ ( f(x) \circ c(x) )^\top }_{1 \times n}$ 
    \item $P_4(x) = \underbrace{ \ell(x)^{-3} }_{ \mathrm{scalar} } \cdot \underbrace{ (f(x) \circ c(x)) }_{n \times 1} \cdot \underbrace{ (f(x) \circ c(x))^\top }_{1 \times n}$
    \item $P_5 (x) = \underbrace{ \ell(x)^{-2} }_{ \mathrm{scalar} } \cdot \underbrace{ B(x) }_{n \times n}$
\end{itemize}

We define $P(x) \in \R^{n \times n}$ as follows
\begin{align*}
 P(x) := 32 \gamma_c ( P_1(x) - P_2(x) - P_3(x) + P_4(x) ) - P_5(x)
\end{align*}

Note that $H_1(x) = A^\top P(x) A$.
\end{definition}

\subsection{Lower Bound Property for Matrix Function \texorpdfstring{$P(x)$}{}}\label{subapp:lower_bound_P}

\begin{lemma}\label{lem:lower_bound_P}
    If the given conditions are satisfied
    \begin{itemize}
        \item Given a matrix $A \in \R^{n \times d}$.
        \item $\gamma \in (0, 1)$
        \item Denote $f(x)$ as Definition~\ref{def:f:formal}
        \item Denote $c(x)$ as Definition~\ref{def:c:formal}
        \item Let $\ell(x)$ be denoted as Definition~\ref{def:ell:formal}
        \item Denote $P(x)$ as Definition~\ref{def:P}
        \item $\ell(x) \geq \gamma$
        \item $\| f(x) \|_2 \leq 1$
    \end{itemize}
    we have
    \begin{align*}
         P(x) \succeq - 104 \gamma^{-3} \cdot I_n
    \end{align*}
\end{lemma}

\begin{proof}
    We have
    \begin{align*}
        P(x)
        = & ~ 16 ( P_1(x) - P_2(x) - P_3(x) + P_4(x) ) - P_5(x) \\
        \succeq & ~ 32 \gamma_c ( - P_2(x) - P_3(x) + P_4(x) ) - P_5(x) \\
        \succeq & ~ 32 \gamma_c ( - P_2(x) - P_3(x) ) - P_5(x) \\
        \succeq & ~ - 32 \gamma_c P_{2.5}(x) - P_5(x) \\
        \succeq & ~ - 32 \gamma_c \cdot 6 \gamma^{-3} \cdot f(x) f(x)^\top - P_5(x) \\
        \succeq & ~ - 32 \gamma_c \cdot 6 \gamma^{-3} \cdot f(x) f(x)^\top - 8 \gamma^{2} \cdot I_n \\
        \succeq & ~ - 192 \gamma_c \gamma^{-3} \cdot f(x) f(x)^\top - 8 \gamma^{2} \cdot I_n \\
        \succeq & ~ - 192 \gamma_c \gamma^{-3} \cdot \| f(x) \|_2^2 \cdot I_n - 8 \gamma^{2} \cdot I_n \\
        \succeq & ~ - 192 \gamma_c \gamma^{-3} \cdot I_n - 8 \gamma^{2} \cdot I_n \\
        \succeq & ~ - 192 \gamma_c \gamma^{-3} \cdot I_n - 8 \gamma^{3} \cdot I_n \\
        \succeq & ~ - 200 \gamma_c \gamma^{-3} \cdot I_n
    \end{align*}
    where the first equality follows from Definition~\ref{def:P}, the second equality follows from Lemma~\ref{lem:lower_bound_P1}, the third equality follows from Lemma~\ref{lem:lower_bound_P4}, the fourth equality follows from Lemma~\ref{lem:P_2.5}, the fifth equality follows from Lemma~\ref{lem:lower_bound_P2.5}, the sixth equality follows from Lemma~\ref{lem:lower_bound_P5}, the seventh equality follows from simple algebra, the eighth equality follows from Fact~\ref{fact:psd}, the ninth equality follows from $\| f(x) \|_2 \leq 1$, the tenth equality follows from $\gamma \in (0, 1)$, the 1first equality follows from simple algebra.
\end{proof}

\subsection{Helpful Lemma}\label{subapp:psd_helpful_lemma}

\begin{lemma}\label{lem:B_lower_bound}
    If the given conditions are satisfied
    \begin{itemize}
        \item Let $B(x) \in \R^{n \times n}$ be denoted as Definition~\ref{def:B:formal}
        \item Let $f(x) \geq 0_n$
        \item Let $b \geq 0_n$
        \item $\| f(x) \|_2 \leq 1$
        \item $\| b \|_2 \leq 1$
    \end{itemize}
    we have
    \begin{itemize}
        \item (see Part 5 of Lemma 6.2 in \cite{dls23})
        \begin{align*}
            - 4 I_n \preceq B(x) \preceq 8 I_n
        \end{align*}
    \end{itemize}
\end{lemma}

\begin{lemma}\label{lem:P_2.5}
We define 
\begin{align*}
    P_{2.5} = \ell(x)^{-3} \cdot | \langle f(x), c(x) \rangle | \cdot (  ( f(x) \circ c(x) ) \cdot (f(x) \circ c(x) )^\top  + f(x) f(x)^\top )
\end{align*}
Then, it is obvious that that
\begin{align*}
  - P_{2.5}(x) \preceq  P_2 (x) + P_3(x) \preceq P_{2.5}(x) 
\end{align*}    
\end{lemma}
\begin{proof}

The reason is
\begin{align*}
- aa^\top + bb^\top 
\preceq ab^\top + b a^\top \preceq aa^\top + bb^\top
\end{align*}

\end{proof}

\subsection{Lower Bound Property for Matrix Function \texorpdfstring{$P_1(x)$}{}}\label{subapp:lower_bound_P1}

\begin{lemma}\label{lem:lower_bound_P1}
    If the given conditions are satisfied
    \begin{itemize}
        \item Given a matrix $A \in \R^{n \times d}$.
        \item $\gamma \in (0, 1)$
        \item Denote $f(x)$ as Definition~\ref{def:f:formal}
        \item Denote $c(x)$ as Definition~\ref{def:c:formal}
        \item Let $\ell(x)$ be denoted as Definition~\ref{def:ell:formal}
        \item Denote $P_1(x)$ as Definition~\ref{def:P}
        \item $\ell(x) \geq \gamma$
    \end{itemize}
    we have
    \begin{align*}
        0 \preceq P_1(x) \preceq 4 \gamma^{-3} \cdot f(x) \cdot f(x)^\top
    \end{align*}
\end{lemma}

\begin{proof}
    On one hand, we can show that
    \begin{align*}
        P_1(x)
        = & ~ \ell(x)^{-3} \cdot \langle f(x), c(x) \rangle^2 \cdot f(x) \cdot f(x)^\top \\
        \succeq & ~ 0
    \end{align*}
    where the first equality follows from Part 1 of Lemma~\ref{lem:func_bounds}, the second equality follows from Part 2 of Lemma~\ref{lem:func_bounds}.

    On the other hand, we have
    \begin{align*}
        P_1(x)
        = & ~ \ell(x)^{-3} \cdot \langle f(x), c(x) \rangle^2 \cdot f(x) \cdot f(x)^\top \\
        leq & ~ \gamma^{-3} \cdot \langle f(x), c(x) \rangle^2 \cdot f(x) \cdot f(x)^\top \\
        \preceq & ~ 2^2 \gamma^{-3} \cdot f(x) \cdot f(x)^\top \\
        = & ~ 4 \gamma^{-3} \cdot f(x) \cdot f(x)^\top 
    \end{align*}
    where the first equality follows from Definition~\ref{def:P}, the second equality follows from $\ell(x) \geq \gamma$, the third equality follows from Part 2 of Lemma~\ref{lem:func_bounds}, the fourth equality follows from simple algebra.

\end{proof}

\subsection{Lower Bound Property for Matrix Function \texorpdfstring{$P_{2.5}(x)$}{}}\label{subapp:lower_bound_P2.5}

\begin{lemma}\label{lem:lower_bound_P2.5}
    If the given conditions are satisfied
    \begin{itemize}
        \item Given a matrix $A \in \R^{n \times d}$.
        \item $\gamma \in (0, 1)$
        \item Denote $f(x)$ as Definition~\ref{def:f:formal}
        \item Denote $c(x)$ as Definition~\ref{def:c:formal}
        \item Let $\ell(x)$ be denoted as Definition~\ref{def:ell:formal}
        \item Let $P_{2.5}(x)$ be denoted as Lemma~\ref{lem:P_2.5}
        \item $\ell(x) \geq \gamma$
    \end{itemize}
    we have
    \begin{align*}
        0 \preceq P_{2.5}(x) \preceq 6 \gamma^{-3} \cdot f(x) f(x)^\top
    \end{align*}
\end{lemma}

\begin{proof}
    On one hand, we can show that
    \begin{align*}
        P_{2.5}(x)
        = & ~ \ell(x)^{-3} \cdot | \langle f(x), c(x) \rangle | \cdot (  ( f(x) \circ c(x) ) \cdot (f(x) \circ c(x) )^\top  + f(x) f(x)^\top ) \\
        \succeq & ~ 0
    \end{align*}
    where the first equality follows from the Definition of $P_{2.5}(x)$, the second equality follows from Part 2 of Lemma~\ref{lem:func_bounds}.

    On the other hand, we have
    \begin{align*}
        P_{2.5}(x)
        = & ~ \ell(x)^{-3} \cdot | \langle f(x), c(x) \rangle | \cdot (  ( f(x) \circ c(x) ) \cdot (f(x) \circ c(x) )^\top  + f(x) f(x)^\top ) \\
        \leq & ~ \gamma^{-3} \cdot | \langle f(x), c(x) \rangle | \cdot (  ( f(x) \circ c(x) ) \cdot (f(x) \circ c(x) )^\top  + f(x) f(x)^\top ) \\
        \leq & ~ \gamma^{-3} \cdot 2 \cdot (  ( f(x) \circ c(x) ) \cdot (f(x) \circ c(x) )^\top  + f(x) f(x)^\top ) \\
        \preceq & ~ \gamma^{-3} \cdot 2 \cdot ( \| c(x) \|_\infty + 1 ) \cdot f(x) f(x)^\top  \\
        \leq & ~ \gamma^{-3} \cdot 2 \cdot ( \| c(x) \|_2 + 1 ) \cdot f(x) f(x)^\top  \\
        \leq & ~ \gamma^{-3} \cdot 2 \cdot ( 2 + 1 ) \cdot f(x) f(x)^\top  \\
        = & ~ 6 \gamma^{-3} \cdot f(x) f(x)^\top
    \end{align*}
    where the first equality follows from the Definition of $P_{2.5}(x)$, the second equality follows from $\ell(x) \geq \gamma$, the third equality follows from Part 2 of Lemma~\ref{lem:func_bounds}, the fourth equality follows from Fact~\ref{fact:psd}, the fifth equality follows from Fact~\ref{fact:vec_norm_bound}, the sixth equality follows from Part 1 of Lemma~\ref{lem:vec_func_norm_bound}, the last equality follows from simple algebra.
\end{proof}

\subsection{Lower Bound Property for Matrix Function \texorpdfstring{$P_4(x)$}{}}\label{subapp:lower_bound_P4}

\begin{lemma}\label{lem:lower_bound_P4}
    If the given conditions are satisfied
    \begin{itemize}
        \item Given a matrix $A \in \R^{n \times d}$.
        \item $\gamma \in (0, 1)$
        \item Denote $f(x)$ as Definition~\ref{def:f:formal}
        \item Denote $c(x)$ as Definition~\ref{def:c:formal}
        \item Let $\ell(x)$ be denoted as Definition~\ref{def:ell:formal}
        \item Denote $P_4(x)$ as Definition~\ref{def:P}
        \item $\ell(x) \geq \gamma$
    \end{itemize}
    we have
    \begin{align*}
       0 \preceq P_4(x) \preceq 2 \gamma^{-3} \cdot f(x) \cdot f(x)^\top
    \end{align*}
\end{lemma}

\begin{proof}
    On one hand, we can show that
    \begin{align*}
        P_4(x)
        = & ~ \ell(x)^{-3} \cdot ( f(x) \circ c(x) ) \cdot ( f(x) \circ c(x) )^\top \\
        \succeq & ~ 0
    \end{align*}
    where the first equality follows from Definition~\ref{def:P}, the second equality follows from Fact~\ref{fact:psd}.

    On the other hand, we have
    \begin{align*}
        P_4(x)
        = & ~ \ell(x)^{-3} \cdot ( f(x) \circ c(x) ) \cdot ( f(x) \circ c(x) )^\top \\
        \leq & ~ \gamma^{-3} \cdot ( f(x) \circ c(x) ) \cdot ( f(x) \circ c(x) )^\top \\
        \leq & ~ \gamma^{-3} \cdot \| c(x) \|_\infty \cdot f(x) f(x)^\top \\
        \leq & ~ \gamma^{-3} \cdot \| c(x) \|_2 \cdot f(x) f(x)^\top \\
        \leq & ~ \gamma^{-3} \cdot 2 \cdot f(x) f(x)^\top \\
        = & ~ 2 \gamma^{-3} \cdot f(x) f(x)^\top
    \end{align*}
    where the first equality follows from Definition~\ref{def:P}, the second equality follows from $\ell(x) \geq \gamma$, the third equality follows from Fact~\ref{fact:psd}, the fourth equality follows from Fact~\ref{fact:vec_norm_bound}, the fifth equality follows from Part 1 of Lemma~\ref{lem:vec_func_norm_bound}, the last equality follows from simple algebra.
\end{proof}

\subsection{Lower Bound Property for Matrix Function \texorpdfstring{$P_5(x)$}{}}\label{subapp:lower_bound_P5}

\begin{lemma}\label{lem:lower_bound_P5}
    If the given conditions are satisfied
    \begin{itemize}
        \item Given a matrix $A \in \R^{n \times d}$.
        \item $\gamma \in (0, 1)$
        \item Denote $f(x)$ as Definition~\ref{def:f:formal}
        \item Denote $c(x)$ as Definition~\ref{def:c:formal}
        \item Let $\ell(x)$ be denoted as Definition~\ref{def:ell:formal}
        \item Denote $P_5(x)$ as Definition~\ref{def:P}
        \item $\ell(x) \geq \gamma$
    \end{itemize}
    we have
    \begin{align*}
       - \frac{1}{4} I_n \preceq P_5(x) \preceq 8 \gamma^{2} \cdot I_n
    \end{align*}
\end{lemma}

\begin{proof}
    On one hand, we can show that
    \begin{align*}
        P_5(x)
        = & ~ \ell(x)^{-2} \cdot B(x) \\
        \succeq & ~ \ell(x)^{-2} \cdot ( - 4 I_n ) \\
        \geq & ~ \frac{1}{16} \cdot ( - 4 I_n ) \\
        = & ~ - \frac{1}{4} I_n
    \end{align*}
    where the first equality follows from Definition~\ref{def:P}, the second equality follows from Lemma~\ref{lem:B_lower_bound}, the third equality follows from Part 1 of Lemma~\ref{lem:func_bounds}, the fourth equality follows from simple algebra.

    On the other hand, we have
    \begin{align*}
        P_5(x)
        = & ~ \ell(x)^{-2} \cdot B(x) \\
        \preceq & ~ \ell(x)^{-2} \cdot 8 I_n \\
        \leq & ~ \gamma^{-2} \cdot 8 I_n \\
        = & ~ 8 \gamma^{2} \cdot I_n
    \end{align*}
    where the first equality follows from Definition~\ref{def:P}, the second equality follows from Lemma~\ref{lem:B_lower_bound}, the third equality follows from $\ell(x) \geq \gamma$, the fourth equality follows from simple algebra.
\end{proof}

\section{Hessian is Lipschitz}\label{sec:lipschitz}\label{app:lipschitz}

In this appendix, we provide our result that Hessian of $L$ is Lipschitz. In Appendix~\ref{subapp:lipschitz_main_result}, we provide our result and proof of $\|\nabla^2 L(x) - \nabla^2 L(y)\| \leq ( 13344 \gamma_c + 2 ) \gamma^{-4} \beta^{-2} n^{1.5} \exp(40R^2) \| x - y \|_2$. In Appendix~\ref{subapp:lipschitz_helpful_lemma}, we provide some helpful lemmas to simplify proofs. In Appendix~\ref{subapp:def_Q}, we divide $H_1(x)$ to several part to help us to compute its Lipschitz property. In Appendix~\ref{subapp:lipschitz_Q1}, we provide our result and proof of Lipschitz property for matrix function $Q_1(x)$. In Appendix~\ref{subapp:lipschitz_Q2}, we provide our result and proof of Lipschitz property for matrix function $Q_2(x)$. In Appendix~\ref{subapp:lipschitz_Q3}, we provide our result and proof of Lipschitz property for matrix function $Q_3(x)$. In Appendix~\ref{subapp:lipschitz_Q4}, we provide our result and proof of Lipschitz property for matrix function $Q_4(x)$. In Appendix~\ref{subapp:lipschitz_Q5}, we provide our result and proof of Lipschitz property for matrix function $Q_5(x)$.

\subsection{Main Result}\label{subapp:lipschitz_main_result}

\begin{lemma}[Formal version of Lemma~\ref{lem:lipschitz:informal}]\label{lem:lipschitz:formal}
If the given conditions are satisfied
\begin{itemize}
    \item Given two matrices $A_1, A_2 \in \R^{n \times d}$, where $A_1$ is the part of data has copyright
    \item Let $L$ be denoted as Definition~\ref{def:L:formal}
    \item Let $\gamma \in (0,1)$
    \item Let $\beta \in (0, 0.1)$
    \item Let $\gamma_c > 0$ denote a scalar
    \item Let $R \geq 4$
    \item $\ell(x) \geq \gamma$
    \item $H(x) := \frac{\d^2 L}{\d x^2}$
\end{itemize}
Then, we have
\begin{itemize}
    \item 
    \begin{align*}
        \| H(x) - H(y) \| \leq ( 13344 \gamma_c + 2 ) \gamma^{-4} \beta^{-2} n^{1.5} \exp(40R^2) \| x - y \|_2
    \end{align*}
\end{itemize}
\end{lemma}

\begin{proof}
    \begin{align*}
        & ~ \| H(x) - H(y) \| \\
        = & ~ \| ( H_1(x) + H_2(x) + H_3(x) ) - ( H_1(y) + H_2(y) + H_3(y) ) \| \\
        \leq & ~ \| H_1(x) - H_1(y) \| + \| H_2(x) - H_2(y) \| + \| H_3(x) - H_3(y) \| \\
        \leq & ~ 13344 \gamma_c \gamma^{-4} \beta^{-2} n^{1.5} \exp(40R^2) \| x - y \|_2 + \| H_2(x) - H_2(y) \| + \| H_3(x) - H_3(y) \| \\
        \leq & ~ 13344 \gamma_c \gamma^{-4} \beta^{-2} n^{1.5} \exp(40R^2) \| x - y \|_2 + \beta^{-2} n^{1.5}\exp(20R^2)\|x - y\|_2 + \| H_3(x) - H_3(y) \| \\
        \leq & ~ 13344 \gamma_c \gamma^{-4} \beta^{-2} n^{1.5} \exp(40R^2) \| x - y \|_2 + \beta^{-2} n^{1.5}\exp(20R^2)\|x - y\|_2 + \beta^{-2} n^{1.5}\exp(20R^2)\|x - y\|_2 \\
        \leq & ~ 13344 \gamma_c \gamma^{-4} \beta^{-2} n^{1.5} \exp(40R^2) \| x - y \|_2 + 2\beta^{-2} n^{1.5}\exp(20R^2)\|x - y\|_2 \\
        \leq & ~ 13344 \gamma_c \gamma^{-4} \beta^{-2} n^{1.5} \exp(40R^2) \| x - y \|_2 + 2 \gamma^{-4} \beta^{-2} n^{1.5}\exp(20R^2)\|x - y\|_2 \\
        \leq & ~ 13344 \gamma_c \gamma^{-4} \beta^{-2} n^{1.5} \exp(40R^2) \| x - y \|_2 + 2 \gamma^{-4} \beta^{-2} n^{1.5}\exp(40R^2)\|x - y\|_2 \\
        \leq & ~ ( 13344 \gamma_c + 2 ) \gamma^{-4} \beta^{-2} n^{1.5} \exp(40R^2) \| x - y \|_2
    \end{align*}
    where the first equality follows from Lemma~\ref{lem:Hessian_L}, the second equality follows from Fact~\ref{fact:vec_norm_bound}, the third equality follows from Lemma~\ref{lem:H1_lipschitz}, the fourth equality follows from Lemma~\ref{lem:dls23_lipschitz_bound}, the fifth equality follows from Lemma~\ref{lem:B_lipschitz_bound}, the sixth equality follows from simple algebra, the seventh equality follows from $\gamma \in (0, 1)$, the eighth equality follows from $\exp(20R^2) \leq \exp(40R^2)$, the last equality follows from simple algebra.
\end{proof}

\subsection{Helpful Lemma}\label{subapp:lipschitz_helpful_lemma}

\begin{lemma}\label{lem:H1_lipschitz}
If the following condition holds
\begin{itemize}
    \item Let $Q(x), Q_1(x), Q_2(x), Q_3(x), Q_4(x), Q_5(x)$ be denoted as Definition~\ref{def:Q}
    \item Let $\gamma \in (0,1)$
    \item Let $\beta \in (0, 0.1)$
    \item Let $\gamma_c > 0$ denote a scalar
    \item Let $R \geq 4$
    \item Let $R_f := \beta^{-2} n^{1.5} \exp(3R^2)$
    \item $\ell(x) \geq \gamma$
    \item $Q(x) = 32 \gamma_c ( Q_1(x) - Q_2(x) - Q_3(x) + Q_4(x) ) - Q_5(x)$
    \item Let $H_1(x) := \frac{\d^2 \gamma_c \cdot \ell(x)^{-1}}{\d x^2}$
    \item $\| A \| \leq R$
\end{itemize}
Then, we have
\begin{itemize}
    \item Part 1.
    \begin{align*}
        \| Q(x) - Q(y) \| \leq 6672 \gamma^{-4} \beta^{-2} n^{1.5} \exp(20R^2) \| x - y \|_2
    \end{align*}
    \item Part 2.
    \begin{align*}
        \| H_1(x) - H_1(y) \| \leq 6672 \gamma^{-4} \beta^{-2} n^{1.5} \exp(40R^2) \| x - y \|_2
    \end{align*}
\end{itemize}
\end{lemma}

\begin{proof}
    \textbf{Proof of Part 1.}
    \begin{align}\label{eq:Q_lipschitz_tool1}
        \| Q(x) - Q(y) \|
        = & ~ \| ( 32 \gamma_c ( Q_1(x) - Q_2(x) - Q_3(x) + Q_4(x) ) - Q_5(x) ) \notag \\
        & ~ - ( 32 \gamma_c ( Q_1(y) - Q_2(y) - Q_3(y) + Q_4(y) ) - Q_5(y) ) \| \notag \\
        \leq & ~ 32 \gamma_c \| Q_1(x) - Q_1(y) \| + 16 \| Q_2(x) - Q_2(y) \| \notag \\
        & ~ + 32 \gamma_c \| Q_3(x) - Q_3(y) \| + 16 \| Q_4(x) - Q_4(y) \| + \| Q_5(x) - Q_5(y) \| ) \notag \\
        \leq & ~ 32 \gamma_c ( \| Q_1(x) - Q_1(y) \| + \| Q_2(x) - Q_2(y) \| \notag \\
        & ~ + \| Q_3(x) - Q_3(y) \| + \| Q_4(x) - Q_4(y) \| + \| Q_5(x) - Q_5(y) \| )
    \end{align}
    where the first equality follows from the Definition of $Q(x)$, the second equality follows from Fact~\ref{fact:vec_norm_bound}, the third equality follows from simple algebra.

    Then we combine Lemma~\ref{lem:lipschitz_Q1}, Lemma~\ref{lem:lipschitz_Q2}, Lemma~\ref{lem:lipschitz_Q3}, Lemma~\ref{lem:lipschitz_Q4}, we show that
    \begin{align}\label{eq:Q_lipschitz_tool2}
        & ~ \| Q_1(x) - Q_1(y) \| + \| Q_2(x) - Q_2(y) \|  + \| Q_3(x) - Q_3(y) \| + \| Q_4(x) - Q_4(y) \| \notag \\
        \leq & ~ 68 \gamma^{-4} R_f \| x - y \|_2 + 64 \gamma^{-4} R_f \| x - y \|_2 + 64 \gamma^{-4} R_f \| x - y \|_2 + 132 \gamma^{-4} R_f \| x - y \|_2 \notag \\
        = & ~ 328 \gamma^{-4} R_f \| x - y \|_2 \notag \\
        = & ~ 328 \gamma^{-4} \beta^{-2} n^{1.5} \exp(3R^2) \| x - y \|_2 \notag \\
        \leq & ~ 328 \gamma^{-4} \beta^{-2} n^{1.5} \exp(20R^2) \| x - y \|_2
    \end{align}
    where the second equality follows from simple algebra, the third equality follows from $R_f = \beta^{-2} n^{1.5} \exp(3R^2)$, the fourth equality follows from $\exp(3R^2) \leq \exp(20R^2)$.

    So we have
    \begin{align*}
        \| Q(x) - Q(y) \|
        \leq & ~ 32 \gamma_c ( \| Q_1(x) - Q_1(y) \| + \| Q_2(x) - Q_2(y) \| \\
        & ~ + \| Q_3(x) - Q_3(y) \| + \| Q_4(x) - Q_4(y) \| + \| Q_5(x) - Q_5(y) \| ) \\
        \leq & ~ 32 \gamma_c ( 328 \gamma^{-4} \beta^{-2} n^{1.5} \exp(20R^2) \| x - y \|_2 + \| Q_5(x) - Q_5(y) \| ) \\
        \leq & ~ 32 \gamma_c ( 328 \gamma^{-4} \beta^{-2} n^{1.5} \exp(20R^2) \| x - y \|_2 +  89\gamma^{-3} \beta^{-2} n^{1.5}\exp(20R^2)\|x - y\|_2) \\
        = & ~ 13344 \gamma_c \gamma^{-4} \beta^{-2} n^{1.5} \exp(20R^2) \| x - y \|_2
    \end{align*}
    where the first equality follows from Eq~\eqref{eq:Q_lipschitz_tool1}, the second equality follows from Eq~\eqref{eq:Q_lipschitz_tool2}, the third equality follows from Lemma~\ref{lem:lipschitz_Q5}, the fourth equality follows from simple algebra.

    \textbf{Proof of Part 2.}
    \begin{align*}
        \| H_1(x) - H_1(y) \|
        = & ~ A^\top \| Q(x) - Q(y) \| A \\
        \leq & ~ \| A \| \cdot \| Q(x) - Q(y) \| \cdot \| A \| \\
        \leq & ~ R^2 \cdot \| Q(x) - Q(y) \| \\
        \leq & ~ R^2 \cdot 13344 \gamma_c \gamma^{-4} \beta^{-2} n^{1.5} \exp(20R^2) \| x - y \|_2 \\
        \leq & ~ 13344 \gamma_c \gamma^{-4} \beta^{-2} n^{1.5} \exp(40R^2) \| x - y \|_2
    \end{align*}
    where the first equality follows from Definition~\ref{def:Q}, the second equality follows from simple algebra, the third equality follows from $\| A \| \leq R$, the fourth equality follows from Part 1 of Lemma~\ref{lem:H1_lipschitz}, the fifth equality follows from simple algebra.
\end{proof}

\begin{lemma}[Lemma 7.1 of \cite{dls23}]\label{lem:dls23_lipschitz_bound}
    If the following condition holds
    \begin{itemize}
        \item Let $\beta \in (0, 0.1)$
        \item Let $R \geq 4$
        \item Let $w \in \R^{n \times d}$
        \item Let $W = \diag(w)$
        \item Let $L_{reg}(x) = \| W Ax \|_2$
        \item Let $H_2(x) = \frac{\d^2 ( L_{reg}(x) + 0.5 \ell_1(x) )}{\d x^2}$
    \end{itemize}
    then we have 
    \begin{align*}
        \| H_2(x) - H_2(y) \| \leq \beta^{-2} n^{1.5}\exp(20R^2)\|x - y\|_2
    \end{align*}
\end{lemma}

\subsection{Definition of Matrix Functions \texorpdfstring{$Q_i(x)$}{}}\label{subapp:def_Q}
\begin{definition}\label{def:Q}
If the given conditions are satisfied
\begin{itemize}
    \item Let $\ell(x)$ be denoted as Definition~\ref{def:ell:formal} 
    \item Denote $f(x)$ as Definition~\ref{def:f:formal}
    \item Denote $c(x)$ as Definition~\ref{def:c:formal}
    \item Denote $B(x)$ as Definition~\ref{def:B:formal}
    \item Let $H_1(x) := \frac{\d^2 \gamma_c \cdot \ell_1(x)^{-1}}{\d x^2}$
\end{itemize}
We define $Q_1(x), Q_2(x), Q_3(x), Q_4(x), Q_5(x) \in \R^{n \times n}$ as follows
\begin{itemize}
    \item $Q_1(x) = \underbrace{ \ell(x)^{-3} }_{\mathrm{scalar}} \cdot \underbrace{ \langle f(x), c(x) \rangle^2 }_{ \mathrm{scalar} } \cdot \underbrace{ f(x) }_{n \times 1} \cdot \underbrace{ f(x)^\top }_{1 \times n}$
    \item $Q_2(x) = \underbrace{ \ell(x)^{-3} }_{ \mathrm{scalar} } \cdot \underbrace{ \langle f(x), c(x) \rangle }_{ \mathrm{scalar} } \cdot \underbrace{ ( f(x) \circ c(x) ) }_{ n \times 1 } \cdot \underbrace{ f(x)^\top }_{1 \times n}$
    \item $Q_3(x) = \underbrace{ \ell(x)^{-3} }_{ \mathrm{scalar} } \cdot \underbrace{ \langle f(x), c(x) \rangle }_{ \mathrm{scalar} } \cdot \underbrace{ f(x) }_{n \times 1} \cdot \underbrace{ ( f(x) \circ c(x) )^\top }_{1 \times n}$ 
    \item $Q_4(x) = \underbrace{ \ell(x)^{-3} }_{ \mathrm{scalar} } \cdot \underbrace{ (f(x) \circ c(x)) }_{n \times 1} \cdot \underbrace{ (f(x) \circ c(x))^\top }_{1 \times n}$
    \item $Q_5 (x) = \underbrace{ \ell(x)^{-2} }_{ \mathrm{scalar} } \cdot \underbrace{ B(x) }_{n \times n}$
\end{itemize}

We define $Q(x) \in \R^{n \times n}$ as follows
\begin{align*}
 Q(x) := 32 \gamma_c ( Q_1(x) - Q_2(x) - Q_3(x) + Q_4(x) ) - Q_5(x)
\end{align*}

Note that $H_1(x) = A^\top Q(x) A$.
\end{definition}

\subsection{Lipschitz Property for Matrix Function \texorpdfstring{$Q_1(x)$}{}}\label{subapp:lipschitz_Q1}

\begin{lemma}\label{lem:lipschitz_Q1}
If the given conditions are satisfied
\begin{itemize}
    \item Denote $Q_1(x)$ as Definition~\ref{def:Q}
    \item Let $\gamma \in (0,1)$
    \item Let $\beta \in (0, 0.1)$
    \item Let $R \geq 4$
    \item Let $R_f := \beta^{-2} n^{1.5} \exp(3R^2)$
    \item $\ell(x) \geq \gamma$
\end{itemize}
Then, we have
\begin{itemize}
    \item 
    \begin{align*}
        \| Q_1(x) - Q_1(y) \| \leq 68 \gamma^{-4} \cdot R_f \cdot \| x - y \|_2
    \end{align*}
\end{itemize}
\end{lemma}

\begin{proof}
    \begin{align*}
        \| Q_1(x) - Q_1(y) \| = & ~ \| \ell(x)^{-3} \cdot \langle f(x), c(x) \rangle^2 \cdot f(x) f(x)^\top - \ell(y)^{-3} \cdot \langle f(y), c(y) \rangle^2 \cdot f(y) f(y)^\top \| \\
        \leq & ~ | \ell(x)^{-3} \cdot \langle f(x), c(x) \rangle^2 | \cdot \| f(x) \|_2 \cdot \| f(x) - f(y) \|_2 \\
        & ~ + | \ell(x)^{-3} \cdot \langle f(x), c(x) \rangle^2 | \cdot \| f(x) - f(y) \|_2 \cdot \| f(y) \|_2 \\
        & ~ + \| f(x) \|_2 \cdot | \ell(x)^{-3} \cdot \langle f(x), c(x) \rangle^2 - \ell(y)^{-3} \cdot \langle f(y), c(y) \rangle^2 | \cdot \| f(x) \|_2
    \end{align*}
    where the first equality follows from Part 1 of Definition~\ref{def:Q}, the second equality follows from Part 3 of Lemma~\ref{lem:partializ_bound_tool}.

    We define
    \begin{align*}
        Q_{1, 1}(x) := & ~ | \ell(x)^{-3} \cdot \langle f(x), c(x) \rangle^2 | \cdot \| f(x) \|_2 \cdot \| f(x) - f(y) \|_2 \\
        Q_{1, 2}(x) := & ~ | \ell(x)^{-3} \cdot \langle f(x), c(x) \rangle^2 | \cdot \| f(x) - f(y) \|_2 \cdot \| f(y) \|_2 \\
        Q_{1, 3}(x) := & ~ \| f(y) \|_2 \cdot | \ell(x)^{-3} \cdot \langle f(x), c(x) \rangle^2 - \ell(y)^{-3} \cdot \langle f(y), c(y) \rangle^2 | \cdot \| f(y) \|_2
    \end{align*}
    where
    \begin{align*}
        \| Q_1(x) - Q_1(x) \| \leq Q_{1, 1}(x) + Q_{1, 2}(x) + Q_{1, 3}(x)
    \end{align*}
    So we can show that
    \begin{align}\label{eq:Q_1_1}
        Q_{1, 1}(x) = & ~ | \ell(x)^{-3} \cdot \langle f(x), c(x) \rangle^2 | \cdot \| f(x) \|_2 \cdot \| f(x) - f(y) \|_2 \notag \\
        \leq & ~ 4 \gamma^{-3} \cdot \| f(x) \|_2 \cdot \| f(x) - f(y) \|_2 \notag \\
        \leq & ~ 4 \gamma^{-3} \cdot \| f(x) - f(y) \|_2 \notag \\
        \leq & ~ 4 \gamma^{-3} \cdot R_f \cdot \| x - y \|_2 
    \end{align}
    where the first follows from the Definition of $Q_{1, 1}(x)$, the second equality follows from Part 3 of Lemma~\ref{lem:func_bounds}, the third equality follows from $\|f(x)\|_2 \leq 1$, the fourth equality follows from Part 1 of Lemma~\ref{lem:vec_lipschitz_tool}.
    \begin{align}\label{eq:Q_1_2}
        Q_{2, 1}(x) = & ~ | \ell(x)^{-3} \cdot \langle f(x), c(x) \rangle^2 | \cdot \| f(x) - f(y) \|_2 \cdot \| f(y) \|_2 \notag \\
        \leq & ~ 4 \gamma^{-3} \cdot \| f(x) - f(y) \|_2 \cdot \| f(y) \|_2 \notag \\
        \leq & ~ 4 \gamma^{-3} \cdot \| f(x) - f(y) \|_2 \notag \\
        \leq & ~ 4 \gamma^{-3} \cdot R_f \cdot \| x - y \|_2 
    \end{align}
    where the first follows from the Definition of $Q_{2, 1}(x)$, the second equality follows from Part 3 of Lemma~\ref{lem:func_bounds}, the third equality follows from $\|f(x)\|_2 \leq 1$, the fourth equality follows from Part 1 of Lemma~\ref{lem:vec_lipschitz_tool}.
    \begin{align}\label{eq:Q_1_3}
         Q_{1, 3}(x) = & ~ \| f(y) \|_2 \cdot | \ell(x)^{-3} \cdot \langle f(x), c(x) \rangle^2 - \ell(y)^{-3} \cdot \langle f(y), c(y) \rangle^2 | \cdot \| f(y) \|_2 \notag \\ 
         \leq & ~ | \ell(x)^{-3} \cdot \langle f(x), c(x) \rangle^2 - \ell(y)^{-3} \cdot \langle f(y), c(y) \rangle^2 \notag \\ 
         \leq & ~ 60 \gamma^{-4} \cdot R_f \cdot \| x - y \|_2
    \end{align}
    where the first equality follows from the Definition of $Q_{1, 3}(x)$, the second equality follows from $\|f(x)\|_2 \leq 1$, the third equality follows from Part 2 of Lemma~\ref{lem:scalar_lipschitz_tool_ell_f_c}.

    So we have
    \begin{align*}
        \| Q_1(x) - Q_1(x) \| \leq & ~ Q_{1, 1}(x) + Q_{1, 2}(x) + Q_{1, 3}(x) \\
        \leq & ~ 4 \gamma^{-3} \cdot R_f \cdot \| x - y \|_2 + Q_{1, 2}(x) + Q_{1, 3}(x) \\
        \leq & ~ 4 \gamma^{-3} \cdot R_f \cdot \| x - y \|_2 + 4 \gamma^{-3} \cdot R_f \cdot \| x - y \|_2 + Q_{1, 3}(x) \\
        \leq & ~ 4 \gamma^{-3} \cdot R_f \cdot \| x - y \|_2 + 4 \gamma^{-3} \cdot R_f \cdot \| x - y \|_2 + 60 \gamma^{-4} \cdot R_f \cdot \| x - y \|_2 \\
        \leq & ~ 68 \gamma^{-4} R_f \cdot \| x - y \|_2
    \end{align*}
    where the second equality follows from Eq.~\eqref{eq:Q_1_1}, the third equality follows from Eq.~\eqref{eq:Q_1_2}, the fourth equality follows from Eq.~\eqref{eq:Q_1_3}, the fifth equality follows from simple algebra.

\end{proof}

\subsection{Lipschitz Property for Matrix Function \texorpdfstring{$Q_2(x)$}{}}\label{subapp:lipschitz_Q2}
\begin{lemma}\label{lem:lipschitz_Q2}
If the given conditions are satisfied
\begin{itemize}
    \item Denote $Q_2(x)$ as Definition~\ref{def:Q}
    \item Let $\gamma \in (0,1)$
    \item Let $\beta \in (0, 0.1)$
    \item Let $R \geq 4$
    \item Let $R_f := \beta^{-2} n^{1.5} \exp(3R^2)$
    \item $\ell(x) \geq \gamma$
\end{itemize}
We have
\begin{itemize}
    \item 
    \begin{align*}
        \| Q_2(x) - Q_2(y) \| \leq 64 \gamma^{-4} \cdot R_f \| x - y \|_2
    \end{align*} 
\end{itemize}
\end{lemma}

\begin{proof}

    \begin{align*}
        \| Q_2(x) - Q_2(y) \| 
        = & ~ \| \ell(x)^{-3} \langle f(x), c(x) \rangle ( f(x) \circ c(x) ) f(x)^\top - \ell(y)^{-3} \langle f(y), c(y) \rangle ( f(y) \circ c(y) ) f(y)^\top \| \\
        \leq & ~ | \ell(x)^{-3} \cdot \langle f(x), c(x) \rangle | \cdot \| f(x) \circ c(x) \|_2 \cdot \| f(x) - f(y) \|_2 \\
        & ~ + | \ell(x)^{-3} \cdot \langle f(x), c(x) \rangle | \cdot \| f(x) \circ c(x) - f(y) \circ c(y) \|_2 \cdot \| f(y) \|_2 \\
        & ~ + \| f(y) \circ c(y) \|_2 \cdot | \ell(x)^{-3} \cdot \langle f(x), c(x) \rangle - \ell(y)^{-3} \cdot \langle f(y), c(y) \rangle | \cdot \| f(y) \|_2
    \end{align*}
    where the first equality follows from Part 2 of Definition~\ref{def:Q}, the second equality follows from Part 3 of Lemma~\ref{lem:partializ_bound_tool}.

    We define
    \begin{align*}
        Q_{2, 1}(x) := & ~ | \ell(x)^{-3} \cdot \langle f(x), c(x) \rangle | \cdot \| f(x) \circ c(x) \|_2 \cdot \| f(x) - f(y) \|_2 \\
        Q_{2, 2}(x) := & ~ | \ell(x)^{-3} \cdot \langle f(x), c(x) \rangle | \cdot \| f(x) \circ c(x) - f(y) \circ c(y) \|_2 \cdot \| f(y) \|_2 \\
        Q_{2, 3}(x) := & ~ \| f(y) \circ c(y) \|_2 \cdot | \ell(x)^{-3} \cdot \langle f(x), c(x) \rangle - \ell(y)^{-3} \cdot \langle f(y), c(y) \rangle | \cdot \| f(y) \|_2
    \end{align*}
    where
    \begin{align*}
        \| Q_2(x) - Q_2(y) \| \leq Q_{2, 1}(x) + Q_{2, 2}(x) + Q_{2, 3}(x)
    \end{align*}
    So we can show that
    \begin{align}\label{eq:Q_2_1}
        Q_{2, 1}(x) = & ~ | \ell(x)^{-3} \cdot \langle f(x), c(x) \rangle | \cdot \| f(x) \circ c(x) \|_2 \cdot \| f(x) - f(y) \|_2 \notag \\
        \leq & ~ 2 \gamma^{-3} \cdot \| f(x) \circ c(x) \|_2 \cdot \| f(x) - f(y) \|_2 \notag \\
        \leq & ~ 2 \gamma^{-3} \cdot 2 \cdot \| f(x) - f(y) \|_2 \notag \\
        \leq & ~ 2 \gamma^{-3} \cdot 2 \cdot R_f \cdot \| x - y \|_2 \notag \\
        = & ~ 4 \gamma^{-3} \cdot R_f \cdot \| x - y \|_2
    \end{align}
    where the first equality follows from the Definition of $Q_{2, 1}(x)$, the second equality follows from Part 3 of Lemma~\ref{lem:func_bounds}, the third equality follows from Part 2 of Lemma~\ref{lem:vec_func_norm_bound}, the fourth equality follows from Part 1 of Lemma~\ref{lem:vec_lipschitz_tool}, the last equality follows from simple algebra.
    \begin{align}\label{eq:Q_2_2}
        Q_{2, 2}(x) = & ~ | \ell(x)^{-3} \cdot \langle f(x), c(x) \rangle | \cdot \| f(x) \circ c(x) - f(y) \circ c(y) \|_2 \cdot \| f(y) \|_2 \notag \\
        \leq & ~ 2 \gamma^{-3} \cdot \| f(x) \circ c(x) - f(y) \circ c(y) \|_2 \cdot \| f(y) \|_2 \notag \\
        \leq & ~ 2 \gamma^{-3} \cdot \| f(x) \circ c(x) - f(y) \circ c(y) \|_2 \notag \\
        \leq & ~ 2 \gamma^{-3} \cdot 3 \cdot R_f \cdot \| x - y \|_2 \notag \\
        = & ~ 6 \gamma^{-3} \cdot R_f \cdot \| x - y \|_2
    \end{align}
    where the first equality follows from the Definition of $Q_{2, 1}(x)$, the second equality follows from Part 3 of Lemma~\ref{lem:func_bounds}, the third equality follows from $\|f(x)\|_2 \leq 1$, the fourth equality follows from Part 3 of Lemma~\ref{lem:vec_lipschitz_tool}, the last equality follows from simple algebra.
    \begin{align}\label{eq:Q_2_3}
        Q_{2, 3}(x) = & ~ \| f(y) \circ c(y) \|_2 \cdot | \ell(x)^{-3} \cdot \langle f(x), c(x) \rangle - \ell(y)^{-3} \cdot \langle f(y), c(y) \rangle | \cdot \| f(y) \|_2 \notag \\
        \leq & ~ \| f(y) \circ c(y) \|_2 \cdot | \ell(x)^{-3} \cdot \langle f(x), c(x) \rangle - \ell(y)^{-3} \cdot \langle f(y), c(y) \rangle | \notag \\
        \leq & ~ 2 \cdot | \ell(x)^{-3} \cdot \langle f(x), c(x) \rangle - \ell(y)^{-3} \cdot \langle f(y), c(y) \rangle | \notag \\
        \leq & ~ 2 \cdot 27 \gamma^{-4} \cdot R_f \| x - y \|_2 \notag \\
        \leq & ~ 54\gamma^{-4} \cdot R_f \| x - y \|_2
    \end{align}
    where the first equality follows from the Definition of $Q_{2, 3}(x)$, the second equality follows from $\|f(x)\|_2 \leq 1$, the third equality follows from Part 2 of Lemma~\ref{lem:vec_func_norm_bound}, the fourth equality follows from Part 1 of Lemma~\ref{lem:scalar_lipschitz_tool_ell_f_c}, the last equality follows from simple algebra.

    So we have
    \begin{align*}
        \| Q_2(x) - Q_2(y) \| 
        \leq & ~ Q_{2, 1}(x) + Q_{2, 2}(x) + Q_{2, 3}(x) \\
        \leq & ~ 4 \gamma^{-3} \cdot R_f \cdot \| x - y \|_2 + Q_{2, 2}(x) + Q_{2, 3}(x) \\
        \leq & ~ 4 \gamma^{-3} \cdot R_f \cdot \| x - y \|_2 + 6 \gamma^{-3} \cdot R_f \cdot \| x - y \|_2 + Q_{2, 3}(x) \\
        \leq & ~ 4 \gamma^{-3} \cdot R_f \cdot \| x - y \|_2 + 6 \gamma^{-3} \cdot R_f \cdot \| x - y \|_2 + 54 \gamma^{-4} \cdot R_f \| x - y \|_2 \\
        \leq & ~ 64 \gamma^{-4} \cdot R_f \| x - y \|_2
    \end{align*}
    where the second equality follows from Eq.~\eqref{eq:Q_2_1}, the third equality follows from Eq.~\eqref{eq:Q_2_2}, the fourth equality follows from Eq.~\eqref{eq:Q_2_3}, the fifth equality follows from simple algebra.

    \end{proof}

\subsection{Lipschitz Property for Matrix Function \texorpdfstring{$Q_3(x)$}{}}\label{subapp:lipschitz_Q3}

\begin{lemma}\label{lem:lipschitz_Q3}
If the given conditions are satisfied
\begin{itemize}
    \item Denote $Q_3(x)$ as Definition~\ref{def:Q}
    \item Let $\gamma \in (0,1)$
    \item Let $\beta \in (0, 0.1)$
    \item Let $R \geq 4$
    \item Let $R_f := \beta^{-2} n^{1.5} \exp(3R^2)$
    \item $\ell(x) \geq \gamma$
\end{itemize}
We have
\begin{itemize}
    \item 
    \begin{align*}
        \| Q_3(x) - Q_3(y) \| \leq 64 \gamma^{-4} \cdot R_f \| x - y \|_2
    \end{align*}
\end{itemize}
\end{lemma}

    \begin{proof}

    \textbf{Proof of Part 3.}
    \begin{align*}
        \| Q_3(x) - Q_3(y) \|
        = & ~ \| \ell(x)^{-3} \langle f(x), c(x) \rangle f(x) ( f(x) \circ c(x) )^\top - \ell(y)^{-3} \langle f(y), c(y) \rangle f(y) ( f(y) \circ c(y) )^\top \| \\
        \leq & ~ | \ell(x)^{-3} \cdot \langle f(x), c(x) \rangle | \cdot \| f(x) \|_2 \cdot \| f(x) \circ c(x) - f(y) \circ c(y) \|_2 \\
        & ~ + | \ell(x)^{-3} \cdot \langle f(x), c(x) \rangle | \cdot \| f(x) - f(y) \|_2 \cdot \| f(y) \circ c(y) \|_2 \\
        & ~ + \| f(y) \|_2 \cdot | \ell(x)^{-3} \cdot \langle f(x), c(x) \rangle - \ell(y)^{-3} \cdot \langle f(y), c(y) \rangle | \cdot \| f(y) \circ c(y) \|_2 
    \end{align*}
    where the first equality follows from the Definition of $Q_3(x)$, the second equality follows from Part 3 of Lemma~\ref{lem:partializ_bound_tool}.

    We define
    \begin{align*}
        Q_{3, 1}(x) := & ~ | \ell(x)^{-3} \cdot \langle f(x), c(x) \rangle | \cdot \| f(x) \|_2 \cdot \| f(x) \circ c(x) - f(y) \circ c(y) \|_2 \\
        Q_{3, 2}(x) := & ~ | \ell(x)^{-3} \cdot \langle f(x), c(x) \rangle | \cdot \| f(x) - f(y) \|_2 \cdot \| f(y) \circ c(y) \|_2 \\
        Q_{3, 3}(x) := & ~ \| f(y) \|_2 \cdot | \ell(x)^{-3} \cdot \langle f(x), c(x) \rangle - \ell(y)^{-3} \cdot \langle f(y), c(y) \rangle | \cdot \| f(y) \circ c(y) \|_2 
    \end{align*}
    where
    \begin{align*}
        \| Q_3(x) - Q_3(y) \| \leq Q_{3, 1}(x) + Q_{3, 2}(x) + Q_{3, 3}(x)
    \end{align*}
    So we can show that
    \begin{align}\label{eq:Q_3_1}
        Q_{3, 1}(x)
        = & ~ | \ell(x)^{-3} \cdot \langle f(x), c(x) \rangle | \cdot \| f(x) \|_2 \cdot \| f(x) \circ c(x) - f(y) \circ c(y) \|_2 \notag \\
        \leq & ~ 2 \gamma^{-3} \cdot \| f(x) \|_2 \cdot \| f(x) \circ c(x) - f(y) \circ c(y) \|_2 \notag \\
        \leq & ~ 2 \gamma^{-3} \cdot \| f(x) \circ c(x) - f(y) \circ c(y) \|_2 \notag \\
        \leq & ~ 2 \gamma^{-3} \cdot 3 R_f \| x - y \|_2 \notag \\
        = & ~ 6 \gamma^{-3} \cdot R_f \| x - y \|_2
    \end{align}
    where the first equality follows from the Definition of $Q_{3, 1}(x)$, the second equality follows from Part 3 of Lemma~\ref{lem:func_bounds}, the third equality follows from $\|f(x)\|_2 \leq 1$, the fourth equality follows from Part 3 of Lemma~\ref{lem:vec_lipschitz_tool}, the last equality follows from simple algebra.
    \begin{align}\label{eq:Q_3_2}
        Q_{3, 2}(x) 
        = & ~ | \ell(x)^{-3} \cdot \langle f(x), c(x) \rangle | \cdot \| f(x) - f(y) \|_2 \cdot \| f(y) \circ c(y) \|_2 \notag \\
        \leq & ~ 2 \gamma^{-3} \cdot \| f(x) - f(y) \|_2 \cdot \| f(y) \circ c(y) \|_2 \notag \\
        \leq & ~ 2 \gamma^{-3} \cdot \| f(x) - f(y) \|_2 \cdot 2 \notag \\
        \leq & ~ 2 \gamma^{-3} \cdot R_f \cdot \| x - y \|_2 \cdot 2 \notag \\
        = & ~ 4 \gamma^{-3} \cdot R_f \cdot \| x - y \|_2
    \end{align}
    where the first equality follows from the Definition of $Q_{3, 2}(x)$, the second equality follows from Part 3 of Lemma~\ref{lem:func_bounds}, the third equality follows from Part 2 of Lemma~\ref{lem:vec_func_norm_bound}, the fourth equality follows from Part 1 of Lemma~\ref{lem:vec_lipschitz_tool}, the last equality follows from simple algebra.
    \begin{align}\label{eq:Q_3_3}
        Q_{3, 3}(x)
        = & ~ \| f(y) \|_2 \cdot | \ell(x)^{-3} \cdot \langle f(x), c(x) \rangle - \ell(y)^{-3} \cdot \langle f(y), c(y) \rangle | \cdot \| f(y) \circ c(y) \|_2 \notag \\
        \leq & ~ | \ell(x)^{-3} \cdot \langle f(x), c(x) \rangle - \ell(y)^{-3} \cdot \langle f(y), c(y) \rangle | \cdot \| f(y) \circ c(y) \|_2 \notag \\
        \leq & ~ | \ell(x)^{-3} \cdot \langle f(x), c(x) \rangle - \ell(y)^{-3} \cdot \langle f(y), c(y) \rangle | \cdot 2 \notag \\
        \leq & ~ 27 \gamma^{-4} \cdot R_f \| x - y \|_2 \cdot 2 \notag \\
        = & ~ 54 \gamma^{-4} \cdot R_f \| x - y \|_2
    \end{align}
    where the first equality follows from the Definition of $Q_{3, 3}(x)$, the second equality follows from $\|f(x)\|_2 \leq 1$, the third equality follows from Part 2 of Lemma~\ref{lem:vec_func_norm_bound}, the fourth equality follows from Part 1 of Lemma~\ref{lem:scalar_lipschitz_tool_ell_f_c}, the last equality follows from simple algebra.

    So we have
    \begin{align*}
        \| Q_3(x) - Q_3(y) \| 
        \leq & ~ Q_{3, 1}(x) + Q_{3, 2}(x) + Q_{3, 3}(x) \\
        \leq & ~ 6 \gamma^{-3} \cdot R_f \cdot \| x - y \|_2 + Q_{3, 2}(x) + Q_{3, 3}(x) \\
        \leq & ~ 6 \gamma^{-3} \cdot R_f \cdot \| x - y \|_2 + 4 \gamma^{-3} \cdot R_f \cdot \| x - y \|_2 + Q_{3, 3}(x) \\
        \leq & ~ 6 \gamma^{-3} \cdot R_f \cdot \| x - y \|_2 + 4 \gamma^{-3} \cdot R_f \cdot \| x - y \|_2 + 54 \gamma^{-4} \cdot R_f \| x - y \|_2 \\
        \leq & ~ 64 \gamma^{-4} \cdot R_f \| x - y \|_2
    \end{align*}
    where the first equality follows from Eq.~\eqref{eq:Q_3_1}, the second equality follows from Eq.~\eqref{eq:Q_3_2}, the third equality follows from Eq.~\eqref{eq:Q_3_3}, the fourth equality follows from simple algebra.

\end{proof}

\subsection{Lipschitz Property for Matrix Function \texorpdfstring{$Q_4(x)$}{}}\label{subapp:lipschitz_Q4}

\begin{lemma}\label{lem:lipschitz_Q4}
If the given conditions are satisfied
\begin{itemize}
    \item Denote $Q_4(x)$ as Definition~\ref{def:Q}
    \item Let $\gamma \in (0,1)$
    \item Let $\beta \in (0, 0.1)$
    \item Let $R \geq 4$
    \item Let $R_f := \beta^{-2} n^{1.5} \exp(3R^2)$
    \item $\ell(x) \geq \gamma$
\end{itemize}
We have
\begin{itemize} 
    \item 
    \begin{align*}
        \| Q_4(x) - Q_4(y) \| \leq 132 \gamma^{-3} \cdot R_f \| x - y \|_2
    \end{align*} 
\end{itemize}
\end{lemma}

\begin{proof}

    \begin{align*}
        \| Q_4(x) - Q_4(y) \| 
        = & ~ \| \ell(x)^{-3} \langle f(x), c(x) \rangle ( f(x) \circ c(x) ) ( f(x) \circ c(x) )^\top \\ 
        & ~ - \ell(y)^{-2} \langle f(y), c(y) \rangle ( f(y) \circ c(y) ) ( f(y) \circ c(y) )^\top \| \\
        \leq & ~ | \ell(x)^{-3} \cdot \langle f(x), c(x) \rangle | \cdot \| f(x) \circ c(x) \|_2 \cdot \| f(x) \circ c(x) - f(y) \circ c(y) \|_2 \\
        & ~ + | \ell(x)^{-3} \cdot \langle f(x), c(x) \rangle | \cdot \| f(x) \circ c(x) - f(y) \circ c(y) \|_2 \cdot \| f(y) \circ c(y) \|_2 \\
        & ~ + \| f(y) \circ c(y) \|_2 \cdot | \ell(x)^{-3} \cdot \langle f(x), c(x) \rangle - \ell(y)^{-3} \cdot \langle f(y), c(y) \rangle | \cdot \| f(y) \circ c(y) \|_2
    \end{align*}
    where the first equality follows from the Definition~\ref{def:Q}, the second equality follows from Part 3 of Lemma~\ref{lem:partializ_bound_tool}.

    We define
    \begin{align*}
        Q_{4, 1}(x) := & ~ | \ell(x)^{-3} \cdot \langle f(x), c(x) \rangle | \cdot \| f(x) \circ c(x)\|_2 \cdot \| f(x) \circ c(x) - f(y) \circ c(y) \|_2 \\
        Q_{4, 2}(x) := & ~ | \ell(x)^{-3} \cdot \langle f(x), c(x) \rangle | \cdot \| f(x) \circ c(x) - f(y) \circ c(y) \|_2 \cdot \| f(y) \circ c(y) \|_2 \\
        Q_{4, 3}(x) := & ~ \| f(y) \circ c(y) \|_2 \cdot | \ell(x)^{-3} \cdot \langle f(x), c(x) \rangle - \ell(y)^{-3} \cdot \langle f(y), c(y) \rangle | \cdot \| f(y) \circ c(y) \|_2
    \end{align*}
    where
    \begin{align*}
        \| Q_4(x) - Q_4(y) \| \leq Q_{4, 1}(x) + Q_{4, 2}(x) + Q_{4, 3}(x)
    \end{align*}
    So we can show that
    \begin{align}\label{eq:Q_4_1}
        Q_{4, 1}(x) 
        = & ~ | \ell(x)^{-3} \cdot \langle f(x), c(x) \rangle | \cdot \| f(x) \circ c(x)\|_2 \cdot \| f(x) \circ c(x) - f(y) \circ c(y) \|_2 \notag \\
        \leq & ~ 2\gamma^{-3} \cdot \| f(x) \circ c(x)\|_2 \cdot \| f(x) \circ c(x) - f(y) \circ c(y) \|_2 \notag \\
        \leq & ~ 2\gamma^{-3} \cdot 2 \cdot \| f(x) \circ c(x) - f(y) \circ c(y) \|_2 \notag \\
        \leq & ~ 2\gamma^{-3} \cdot 2 \cdot 3 R_f \cdot \| x - y \|_2 \notag \\
        = & ~ 12\gamma^{-3} \cdot R_f \cdot \| x - y \|_2 
    \end{align}
    where the first equality follows from the Definition of $Q_{4, 1}(x)$, the second equality follows from Part 3 of Lemma~\ref{lem:func_bounds}, the third equality follows from Part 2 of Lemma~\ref{lem:vec_func_norm_bound}, the fourth equality follows from Part 3 of Lemma~\ref{lem:vec_lipschitz_tool}, the last equality follows from simple algebra.
    \begin{align}\label{eq:Q_4_2}
        Q_{4, 2}(x) 
        = & ~ | \ell(x)^{-3} \cdot \langle f(x), c(x) \rangle | \cdot \| f(x) \circ c(x) - f(y) \circ c(y) \|_2 \cdot \| f(y) \circ c(y) \|_2 \notag \\
        \leq & ~ 2\gamma^{-3} \cdot \| f(x) \circ c(x) - f(y) \circ c(y) \|_2 \cdot \| f(y) \circ c(y) \|_2 \notag \\
        \leq & ~ 2\gamma^{-3} \cdot \| f(x) \circ c(x) - f(y) \circ c(y) \|_2 \cdot 2 \notag \\
        \leq & ~ 2\gamma^{-3} \cdot 3 R_f \cdot \| x - y \|_2 \cdot 2 \notag \\
        = & ~ 12\gamma^{-3} \cdot R_f \cdot \| x - y \|_2 
    \end{align}
    where the first equality follows from the Definition of $Q_{4, 2}(x)$, the second equality follows from Part 3 of Lemma~\ref{lem:func_bounds}, the third equality follows from Part 2 of Lemma~\ref{lem:vec_func_norm_bound}, the fourth equality follows from Part 3 of Lemma~\ref{lem:vec_lipschitz_tool}, the last equality follows from simple algebra.
    \begin{align}\label{eq:Q_4_3}
        Q_{4, 3}(x) 
        = & ~ \| f(y) \circ c(y) \|_2 \cdot | \ell(x)^{-3} \cdot \langle f(x), c(x) \rangle - \ell(y)^{-3} \cdot \langle f(y), c(y) \rangle | \cdot \| f(y) \circ c(y) \|_2 \notag \\
        \leq & ~ 2 \cdot | \ell(x)^{-3} \cdot \langle f(x), c(x) \rangle - \ell(y)^{-3} \cdot \langle f(y), c(y) \rangle | \cdot 2 \notag \\
        \leq & ~ 2 \cdot 27 \gamma^{-4} \cdot R_f \| x - y \|_2 \cdot 2 \notag \\
        = & ~ 108 \gamma^{-4} \cdot R_f \| x - y \|_2
    \end{align}
    where the first equality follows from the Definition of $Q_{4, 3}(x)$, the second equality follows from Part 2 of Lemma~\ref{lem:vec_func_norm_bound}, the third equality follows Part 1 of Lemma~\ref{lem:scalar_lipschitz_tool_ell_f_c}, the last equality follows from simple algebra.

    So we have
    \begin{align*}
        \| Q_4(x) - Q_4(y) \| 
        \leq & ~ Q_{4, 1}(x) + Q_{4, 2}(x) + Q_{4, 3}(x) \\
        \leq & ~ 12 \gamma^{-3} \cdot R_f \cdot \| x - y \|_2 + Q_{3, 2}(x) + Q_{3, 3}(x) \\
        \leq & ~ 12 \gamma^{-3} \cdot R_f \cdot \| x - y \|_2 + 12 \gamma^{-2} \cdot R_f \cdot \| x - y \|_2 + Q_{3, 3}(x) \\
        \leq & ~ 12 \gamma^{-3} \cdot R_f \cdot \| x - y \|_2 + 12 \gamma^{-3} \cdot R_f \cdot \| x - y \|_2 + 108 \gamma^{-4}  \cdot R_f \| x - y \|_2 \\
        \leq & ~ 132 \gamma^{-4} \cdot R_f \| x - y \|_2
    \end{align*}
    where the second equality follows from Eq.~\eqref{eq:Q_4_1}, the third equality follows from Eq.~\eqref{eq:Q_4_2}, the fourth equality follows from Eq.~\eqref{eq:Q_4_3}, the fifth equality follows from simple algebra.

\end{proof}

\subsection{Lipschitz Property for Matrix Function \texorpdfstring{$Q_5(x)$}{}}\label{subapp:lipschitz_Q5}

\begin{lemma}\label{lem:lipschitz_Q5}
If the given conditions are satisfied
\begin{itemize}
    \item Denote $Q_5(x)$ as Definition~\ref{def:Q}
    \item Let $\gamma \in (0,1)$
    \item Let $\beta \in (0, 0.1)$
    \item Let $R \geq 4$
    \item Let $R_f := \beta^{-2} n^{1.5} \exp(3R^2)$
    \item $\ell(x) \geq \gamma$
\end{itemize}
We have
\begin{itemize}
    \item 
    \begin{align*}
        \| Q_5(x) - Q_5(y) \| \leq 89\gamma^{-3} \cdot \beta^{-2} n^{1.5}\exp(20R^2)\|x - y\|_2
    \end{align*}
\end{itemize}
\end{lemma}

\begin{proof}
    We have
    \begin{align*}
        \| Q_5(x) - Q_5(y) \|
        = & ~ \| \ell(x)^{-2} B(x) - \ell(y)^{-2} B(y) \| \\
        \leq & ~ \| \ell(x)^{-2} B(x) - \ell(x)^{-2} B(y) + \ell(x)^{-2} B(y) - \ell(y)^{-2} B(y) \| \\
        \leq & ~ \| \ell(x)^{-2} B(x) - \ell(x)^{-2} B(y)\| + \| \ell(x)^{-2} B(y) - \ell(y)^{-2} B(y) \| \\
        = & ~ | \ell(x)^{-2} | \cdot \| B(x) - B(y)\| + | \ell(x)^{-2} - \ell(y)^{-2} | \cdot \| B(y) \| \\
        \leq & ~ \gamma^{-2} \cdot \| B(x) - B(y)\| + | \ell(x)^{-2} - \ell(y)^{-2} | \cdot \| B(y) \| \\
        \leq & ~ \gamma^{-2} \cdot \| B(x) - B(y)\| + | \ell(x)^{-2} - \ell(y)^{-2} | \cdot 11 \\
        \leq & ~ \gamma^{-2} \cdot \| B(x) - B(y)\| + 8 \gamma^{-3} \cdot R_f \cdot \| x - y \|_2 \cdot 11 \\
        \leq & ~ \gamma^{-2} \cdot \beta^{-2} n^{1.5}\exp(20R^2)\|x - y\|_2 +  8 \gamma^{-3} \cdot R_f \cdot \| x - y \|_2 \cdot 11 \\
        \leq & ~ \gamma^{-3} \cdot \beta^{-2} n^{1.5}\exp(20R^2)\|x - y\|_2 +  8 \gamma^{-3} \cdot R_f \cdot \| x - y \|_2 \cdot 11 \\
        \leq & ~ \gamma^{-3} \cdot \beta^{-2} n^{1.5}\exp(20R^2)\|x - y\|_2 +  8 \gamma^{-3} \cdot \beta^{-2} n^{1.5}\exp(3R^2) \cdot \| x - y \|_2 \cdot 11 \\
        \leq & ~ \gamma^{-3} \cdot \beta^{-2} n^{1.5}\exp(20R^2)\|x - y\|_2 +  8 \gamma^{-3} \cdot \beta^{-2} n^{1.5}\exp(20R^2) \cdot \| x - y \|_2 \cdot 11 \\
        \leq & ~ 89\gamma^{-3} \cdot \beta^{-2} n^{1.5}\exp(20R^2)\|x - y\|_2 \\
        \leq & ~ 89\gamma^{-4} \cdot \beta^{-2} n^{1.5}\exp(20R^2)\|x - y\|_2
    \end{align*}
    where the first equality follows from Definition~\ref{def:Q}, the second, third, fourth equalities follow from Fact~\ref{fact:mat_norm_bound}, the fifth equality follows from $\ell(x) \geq \gamma$, the sixth equality follows from Part 1 of Lemma~\ref{lem:mat_func_norm_bounds}, the seventh equality follows from Part 2 fo Lemma~\ref{lem:scalar_lipschitz_tool_ell}, the eighth equality follows from Lemma~\ref{lem:B_lipschitz_bound}, the ninth equality follows from $\gamma \in (0, 1)$, the tenth equality follows from $R_f = \beta^{-2} n^{1.5} \exp(3R^2)$, the 1first equality follows from $\exp(3R^2) \leq \exp(20R^2)$, the 1second equality follows from simple algebra, the last equality follows from $\gamma \in (0, 1)$.
\end{proof}

\begin{lemma}\label{lem:B_lipschitz_bound}
    If the given conditions are satisfied
    \begin{itemize}
        \item Let $A \in \R^{n \times d}$
        \item Let $\beta \in (0, 0.1)$
        \item Let $R \geq 4$
        \item Let $w \in \R^{n \times d}$
        \item Let $W = \diag(w)$
        \item Let $L_{reg}(x) = \| W Ax \|_2$
        \item Let $H_2(x) = \frac{\d^2 ( L_{reg}(x) + \ell(x) )}{\d x^2}$
    \end{itemize}
    then we have
    \begin{align*}
        \| B(x) - B(y) \| \leq \beta^{-2} n^{1.5}\exp(20R^2)\|x - y\|_2
    \end{align*}

    \begin{proof}
        We have
        \begin{align*}
            \| H_2(x) - H_2(y) \| 
            = & ~ \| \frac{\d^2 ( L_{reg}(x) + \ell(x) )}{\d x^2} - \frac{\d^2 ( L_{reg}(y) + \ell(y) )}{\d y^2} \| \\
            = & ~ \| \frac{\d^2 L_{reg}(x)}{\d x^2} + \frac{\d^2 \ell(x) }{\d x^2} - \frac{\d^2 L_{reg}(y)}{\d y^2} - \frac{\d^2 \ell(y) }{\d y^2} \| \\
        \end{align*}
        where the first equality follows from the Definition of $H(x)$, the second equality follows from simple differential rule.

        So we have
        \begin{align*}
            \| B(x) - B(y) \|
            \leq & ~ A^\top \cdot \| B(x) - B(y) \| \cdot A \\
            = & ~ \| \frac{\d^2 \ell(x) }{\d x^2} - \frac{\d^2 \ell(y) }{\d y^2} \| \\
            \leq & ~ \| \frac{\d^2 L_{reg}(x)}{\d x^2} + \frac{\d^2 \ell(x) }{\d x^2} - \frac{\d^2 L_{reg}(y)}{\d y^2} - \frac{\d^2 \ell(y) }{\d y^2} \| \\
            = & ~ \| H(x) - H(y) \| \\
            \leq & ~ \beta^{-2} n^{1.5}\exp(20R^2)\|x - y\|_2
        \end{align*}
        where the first equality follows from simple algebra, the second equality follows from Lemma~\ref{lem:rewrite_B}, the third equality follows from , the fourth equality follows from the Definition of $H(x)$, the fifth equality follows from Lemma~\ref{lem:dls23_lipschitz_bound}.
    \end{proof}
\end{lemma}

\section{Lipschitz Tools}\label{app:lipschitz_tool}

In this appendix, we provide a set of tools that can assist in the computation of the Lipschitz property. In Appendix~\ref{subapp:tool_sclar}, we provide Lipschitz tool of some scalar functions. In Appendix~\ref{subapp:tool_vector}, we provide Lipschitz tool of some vector functions.

\subsection{Lipschitz Tool: Scalar Function}\label{subapp:tool_sclar}

\begin{lemma}\label{lem:scalar_lipschitz_tool_ell}
    If the given conditions are satisfied
    \begin{itemize}
        \item Let $A \in \R^{n \times d}$
        \item Let $\beta \in (0, 0.1)$
        \item Let $R \geq 4$
        \item Let $\gamma \in (0, 1)$
        \item Let $x, y \in \R^d$ satisfy $\|A(x - y)\|_\infty < 0.01$
        \item Let $R_f := \beta^{-2} n^{1.5} \exp{3R^2}$ 
        \item Let $\ell(x)$ be denoted as Definition~\ref{def:ell:formal}
        \item $\ell(x) \geq \gamma$
    \end{itemize}
    We have
    \begin{itemize}
        \item Part 1.
        \begin{align*}
            | \ell(x) - \ell(y) | \leq 4 R_f \cdot \|x - y\|_2
        \end{align*}
        \item Part 2. Let $p \in \{1, 2, 3\}$, then we have
        \begin{align*}
            | \ell(x)^{-p} - \ell(y)^{-p} | \leq 4p \gamma^{-(1 + p)} \cdot R_f \cdot \|x - y\|_2
        \end{align*}
    \end{itemize}
\end{lemma}

\begin{proof}
    \textbf{Proof of Part 1.}
    \begin{align*}
        | \ell(x) - \ell(y) |
        = & ~ | \langle c(x), c(x) \rangle - \langle c(y), c(y) \rangle | \\
        = & ~ | \langle c(x) - c(y), c(x) + c(y) \rangle | \\
        \leq & ~ \| c(x) - c(y) \|_2 \cdot \| c(x) + c(y) \|_2 \\
        \leq & ~ (\|c(x)\|_2 + \|c(y)\|_2) \cdot \| c(x) - c(y) \|_2 \\
        \leq & ~ 4 \cdot \| c(x) - c(y) \|_2 \\
        \leq & ~ 4 R_f \cdot \|x - y\|_2
    \end{align*}
    where the first equality follows from Fact~\ref{fact:vector}, the second equality follows from Fact~\ref{fact:vec_norm_bound} (Cauchy-Schwarz inequality), the third equality follows from Fact~\ref{fact:vec_norm_bound}, the fourth equality follows from Part 1 of Lemma~\ref{lem:vec_func_norm_bound}, the fifth equality follows from Part 2 of Lemma~\ref{lem:vec_lipschitz_tool}.

    \textbf{Proof of Part 2.}
    \begin{align*}
        | \ell(x)^{-p} - \ell(y)^{-p} | 
        = & ~ | \ell(x) - \ell(y) | \cdot | \sum_{i = 0}^{p - 1} \ell(x)^{-p + i}\ell(y)^{-i - 1}| \\
        \leq & ~ | \ell(x) - \ell(y) | \cdot | \sum_{i = 0}^{p - 1} \gamma^{-p + i}\gamma^{-i - 1}| \\
        = & ~ | \ell(x) - \ell(y) | \cdot | \sum_{i = 0}^{p - 1} \gamma^{-p - 1} | \\
        = & ~ | \ell(x) - \ell(y) | \cdot (p - 1 + 1) \cdot \gamma^{-p - 1} \\
        \leq & ~ 4 R_f \cdot \|x - y\|_2 \cdot (p - 1 + 1) \cdot \gamma^{-p - 1} \\
        = & ~ 4p \gamma^{-(1 + p)} \cdot R_f \cdot \|x - y\|_2
    \end{align*}
    where the first equality follows from $a^p - b^p = (a-b) ( a^{p-1} + a^{p-2} b + \cdots + a b^{p-2} +  b^{p-1} )$, the second equality follows from $\ell(x) \geq \gamma$, the third equality follows from simple algebra, the fourth equality follows from simple algebra, the fifth equality follows from Part 1 of Lemma~\ref{lem:scalar_lipschitz_tool_ell}, the last equality follows from simple algebra.

\end{proof}

\begin{lemma}\label{lem:scalar_lipschitz_tool_f_c}
    If the given conditions are satisfied
    \begin{itemize}
        \item Let $A \in \R^{n \times d}$
        \item Let $\beta \in (0, 0.1)$
        \item Let $R \geq 4$
        \item Let $x, y \in \R^d$ satisfy $\|A(x - y)\|_\infty < 0.01$
        \item Let $R_f := \beta^{-2} n^{1.5} \exp{3R^2}$ 
        \item Denote $f(x)$ as Definition~\ref{def:f:formal}
        \item Denote $c(x)$ as Definition~\ref{def:c:formal}
    \end{itemize}
    We have
    \begin{itemize}
        \item Part 1.
        \begin{align*}
            | \langle f(x), c(x) \rangle - \langle f(y), c(y) \rangle |\leq 3 R_f \cdot \|x - y\|_2
        \end{align*}
        \item Part 2.
        \begin{align*}
            | \langle f(x), c(x) \rangle^2 - \langle f(y), c(y) \rangle^2 |\leq 12 R_f \cdot \|x - y\|_2
        \end{align*}
    \end{itemize}
\end{lemma}

\begin{proof}
    \textbf{Proof of Part 1.}
    \begin{align*}
        | \langle f(x), c(x) \rangle - \langle f(y), c(y) \rangle |
        = & ~ | f(x)^\top c(x) - f(y)^\top c(y) | \\
        \leq & ~ \|f(x)\|_2 \cdot \|c(x) - c(y)\|_2 + \|f(x) - f(y)\|_2 \cdot \|c(x)\|_2 \\
        \leq & ~ \|c(x) - c(y)\|_2 + \|f(x) - f(x)\|_2 \cdot \|c(y)\|_2 \\
        \leq & ~ \|c(x) - c(y)\|_2 + 2 \|f(x)^\top - f(x)^\top\|_2 \\
        \leq & ~ R_f \cdot \|x - y\|_2 + 2 \|f(x)^\top - f(x)^\top\|_2 \\
        \leq & ~ R_f \cdot \|x - y\|_2 + 2 R_f \cdot \|x - y\|_2 \\
        \leq & ~ 3 R_f \cdot \|x - y\|_2
    \end{align*}
    where the first equality follows from Fact~\ref{fact:vector}, the second equality follows from Part 1 of  Lemma~\ref{lem:partializ_bound_tool}, the third equality follows from $\|f(x)\| \leq 1$, the fourth equality follows from Part 1 of Lemma~\ref{lem:vec_func_norm_bound}, the fifth equality follows from Part 2 of Lemma~\ref{lem:vec_lipschitz_tool}, the sixth equality follows from Part 1 of Lemma~\ref{lem:vec_lipschitz_tool}, the last equality follows from simple algebra.

    \textbf{Proof of Part 2.}
    \begin{align*}
        | \langle f(x), c(x) \rangle^2 - \langle f(y), c(y) \rangle^2 |
        \leq & ~ | \langle f(x), c(x) \rangle + \langle f(y), c(y) \rangle | \cdot | \langle f(x), c(x) \rangle - \langle f(y), c(y) \rangle | \\
        \leq & ~ | 2 + 2| \cdot | \langle f(x), c(x) \rangle - \langle f(y), c(y) \rangle | \\
        \leq & ~ | 2 + 2| \cdot 3 R_f \cdot \|x - y\|_2 \\
        = & ~ 12 R_f \cdot \|x - y\|_2
    \end{align*}
    where the first equality follows from simple algebra, the second equality follows from Part 2 of Lemma~\ref{lem:func_bounds}, the third equality follows from Part 1 of Lemma~\ref{lem:scalar_lipschitz_tool_f_c}, the fourth equality follows from simple algebra.
\end{proof}

\begin{lemma}\label{lem:scalar_lipschitz_tool_ell_f_c}
    If the given conditions are satisfied
    \begin{itemize}
        \item Let $A \in \R^{n \times d}$
        \item Let $\beta \in (0, 0.1)$
        \item Let $R \geq 4$
        \item Let $\gamma \in (0, 1)$
        \item Let $x, y \in \R^d$ satisfy $\|A(x - y)\|_\infty < 0.01$
        \item Let $R_f := \beta^{-2} n^{1.5} \exp{3R^2}$ 
        \item Denote $f(x)$ as Definition~\ref{def:f:formal}
        \item Denote $c(x)$ as Definition~\ref{def:c:formal}
        \item Let $\ell(x)$ be denoted as Definition~\ref{def:ell:formal}
        \item $\ell(x) \geq \gamma$
    \end{itemize}
    We have
    \begin{itemize}
        \item Part 1.
        \begin{align*}
            | \ell(x)^{-3} \cdot \langle f(x), c(x) \rangle - \ell(y)^{-3} \cdot \langle f(y), c(y) \rangle | \leq 27 \gamma^{-4} \cdot R_f \| x - y \|_2
        \end{align*}
        \item Part 2.
        \begin{align*}
            | \ell(x)^{-3} \cdot \langle f(x), c(x) \rangle^2 - \ell(y)^{-3} \cdot \langle f(y), c(y) \rangle^2 | \leq 60 \gamma^{-4} \cdot R_f \cdot \| x - y \|_2
        \end{align*}
    \end{itemize}
\end{lemma}

\begin{proof}
    \textbf{Proof of Part 1.}
    \begin{align*}
        & ~ | \ell(x)^{-3} \cdot \langle f(x), c(x) \rangle - \ell(y)^{-3} \cdot \langle f(y), c(y) \rangle | \\
        = & ~ | \ell(x)^{-3} | \cdot | \langle f(x), c(x) \rangle - \langle f(y), c(y) \rangle | + | \ell(x)^{-3} - \ell(y)^{-3} | \cdot | \langle f(y), c(y) \rangle | \\
        \leq & ~ \gamma^{-3} \cdot | \langle f(x), c(x) \rangle - \langle f(y), c(y) \rangle | + | \ell(x)^{-3} - \ell(y)^{-3} | \cdot | \langle f(y), c(y) \rangle | \\
        \leq & ~ \gamma^{-3} \cdot 3 R_f \cdot \| x - y \|_2 + | \ell(x)^{-3} - \ell(y)^{-3} | \cdot | \langle f(y), c(y) \rangle | \\
        \leq & ~ \gamma^{-3} \cdot 3 R_f \cdot \| x - y \|_2 + 12 \gamma^{-4} \cdot R_f \cdot \| x - y \|_2 \cdot | \langle f(y), c(y) \rangle | \\
        \leq & ~ \gamma^{-3} \cdot 3 R_f \cdot \| x - y \|_2 + 12 \gamma^{-4} \cdot R_f \cdot \| x - y \|_2 \cdot 2 \\
        = & ~ ( 3 \gamma^{-3} + 24 \gamma^{-4} ) \cdot R_f \| x - y \|_2  \\
        \leq & ~ 27 \gamma^{-4} \cdot R_f \| x - y \|_2 
    \end{align*}
    where the first equality follows from Part 4 of Lemma~\ref{lem:partializ_bound_tool}, the second equality follows from $\ell(x) \geq \gamma$, the third equality follows from Part 1 of Lemma~\ref{lem:scalar_lipschitz_tool_f_c}, the fourth equality follows from Part 2 of Lemma~\ref{lem:scalar_lipschitz_tool_ell}, the fifth equality follows from Part 2 of Lemma~\ref{lem:func_bounds}, the sixth equality follows from simple algebra, the last equality follows from $\gamma \in (0, 1)$.

    \textbf{Proof of Part 2.}
    \begin{align*}
        & ~ | \ell(x)^{-3} \cdot \langle f(x), c(x) \rangle^2 - \ell(y)^{-3} \cdot \langle f(y), c(y) \rangle^2 | \\
        \leq & ~ | \ell(x)^{-3} \cdot \langle f(x), c(x) \rangle | \cdot | \langle f(x), c(x) \rangle - \langle f(y), c(y) \rangle | \\
        & ~ + | \ell(x)^{-3} \cdot \langle f(x), c(x) \rangle - \ell(y)^{-3} \cdot \langle f(y), c(y) \rangle | \cdot | \langle f(y), c(y) \rangle | \\
        \leq & ~ | \ell(x)^{-3} \cdot \langle f(x), c(x) \rangle | \cdot | \langle f(x), c(x) \rangle - \langle f(y), c(y) \rangle | \\
        & ~ + | \ell(x)^{-3} \cdot \langle f(x), c(x) \rangle - \ell(y)^{-3} \cdot \langle f(y), c(y) \rangle | \cdot 2 \\
        \leq & ~ 2 \gamma^{-3} \cdot | \langle f(x), c(x) \rangle - \langle f(y), c(y) \rangle | + | \ell(x)^{-3} \cdot \langle f(x), c(x) \rangle - \ell(y)^{-3} \cdot \langle f(y), c(y) \rangle | \cdot 2 \\
        \leq & ~ 2 \gamma^{-3} \cdot | \langle f(x), c(x) \rangle - \langle f(y), c(y) \rangle | +  27 \gamma^{-4} \cdot R_f \| x - y \|_2 \cdot 2 \\
        \leq & ~ 2 \gamma^{-3} \cdot 3 R_f \cdot \| x - y \|_2 + 27 \gamma^{-4} \cdot R_f \cdot \| x - y \|_2 \cdot 2 \\
        = & ~ 60 \gamma^{-4} \cdot R_f \cdot \| x - y \|_2 
    \end{align*}
    where the first equality follows from Part 4 of Lemma~\ref{lem:partializ_bound_tool}, the second equality follows from Part 3 of Lemma~\ref{lem:func_bounds}, the third equality follows from $\ell(x) \geq \gamma$, the fourth equality follows from Part 1 of Lemma~\ref{lem:scalar_lipschitz_tool_ell_f_c}, the fifth equality follows from Part 1 of Lemma~\ref{lem:scalar_lipschitz_tool_f_c}, the sixth equality follows from simple algebra.
\end{proof}

\subsection{Lipschitz Tool: Vector Function}\label{subapp:tool_vector}

\begin{lemma}\label{lem:vec_lipschitz_tool}
    If the given conditions are satisfied
    \begin{itemize}
        \item Let $A \in \R^{n \times d}$
        \item Let $b \in \R^{n}$ satisfy that $\| b \|_1 \leq 1$
        \item Let $\beta \in (0, 0.1)$
        \item Let $R \geq 4$
        \item Let $x, y \in \R^d$ satisfy $\|A(x - y)\|_\infty < 0.01$
        \item $\| A \| \leq R$
        \item Let $R_f := \beta^{-2} n^{1.5} \exp{3R^2}$ 
        \item Denote $f(x)$ as Definition~\ref{def:f:formal}
        \item Denote $c(x)$ as Definition~\ref{def:c:formal}
    \end{itemize}
    We have
    \begin{itemize}
        \item Part 1. (see Part 4 of Lemma 7.2 in \cite{dls23})
        \begin{align*}
            \|f(x) - f(y)\|_2 \leq R_f \cdot \|x - y\|_2
        \end{align*}
        \item Part 2. (see Part 5 of Lemma 7.2 in \cite{dls23})
        \begin{align*}
            \|c(x) - c(y)\|_2 \leq R_f \cdot \|x - y\|_2
        \end{align*}
        \item Part 3.
        \begin{align*}
            \| f(x) \circ c(x) - f(y) \circ c(y) \|_2 \leq 3 R_f \cdot \| x - y \|_2
        \end{align*}
        \end{itemize}
\end{lemma}

\begin{proof}
    \textbf{Proof of Part 3.}
    \begin{align*}
        \| f(x) \circ c(x) - f(y) \circ c(y) \|_2
        \leq & ~ \| f(x) \|_2 \cdot \| c(x) - c(y) \|_2 + \| f(x) - f(y) \|_2 \cdot \| c(y) \|_2 \\
        \leq & ~ \| c(x) - c(y) \|_2 + \| f(x) - f(y) \|_2 \cdot \| c(y) \|_2 \\
        \leq & ~ \| c(x) - c(y) \|_2 + 2 \cdot \| f(x) - f(y) \|_2 \\
        \leq & ~ R_f \cdot \| x - y \|_2 + 2 \cdot \| f(x) - f(y) \|_2 \\
        \leq & ~ R_f \cdot \| x - y \|_2 + 2 R_f \cdot \| x - y \|_2 \\
        = & ~ 3 R_f \cdot \| x - y \|_2
    \end{align*}
    where the first equality follows from Part 5 of Lemma~\ref{lem:partializ_bound_tool}, the second equality follows from $\|f(x)\|_2 \leq 1$, the third equality follows from Part 1 of Lemma~\ref{lem:vec_func_norm_bound}, the fourth equality follows from Part 2 of Lemma~\ref{lem:vec_lipschitz_tool}, the fifth equality follows from Part 1 of Lemma~\ref{lem:vec_lipschitz_tool}, the sixth equality follows from simple algebra.

\end{proof}

\section{Helpful Bounds}\label{app:helpful_bounds}

In this appendix, we present a collection of bound that are valuable for facilitating computations in our proofs.

\begin{lemma}\label{lem:vec_func_norm_bound}
    If the given conditions are satisfied
    \begin{itemize}
        \item Let $b \in \R^{n}$ satisfy that $\| b \|_1 \leq 1$
        \item Denote $f(x)$ as Definition~\ref{def:f:formal}
        \item Denote $c(x)$ as Definition~\ref{def:c:formal}
        \item $\|f(x)\|_2 \leq 1$
    \end{itemize}
    then we have
    \begin{itemize}
        \item Part 1. $\|c(x)\|_2 \leq 2 $
        \item Part 2. $\| f(x) \circ c(x) \|_2 \leq 2$
    \end{itemize}
\end{lemma}

\begin{proof}
    \textbf{Proof of Part 1.}
    \begin{align*}
        \|c(x)\|_2 
        = & ~ \|f(x) - b\|_2 \\
        \leq & ~ \|f(x)\|_2 + \|b\|_2 \\
        \leq & ~ 1 + \|b\|_2 \\
        \leq & ~ 1 + 1 = 2
    \end{align*}
    where the first equality follow from Definition~\ref{def:c:formal}, the second equality follows Fact~\ref{fact:vec_norm_bound}, the third equality follows from $\|f(x)\|_2 \leq 1$, the fourth equality follows from $\|b\|_2 \leq 1$.

    \textbf{Proof of Part 2.}
    \begin{align*}
        \| f(x) \circ c(x) \|_2
        \leq & ~ \| f(x) \|_2 \cdot \| c(x) \|_2 \\
        \leq & ~ 1 \cdot \| c(x) \|_2 \\
        \leq & ~ 2
    \end{align*}
    where the first equalities follow from Fact~\ref{fact:vec_norm_bound}, the second equality follows from $\|f(x)\|_2 \leq 1$, the third equality follows from Part 1 of Lemma~\ref{lem:vec_func_norm_bound}.
\end{proof}

\begin{lemma}\label{lem:func_bounds}
    If the given conditions are satisfied
    \begin{itemize}
        \item Let $b \in \R^{n}$ satisfy that $\| b \|_1 \leq 1$
        \item Denote $f(x)$ as Definition~\ref{def:f:formal}
        \item Denote $c(x)$ as Definition~\ref{def:c:formal}
        \item Let $\ell(x)$ be denoted as Definition~\ref{def:ell:formal}
        \item $\|f(x)\|_2 \leq 1$
        \item Let $\gamma \in (0, 1)$
        \item $\ell(x) \geq \gamma$
    \end{itemize}
    then we have
    \begin{itemize}
        \item Part 1. $\ell(x) \leq 4$
        \item Part 2. $0.25 \| b \|_2^2 \leq 0 \leq \langle f(x), c(x) \rangle \leq 2$
        \item Part 3. Let $p, q \in \{1, 2, 3\}$, we have $\ell(x)^{-p} \langle f(x), c(x) \rangle^q \leq 2q\gamma^{-p}$
    \end{itemize}
\end{lemma}

\begin{proof}
    \textbf{Proof of Part 1.}
    \begin{align*}
        \ell(x)
        = & ~ \langle c(x), c(x) \rangle \\
        \leq & ~ \|c(x)\|_2 \cdot \|c(x)\|_2 \\
        \leq & ~ 2 \times 2 = 4
    \end{align*}
    where the first equality follow from Definition~\ref{def:ell:formal}, the second equality follows from Fact~\ref{fact:vec_norm_bound} (Cauchy-Schwarz inequality), the third equality follows from Part 1 of Lemma~\ref{lem:vec_func_norm_bound}.

    \textbf{Proof of Part 2.}
    On one hand, we have
    \begin{align*}
        \langle f(x), c(x) \rangle
        \leq & ~ \|f(x)\|_2 \cdot \|c(x)\|_2 \\
        \leq & ~ 1 \cdot \|c(x)\|_2 \\
        \leq & ~ 2
    \end{align*}
    where the first equality follows from Fact~\ref{fact:vec_norm_bound} (Cauchy-Schwarz inequality), the second equality follows from $\|f(x)\|_2 \leq 1$, the third equality follows from Part 1 of Lemma~\ref{lem:vec_func_norm_bound}.

    On the other hand, we have
    \begin{align*}
        \langle f(x), c(x) \rangle
        = & ~ \langle f(x), f(x) - b \rangle \\
        = & ~ \| f(x) \|_2^2 - \langle f(x), b \rangle \\
        = & ~ \| f(x) \|_2^2 - \langle f(x), b \rangle + 0.25 \| b \|_2^2 - 0.25 \| b \|_2^2 \\
        = & ~ \| f(x) - 0.5 b \|_2^2 - 0.25 \| b \|_2^2 \\
        \geq & ~ 0 \geq - 0.25 \| b \|_2^2 
    \end{align*}
    where the first equality follows from Fact~\ref{fact:vec_norm_bound}, the second, third, fourth fifth equalities follow from simple algebras.

    \textbf{Proof of Part 3.}
    \begin{align*}
        \ell(x)^{-p} \langle f(x), c(x) \rangle^q 
        \leq & ~ \gamma^{-p} \cdot \langle f(x), c(x) \rangle^q \\
        \leq & ~ 2q \gamma^{-p}
    \end{align*}
    where the first equality follows from $\ell(x) > \gamma$, the second equality follows from Part 2 of Lemma~\ref{lem:func_bounds}.
\end{proof}

\begin{lemma}\label{lem:mat_func_norm_bounds}
    If the given conditions are satisfied
    \begin{itemize}
        \item Let $b \in \R^{n}$ satisfy that $\| b \|_1 \leq 1$
        \item Denote $B(x)$ as Definition~\ref{def:B:formal}
        \item $\|f(x)\|_2 \leq 1$
    \end{itemize}
    then we have
    \begin{itemize}
        \item Part 1. $\| B(x) \| \leq 11$
    \end{itemize}
\end{lemma}

\begin{proof}
    \textbf{Proof of Part 1.}
    \begin{align*}
        \| B(x) \| = & ~ \| \langle 3f(x) - 2b, f(x)\rangle \cdot f(x) f(x)^\top + \langle f(x) - b, f(x) \rangle \cdot \diag(f(x)) \\
        & ~ + \diag((2f(x) - b) \circ f(x)) + (b \circ f(x)) \cdot f(x)^\top + f(x) \cdot (b \circ f(x))^\top \| \\
        \leq & ~ \| \langle 3f(x) - 2b, f(x)\rangle \cdot f(x) f(x)^\top \| + \| \langle f(x) - b, f(x) \rangle \cdot \diag(f(x)) \| \\
        & ~ + \| \diag((2f(x) - b) \circ f(x)) \| + \| (b \circ f(x)) \cdot f(x)^\top + f(x) \cdot (b \circ f(x))^\top \|
    \end{align*}
    where the first equality follows from Lemma~\ref{lem:rewrite_B}.

    For the first term
    \begin{align*}
        \| \langle 3f(x) - 2b, f(x)\rangle \cdot f(x) f(x)^\top \|
        \leq & ~ \| \langle 3f(x) - 2b, f(x)\rangle \| \cdot \| f(x) f(x)^\top \| \\ 
        \leq & ~ \| \langle 3f(x) - 2b, f(x)\rangle \| \cdot \| f(x) \|_2 \cdot \| f(x) \|_2 \\ 
        \leq & ~ \| 3f(x) - 2b \|_2 \cdot \| f(x) \|_2 \cdot \| f(x) \|_2 \cdot \| f(x) \|_2 \\ 
        \leq & ~ ( \| 3f(x) \|_2 + \| 2b \|_2 ) \cdot \| f(x) \|_2 \cdot \| f(x) \|_2 \cdot \| f(x) \|_2 \\
        \leq & ~ 3 + \| 2b \|_2 \\
        \leq & ~ 3 + 2 = 5
    \end{align*}
    where the first, second equalities follow from Fact~\ref{fact:mat_norm_bound}, the third, fourth equalities follow from Fact~\ref{fact:vec_norm_bound}, the fifth equality follows from $\|f(x)\|_2 \leq 1$, the sixth equality follows from $\| b \|_2 \leq 1$, the last equality follows from simple algebra.

    For the second term
    \begin{align*}
        \| \langle f(x) - b, f(x) \rangle \cdot \diag(f(x)) \|
        \leq & ~ | \langle f(x) - b, f(x) \rangle | \cdot \| \diag(f(x)) \| \\
        \leq & ~ \| f(x) - b \|_2 \cdot \| f(x) \|_2 \cdot \| \diag(f(x)) \| \\
        \leq & ~ \| f(x) - b \|_2 \cdot \| f(x) \|_2 \cdot \| f(x) \|_2 \\
        \leq & ~ ( \| f(x) \|_2 + \| b \|_2 ) \cdot \| f(x) \|_2 \cdot \| f(x) \|_2 \\
        \leq & ~  \| b \|_2 \\
        \leq & ~ 1
    \end{align*}
    where the first equality follows from Fact~\ref{fact:mat_norm_bound}, the second equality follows from Fact~\ref{fact:vec_norm_bound} (Cauchy-Schwarz inequality), the third, fourth equalities follow from Fact~\ref{fact:vec_norm_bound}, the fifth equality follows from $\|f(x)\|_2 \leq 1$, the sixth equality follows from $\| b \|_2 \leq 1$.

    For the third term
    \begin{align*}
        \| \diag((2f(x) - b) \circ f(x)) \|
        \leq & ~ \| (2f(x) - b) \circ f(x) \|_2 \\
        \leq & ~ \| 2f(x) - b \|_2 \cdot \| f(x) \|_2 \\
        \leq & ~ ( \| 2f(x) \|_2 + \| b \|_2 ) \cdot \| f(x) \|_2 \\
        \leq & ~ 2 + \| b \|_2 \\
        \leq & ~ 2 + 1 = 3
    \end{align*}
    where the first, second, third equalities follow from Fact~\ref{fact:vec_norm_bound}, the fourth equality follows from $\| f(x) \|_2 \leq 1$, the fifth equality follows from $\| b \|_2 = 1$, the last equality follows from simple algebra.

    For the fourth term
    \begin{align*}
        \| (b \circ f(x)) \cdot f(x)^\top + f(x) \cdot (b \circ f(x))^\top \|
        \leq & ~ \| (b \circ f(x)) \cdot f(x)^\top \| + \| f(x) \cdot (b \circ f(x))^\top \| \\
        \leq & ~ \| b \circ f(x) \|_2 \cdot \| f(x) \|_2 + \| f(x) \|_2 \cdot \| b \circ f(x) \|_2 \\
        \leq & ~ 2 \| b \circ f(x) \|_2 \\
        \leq & ~ 2 (\|b\|_2 \cdot \|f(x)\|_2) \\
        \leq & ~ 2 (\|b\|_2 \cdot 1 ) \\
        \leq & ~ 2 (1 \cdot 1) \\
        = & ~ 2
    \end{align*}
    where the first equality follows from Fact~\ref{fact:vec_norm_bound}, the second equality follows from Fact~\ref{fact:vec_norm_bound}, the third equality follows from $\| f(x) \|_2 \leq 1$, the fourth equality follows from $\| b \|_2 = 1$, the last equality follows from simple algebra.

    Then we combine four terms, we have
    \begin{align*}
        \| B(x) \| \leq & ~ 5 + 1 + 3 + 2 = 11
    \end{align*}
    where the last equality follows from simple algebra.
\end{proof}

\section{Minimizing Loss Guarantee}\label{app:main_result}

In this appendix, our objective is to minimize $L$ to its optimal value, ensuring that we achieve the most favorable outcome in terms of our training process. The minimization guarantee of $L$ confirms our main result on optimization of Copyright Regression, it also demonstrates the ease of use of Copyright Regression, which can be optimized on any attention-based model.

We provide our result and proof below.
\begin{theorem}[Minimizing training objective $L$, formal version of Theorem~\ref{the:main_result:informal}]\label{the:main_result:formal}
    Suppose we have matrix $A \in \R^{n \times d}$ and $A_1 \in \R^{n_1 \times d}$, $n_1 \leq n$, vector $b, w \in \R^n$. Let $L$ be defined as Definition~\ref{def:L:formal}, denote $x^*$ as the optimal solution of $L$ where $g(x^*) = \mathbf{0}_d$ and $\| x^* \| \leq R$. Denote $R \geq 10$ be a positive scalar. Denote $M = n^{1.5}\exp(60R^2)$, Let $x_0$ be denoted as an initial point where $M \| x_0 - x^* \|_2 \leq 0.1l$, where $l > 0$ denoted a scalar.

    For any accuracy $\epsilon \in (0, 0.1)$ and any failure probability $\delta \in (0, 0.1)$, there exists a randomized algorithm, with probability $1 - \delta$, it runs $T = \log(\| x_0 - x^* \|_2 / \epsilon)$ iteration and outputs a vector $\widetilde{x} \in \R^d$ such that
    \begin{align*}
        \| \widetilde{x} - x^* \| \leq \epsilon
    \end{align*}
    and the time cost of each iteration is
    \begin{align*}
        O( ( \mathrm{nnz}(A) + d^w ) \cdot \mathrm{poly}( \log( n / \delta ) ) )
    \end{align*}
    Here $w$ is the exponent of matrix multiplication. Currently $w \approx 2.373$.
\end{theorem}

\begin{proof}
\textbf{Proof of Upper bound on $M$.}

By Lemma~\ref{lem:lipschitz:formal}, we have
\begin{align*}
    \| \nabla^2 L(x) - \nabla^2 L(y) \| \leq ( 13344 \gamma_c + 2 ) \gamma^{-4} \beta^{-2} n^{1.5} \exp(40R^2) \| x - y \|_2
\end{align*}

Hence, we have
\begin{align}\label{eq:upper_M}
    M \leq & ~ ( 13344 \gamma_c + 2 ) \gamma^{-4} \beta^{-2} n^{1.5} \exp(40R^2) \| x - y \|_2 \notag \\
    \leq & ~ ( 13344 \gamma_c + 2 ) \gamma^{-4} n^{1.5} \exp(50R^2) \| x - y \|_2 \notag \\
    \leq & ~ ( 13344 \gamma_c + 2 ) (\sqrt{2 \gamma_c})^{-4} n^{1.5} \exp(50R^2) \| x - y \|_2 \notag \\
    \leq & ~ ( 13344 \gamma_c + 2 ) (2 \gamma_c)^{-2} n^{1.5} \exp(50R^2) \| x - y \|_2 \notag \\
    \leq & ~ ( 26688 \gamma_c^{-1} + 4 \gamma_c^{-2} ) n^{1.5} \exp(50R^2) \| x - y \|_2
\end{align}
where the first equality follows from Lemma~\ref{lem:lipschitz:formal}, the second equality follows from Lemma~\ref{lem:lower_beta}, the third equality follows from Lemma~\ref{lem:ell_1(x^*):formal}, the fourth, fifth equalities follow from simple algebras.

Consider that $\gamma_c$ should not be greater than $0.5$, we have
\begin{align*}
    M
    \leq & ~ ( 26688 \gamma_c^{-1} + 4 \gamma_c^{-2} ) n^{1.5} \exp(50R^2) \| x - y \|_2 \\
    \leq & ~ 26692 \gamma_c^{-2} n^{1.5} \exp(50R^2) \| x - y \|_2 \\
    \leq & ~ n^{1.5} \exp(60R^2) \| x - y \|_2 
\end{align*}
where the first equality follows from Eq~\eqref{eq:upper_M}, the second equality follows from $\gamma_c \leq 0.5$, the third equality follows from simple algebra.

\textbf{Proof of Hessian is PD.}

The positive definiteness of the Hessian matrix follows directly from Lemma~\ref{lem:psd:formal}.

\textbf{Proof of Hessian is Lipschitz.}

The Lipschitz property of the Hessian matrix can be established using Lemma~\ref{lem:lipschitz:formal}.

\textbf{Proof of Cost per iteration.}

The cost per iteration can be deduced from Lemma~\ref{lem:iter_cost}.

\textbf{Proof of Convergence per Iteration.}

By utilizing Lemma~\ref{lem:iter_shrink}, it can be shown that:
\begin{align*}
| x_k - x^* |2 \leq 0.4 | x{k-1} - x^* |_2
\end{align*}

\textbf{Proof of Number of Iterations.}

After performing $T$ iterations, we obtain the following result:
\begin{align*}
| x_k - x^* |2 \leq 0.4^T | x{k-1} - x^* |_2
\end{align*}
By appropriately choosing the value of $T$, we can achieve the desired bound. The failure probability is derived from applying the union bound over the $T$ iterations.
\end{proof}

\section{\texorpdfstring{$L$}{} is \texorpdfstring{$\tau_c$}{}-Copyright-Protected}\label{app:copyright_protected}

In this appendix, we show our result that Copyright Regression avoids model outputting copyright data. In Appendix~\ref{subapp:problem_def}, we reaffirm our definition of problem and optimal parameter $x^*$ of Copyright Regression. In Appendix~\ref{subapp:cp_peroperty}, we provide our result and proof of $x^*$ is $\tau_c$-Copyright-Protected, where $\tau_c = \sqrt{2\gamma_c} / n_1 - \epsilon_2 / n_2$.

\subsection{Definitions}\label{subapp:problem_def}

\begin{definition}[$\tau$-Copyright-Protected]\label{def:probelm:formal}
Given a matrix $A \in \R^{n \times d}$ and vector $b \in \R^n$ that $A = \begin{bmatrix}
    A_1 \\
    A_2
\end{bmatrix}$, and $b = \begin{bmatrix}
    b_1 \\
    b_2
\end{bmatrix}$, where $A_1 \in \R^{n_1 \times d}$, $A_2 \in \R^{n_2 \times d}$, $b_1 \in \R^{n_1}$, $b_2 \in \R^{n_2}$ and $n = n_1 + n_2$. $A_1$, $b_1$ are the data has copyright issue and $A_2$, $b_2$ are the data does not have copyright issue. Denote the train objective $L$. Denote $\tau > 0$ a scalar.

If there is a trained model $f_\theta$ with parameter $\theta$ that satisfies
\begin{align*}
    \frac{L(f_\theta(A_1), b_1)}{n_1} \geq \tau + \frac{L(f_\theta(A_2), b_2)}{n_2}
\end{align*}
then we say this model $f_\theta$ is $\tau$-Copyright-Protected.
\end{definition}

\begin{definition}\label{def:x^*:formal}
    Let $x \in \R^d$ be a vector parameter in regression problem. Let $L$ be denoted as Definition~\ref{def:L:formal}, let $\ell_1(x)$ and $\ell_2(x)$ be denoted as Definition~\ref{def:L_copyright:formal}. Denote $\gamma_c > 0$ a scalar. Denote $\epsilon_1, \epsilon_2 \in (0, 0.1)$ two scalars, we define that
    \begin{align*}
        x^* := \mathop{\arg\min}\limits_{x \in \R^d} L
    \end{align*}
    and $x^*$ satisfies 
    \begin{itemize}
        \item
        \begin{align*}
            & ~ 0.5 \ell_1(x^*) + \gamma_c \ell_1(x^*)^{-1} \\
            \leq & ~ \min_{x \in \R^d} ( 0.5 \ell_1(x^*) + \gamma_c \ell_1(x^*)^{-1} ) + \epsilon_1
        \end{align*}
        \item
        \begin{align*}
            \ell_2(x^*) \leq \epsilon_2 \leq \min_{x \in \R^d} \ell_2(x) + \epsilon_2 
        \end{align*}
    \end{itemize}
\end{definition}

\subsection{Copyright-Pretected Property for \texorpdfstring{$x^*$}{}}\label{subapp:cp_peroperty}

\begin{lemma}\label{lem:ell_1(x^*):formal}
    If the given conditions are satisfied
    \begin{itemize}
        \item Let $x^*$ be denoted as Definition~\ref{def:x^*:formal}
        \item Let $\ell_1(x)$ be denoted as Definition~\ref{def:L_copyright:formal}
        \item Denote $\epsilon_1, \epsilon_2 \in (0, 0.1)$ two scalars
        \item Denote $\gamma_c > 0$ a scalar
    \end{itemize}
    we have
    \begin{align*}
        \sqrt{2 \gamma_c} \leq \ell_1(x^*) \leq & ~  ( \sqrt{2 \gamma_c} + \epsilon_1 ) + \sqrt{\epsilon_1^2 + 2 \epsilon_1 \sqrt{2 \gamma_c}}
    \end{align*}
\end{lemma}

\begin{proof}
    To compute the $\min_{x \in \R^d} ( 0.5 \ell_1(x) + \gamma_c \ell_1(x)^{-1} )$, we have
    \begin{align}\label{eq:d_ell+ell-1}
        \frac{\d ( 0.5 \ell_1(x) + \gamma_c \ell_1(x)^{-1} )}{\d \ell_1(x)}
        = & ~ \frac{\d 0.5 \ell_1(x)}{\d \ell_1(x)} + \frac{\d \gamma_c \ell_1(x)^{-1}}{\d \ell_1(x)} \notag \\
        = & ~ 0.5 + \frac{\d \gamma_c \ell_1(x)^{-1}}{\d \ell_1(x)} \notag \\
        = & ~ 0.5 + \gamma_c \frac{\d \ell_1(x)^{-1}}{\d \ell_1(x)} \notag \\
        = & ~ 0.5 - \gamma_c \frac{1}{\ell_1(x)^{-2}}
    \end{align}
    where the first, second, third, fourth equalities follow from simple differential rules.

    Hence, when $\frac{\d ( 0.5 \ell_1(x) + \gamma_c \ell_1(x)^{-1} )}{\d \ell_1(x)} = 0$, applying Eq.~\eqref{eq:d_ell+ell-1}, we have
    \begin{align}\label{eq:ell_1}
        0.5 - \gamma_c \frac{1}{\ell_1(x)^{-2}} = & ~ 0
    \end{align}
    solving Eq.~\eqref{eq:ell_1} yields
    \begin{align*}
        \ell_1(x) = & ~ \sqrt{2 \gamma_c}
    \end{align*}

    Bring $\ell_1(x)$ into the $0.5 \ell_1(x) + \gamma_c \ell_1(x)^{-1}$, and we have
    \begin{align*}
        \min_{x \in \R^d} ( 0.5 \ell_1(x) + \gamma_c \ell_1(x)^{-1} )
        = & ~ 0.5 \sqrt{2 \gamma_c} + \gamma_c \frac{1}{\sqrt{2 \gamma_c}} \\
        = & ~ \sqrt{2 \gamma_c}
    \end{align*}
    where the first equality follows from $\ell_1(x) = \sqrt{2 \gamma_c}$, the second equality follows from simple algebra.

    So we have
    \begin{align}\label{eq:ell_x^*}
        0.5 \ell_1(x^*) + \gamma_c \ell_1(x^*)^{-1}
        \leq & ~ \min_{x \in \R^d} ( 0.5 \ell_1(x^*) + \gamma_c \ell_1(x^*)^{-1} ) + \epsilon_1 \notag \\
        = & ~ \sqrt{2 \gamma_c} + \epsilon_1
    \end{align}
    solving Eq.~\eqref{eq:ell_x^*} yields
    \begin{align*}
        \ell_1(x^*) \leq & ~  ( \sqrt{2 \gamma_c} + \epsilon_1 ) + \sqrt{\epsilon_1^2 + 2 \epsilon_1 \sqrt{2 \gamma_c}}
    \end{align*}
    then we combine two terms, we have
    \begin{align*}
        \sqrt{2 \gamma_c} \leq \ell_1(x^*) \leq & ~  ( \sqrt{2 \gamma_c} + \epsilon_1 ) + \sqrt{\epsilon_1^2 + 2 \epsilon_1 \sqrt{2 \gamma_c}}
    \end{align*}
\end{proof}

\begin{theorem}[Formal version of Theorem~\ref{the:tau_c_protect:informal}]\label{the:tau_c_protect:formal}
    Let $x^*$ be denoted the trained parameter on Copyright Regression. Let $\ell(x)$ be denoted as Definition~\ref{def:ell:informal}, let $\ell(x)$ be the original train objective of Softmax Regression. Denote $\epsilon_2 \in (0 0.1)$ a scalar. Denote $\tau_c := \sqrt{2\gamma_c} / n_1 - \epsilon_2 / n_2$, we have
    \begin{align*}
        \frac{\ell_1(x^*)}{n_1} \geq \tau_c + \frac{\ell_2(x^*)}{n_2}
    \end{align*}
    so $x^*$ in Copyright Regression is $\tau_c$-Copyright-Protected.
    
\end{theorem}

\begin{proof}
    We have
    \begin{align*}
        \frac{\ell_1(x^*)}{n_1} - \frac{\ell_2(x^*)}{n_2} 
        \geq & ~ \frac{\sqrt{2 \gamma_c}}{n_1} - \frac{\ell_2(x)}{n_2} \\
        \geq & ~ \frac{\sqrt{2 \gamma_c}}{n_1} - \frac{\epsilon_2}{n_2}
    \end{align*}
    where the first equality follows from Lemma~\ref{lem:ell_1(x^*):formal}, the second equality follows from Definition~\ref{def:x^*:formal}.

    We define $\tau_c := \frac{\sqrt{2 \gamma_c}}{n_1} - \frac{\epsilon_2}{n_2}$, then we have
    \begin{align*}
        \frac{\ell_1(x^*)}{n_1} - \frac{\ell_2(x^*)}{n_2}
        \geq & ~ \frac{\sqrt{2 \gamma_c}}{n_1} - \frac{\epsilon_2}{n_2} \\
        \geq & ~ \tau_c
    \end{align*}
\end{proof}

\section{Approximate Newton Method}\label{app:approx_method}

In this appendix, we present an adapted version of the Newton method for convex optimization. In Appendix~\ref{subapp:def_update_rule}, we outline the assumptions underlying the conventional Newton method, along with the precise update rule employed by the traditional algorithm. Additionally, in Appendix~\ref{subapp:approx}, we introduce the approximate update rule for the modified Newton method. We also provide an implementation tool for computing the approximation of $\nabla^2 L$, and leverage certain lemmas from \cite{lsz23} to analyze the behavior and performance of the approximate Newton method.

\subsection{Definition and Update Rule}\label{subapp:def_update_rule}

In this section, our primary focus is on examining the local convergence properties of the Newton method. We direct our attention towards solving the optimization problem given by:
\begin{align*}
\min_{x \in \mathbb{R}^d} L(x)
\end{align*}
To facilitate our analysis, we make certain assumptions as follows
\begin{definition}[$(l, m)$-good Loss function, Definition 8.1 in \cite{dls23}]\label{def:good_init}
    Let $L : \R^d \to \R$ be denote as Definition~\ref{def:L:formal}, we say $L$ is $(l, m)$-good when it satisfies the following conditions,
    \begin{itemize}
        \item \textbf{$l$-local Minimum. } We define $l > 0$ to be a positive scalar. If there exists a vector $x^* \in \R^d$ satisfies
        \begin{itemize}
            \item $\nabla L(x^*) = \mathbf{0}_d$
            \item $\nabla^2 L(x^*) \succeq l \cdot \mathbf{I}_d$
        \end{itemize}
        \item \textbf{Hessian is $M$-Lipschitz. } If there exists a positive scalar $M > 0$ with
        \begin{align*}
            \| \nabla^2 L(x) - \nabla^2 L(y) \| \leq M \cdot \| x - y \|_2 
        \end{align*}
        \item \textbf{Good Initialization Point. } Let $x_0$ be denoted as the initialization point. If $r_0 := \| x_0 - x^* \|_2$ satisfies
        \begin{align*}
            r_0 M \leq 0.1 l
        \end{align*}
    \end{itemize}
\end{definition}

We define gradient and Hessian as follows
\begin{definition}[Gradient and Hessian]. 
    The gradient $g: \R^d \to \R^d$ of the loss function is defined as
    \begin{align*}
        g(x) := \nabla L(x)
    \end{align*}
    The Hessian $H: \R^d \to R^{d \times d}$ of the loss function is defined as
    \begin{align*}
        H(x) := \nabla^2 L(x)
    \end{align*}
\end{definition}

With the gradient function $g : \R^d \to \R^d$ and the Hessian matrix $H : \R^d \to \R^{d \times d}$, we define the exact process of the Newton method as follows:
\begin{definition}[Exact update of the Newton method, Definition 8.3 in \cite{dls23}]
    \begin{align*}
        x_{t+1} = x_t - H(x_t)^{-1} \cdot g(x_t)
    \end{align*}
\end{definition}

\subsection{Approximate of Hessian and Update Rule}\label{subapp:approx}

\begin{definition}[Approximate Hessian, Definition 8.4 in \cite{dls23}]\label{def:approx_hess}
    For any Hessian $H(x_t) \in \R^{d \times d}$, we define the mated Hessian $\widetilde{H}(x_t) \in R^{d \times d}$ to be a matrix such that the following holds,
    \begin{align*}
        ( 1 - \epsilon_0 ) \cdot H(x_t) \preceq \widetilde{H}(x_t) \preceq ( 1 + \epsilon_0 ) \cdot H(x_t)
    \end{align*}
\end{definition}

In order to get the approximated Hessian $\widetilde{H}(x_t)$ efficiently, here we state a standard tool (see Lemma 8.5 in \cite{dls23}).

\begin{lemma}[Lemma 8.5 in \cite{dls23}]\label{lem:iter_cost}
    Let $\epsilon_0 = 0.01$ be a constant precision parameter. Let $A \in \R^{n \times d}$ be a real matrix, then for any positive diagonal (PD) matrix $D \in \R^{n \times n}$, there exists an algorithm which runs in time
    \begin{align*}
        O ( ( \mathrm{nnz}(A) + d^w ) \mathrm{poly}( \log( n / \delta ) ) )
    \end{align*}
    and it outputs an $O(d\log(n/\delta))$ sparse diagonal matrix $\widetilde{D} \in \R^{n \times n}$ for which
    \begin{align*}
        (1 - \epsilon_0)A^\top D A \preceq A^\top \widetilde{D} A \preceq (1 + \epsilon_0) A^{\top} D A
    \end{align*}
\end{lemma}

Following the standard of Approximate Newton Hessian literature \cite{bps+20, szs+23, lsz23, dls23}, we consider the following.

\begin{definition}[Approximate update]
    We consider the following process
    \begin{align*}
        x_{t+1} = x_t - \widetilde{H}(x_t)^{-1} \cdot g(x_t)
    \end{align*}
\end{definition}

We state a tool from prior work, 

\begin{lemma}[Iterative shrinking Lemma, Lemma 6.9 in \cite{lsz23}]\label{lem:iter_shrink}
    If the following condition hold 
    \begin{itemize}
        \item Loss Function $L$ is $(l, M)$-good (see Definition~\ref{def:good_init})
        \item Let $\epsilon_0 \in (0, 0.1)$ (see Definition~\ref{def:approx_hess})
        \item Let $r_t := \| x_t - x^* \|_2$
        \item Let $\overline{r}_t := M \cdot r_t$
    \end{itemize}
    Then we have
    \begin{align*}
        r_{t+1} \leq 2 \cdot ( \epsilon_0 + \overline{r}_t / ( l - \overline{r}_t ) ) \cdot r_t
    \end{align*}
\end{lemma}

\begin{lemma}[Upper bound of $\beta$, Lemma 8.3 in \cite{dls23}]\label{lem:lower_beta}
    If the given conditions are satisfied
    \begin{itemize}
        \item $\| A \| \leq R$
        \item $\| x \|_2 \leq R$
        \item Let $\beta$ be the lower bound on $\langle \exp(Ax), \mathbf{1}_n \rangle$
    \end{itemize}
    then we have
    \begin{align*}
        \beta \geq \exp( - R^2)
    \end{align*}
\end{lemma}


\ifdefined\isarxiv

\bibliographystyle{alpha}
\bibliography{ref}

\newcommand{\etalchar}[1]{$^{#1}$}
\begin{thebibliography}{ADH{\etalchar{+}}19b}

\bibitem[ADF{\etalchar{+}}23]{adf+23}
Rohan Anil, Andrew~M Dai, Orhan Firat, Melvin Johnson, Dmitry Lepikhin,
  Alexandre Passos, Siamak Shakeri, Emanuel Taropa, Paige Bailey, Zhifeng Chen,
  et~al.
\newblock Palm 2 technical report.
\newblock {\em arXiv preprint arXiv:2305.10403}, 2023.

\bibitem[ADH{\etalchar{+}}19a]{adh+19a}
Sanjeev Arora, Simon Du, Wei Hu, Zhiyuan Li, and Ruosong Wang.
\newblock Fine-grained analysis of optimization and generalization for
  overparameterized two-layer neural networks.
\newblock In {\em International Conference on Machine Learning}, pages
  322--332. PMLR, 2019.

\bibitem[ADH{\etalchar{+}}19b]{adh+19b}
Sanjeev Arora, Simon~S Du, Wei Hu, Zhiyuan Li, Russ~R Salakhutdinov, and
  Ruosong Wang.
\newblock On exact computation with an infinitely wide neural net.
\newblock {\em Advances in neural information processing systems}, 32, 2019.

\bibitem[AG23]{ag23}
Sanjeev Arora and Anirudh Goyal.
\newblock A theory for emergence of complex skills in language models.
\newblock {\em arXiv preprint arXiv:2307.15936}, 2023.

\bibitem[ALS{\etalchar{+}}22]{als+22}
Josh Alman, Jiehao Liang, Zhao Song, Ruizhe Zhang, and Danyang Zhuo.
\newblock Bypass exponential time preprocessing: Fast neural network training
  via weight-data correlation preprocessing.
\newblock {\em arXiv preprint arXiv:2211.14227}, 2022.

\bibitem[AS23]{as23}
Josh Alman and Zhao Song.
\newblock Fast attention requires bounded entries.
\newblock {\em arXiv preprint arXiv:2302.13214}, 2023.

\bibitem[AZLS19a]{als19a}
Zeyuan Allen-Zhu, Yuanzhi Li, and Zhao Song.
\newblock A convergence theory for deep learning via over-parameterization.
\newblock In {\em International conference on machine learning}, pages
  242--252. PMLR, 2019.

\bibitem[AZLS19b]{als19b}
Zeyuan Allen-Zhu, Yuanzhi Li, and Zhao Song.
\newblock On the convergence rate of training recurrent neural networks.
\newblock {\em Advances in neural information processing systems}, 32, 2019.

\bibitem[BAR23]{bar23}
BARD.
\newblock Try bard, an ai experiment by google.
\newblock {\em Google}, February 2023.

\bibitem[BCE{\etalchar{+}}23]{bce23}
S{\'e}bastien Bubeck, Varun Chandrasekaran, Ronen Eldan, Johannes Gehrke, Eric
  Horvitz, Ece Kamar, Peter Lee, Yin~Tat Lee, Yuanzhi Li, Scott Lundberg,
  et~al.
\newblock Sparks of artificial general intelligence: Early experiments with
  gpt-4.
\newblock {\em arXiv preprint arXiv:2303.12712}, 2023.

\bibitem[BMR{\etalchar{+}}20]{bmr+20}
Tom Brown, Benjamin Mann, Nick Ryder, Melanie Subbiah, Jared~D Kaplan, Prafulla
  Dhariwal, Arvind Neelakantan, Pranav Shyam, Girish Sastry, Amanda Askell,
  et~al.
\newblock Language models are few-shot learners.
\newblock {\em Advances in neural information processing systems},
  33:1877--1901, 2020.

\bibitem[BNX{\etalchar{+}}23]{bnx+23}
Fan Bao, Shen Nie, Kaiwen Xue, Yue Cao, Chongxuan Li, Hang Su, and Jun Zhu.
\newblock All are worth words: A vit backbone for diffusion models.
\newblock In {\em Proceedings of the IEEE/CVF Conference on Computer Vision and
  Pattern Recognition}, pages 22669--22679, 2023.

\bibitem[BPSW20]{bps+20}
Jan van~den Brand, Binghui Peng, Zhao Song, and Omri Weinstein.
\newblock Training (overparametrized) neural networks in near-linear time.
\newblock {\em arXiv preprint arXiv:2006.11648}, 2020.

\bibitem[BSZ23]{bsz23}
Jan van~den Brand, Zhao Song, and Tianyi Zhou.
\newblock Algorithm and hardness for dynamic attention maintenance in large
  language models.
\newblock {\em arXiv preprint arXiv:2304.02207}, 2023.

\bibitem[CDW{\etalchar{+}}21]{cdw+20}
Beidi Chen, Tri Dao, Eric Winsor, Zhao Song, Atri Rudra, and Christopher
  R{\'e}.
\newblock Scatterbrain: Unifying sparse and low-rank attention.
\newblock {\em Advances in Neural Information Processing Systems},
  34:17413--17426, 2021.

\bibitem[CG19]{cg19}
Yuan Cao and Quanquan Gu.
\newblock Generalization bounds of stochastic gradient descent for wide and
  deep neural networks.
\newblock {\em Advances in neural information processing systems}, 32, 2019.

\bibitem[CGH{\etalchar{+}}19]{cgh+19}
Tianle Cai, Ruiqi Gao, Jikai Hou, Siyu Chen, Dong Wang, Di~He, Zhihua Zhang,
  and Liwei Wang.
\newblock Gram-gauss-newton method: Learning overparameterized neural networks
  for regression problems.
\newblock {\em arXiv preprint arXiv:1905.11675}, 2019.

\bibitem[Cha22]{cha22}
ChatGPT.
\newblock Optimizing language models for dialogue.
\newblock {\em OpenAI Blog}, November 2022.

\bibitem[CLP{\etalchar{+}}20]{clp+20}
Beidi Chen, Zichang Liu, Binghui Peng, Zhaozhuo Xu, Jonathan~Lingjie Li, Tri
  Dao, Zhao Song, Anshumali Shrivastava, and Christopher Re.
\newblock Mongoose: A learnable lsh framework for efficient neural network
  training.
\newblock In {\em International Conference on Learning Representations}, 2020.

\bibitem[CND{\etalchar{+}}22]{cnd+22}
Aakanksha Chowdhery, Sharan Narang, Jacob Devlin, Maarten Bosma, Gaurav Mishra,
  Adam Roberts, Paul Barham, Hyung~Won Chung, Charles Sutton, Sebastian
  Gehrmann, et~al.
\newblock Palm: Scaling language modeling with pathways.
\newblock {\em arXiv preprint arXiv:2204.02311}, 2022.

\bibitem[CWR{\etalchar{+}}22]{cwr+22}
He~Cao, Jianan Wang, Tianhe Ren, Xianbiao Qi, Yihao Chen, Yuan Yao, and Lei
  Zhang.
\newblock Exploring vision transformers as diffusion learners.
\newblock {\em arXiv preprint arXiv:2212.13771}, 2022.

\bibitem[DBK{\etalchar{+}}20]{dbk+20}
Alexey Dosovitskiy, Lucas Beyer, Alexander Kolesnikov, Dirk Weissenborn,
  Xiaohua Zhai, Thomas Unterthiner, Mostafa Dehghani, Matthias Minderer, Georg
  Heigold, Sylvain Gelly, et~al.
\newblock An image is worth 16x16 words: Transformers for image recognition at
  scale.
\newblock {\em arXiv preprint arXiv:2010.11929}, 2020.

\bibitem[DCLT18]{dcl+18}
Jacob Devlin, Ming-Wei Chang, Kenton Lee, and Kristina Toutanova.
\newblock Bert: Pre-training of deep bidirectional transformers for language
  understanding.
\newblock {\em arXiv preprint arXiv:1810.04805}, 2018.

\bibitem[DE21]{dal21}
DALL-E.
\newblock Dall·e: Creating images from text.
\newblock {\em OpenAI Research}, January 2021.

\bibitem[DE22]{dal22}
DALL-E2.
\newblock Dall·e 2 pre-training mitigations.
\newblock {\em OpenAI Research}, June 2022.

\bibitem[DLMS23]{dlms23}
Yichuan Deng, Zhihang Li, Sridhar Mahadevan, and Zhao Song.
\newblock Zero-th order algorithm for softmax attention optimization.
\newblock {\em arXiv preprint arXiv:2307.08352}, 2023.

\bibitem[DLS23]{dls23}
Yichuan Deng, Zhihang Li, and Zhao Song.
\newblock Attention scheme inspired softmax regression.
\newblock {\em arXiv preprint arXiv:2304.10411}, 2023.

\bibitem[DZPS18]{dzp+18}
Simon~S Du, Xiyu Zhai, Barnabas Poczos, and Aarti Singh.
\newblock Gradient descent provably optimizes over-parameterized neural
  networks.
\newblock {\em arXiv preprint arXiv:1810.02054}, 2018.

\bibitem[fttc22]{uni22}
United State~Courts for the~9th circuits.
\newblock Copying—access and substantial similarity.
\newblock {\em Model Civil Jury instructions}, December 2022.

\bibitem[Gil19]{gil19}
Jessica~L Gillotte.
\newblock Copyright infringement in ai-generated artworks.
\newblock {\em UC Davis L. Rev.}, 53:2655, 2019.

\bibitem[GMS23]{gms23}
Yeqi Gao, Sridhar Mahadevan, and Zhao Song.
\newblock An over-parameterized exponential regression.
\newblock {\em arXiv preprint arXiv:2303.16504}, 2023.

\bibitem[GSY23a]{gsy23_dp}
Yeqi Gao, Zhao Song, and Xin Yang.
\newblock Differentially private attention computation.
\newblock {\em arXiv preprint arXiv:2305.04701}, 2023.

\bibitem[GSY23b]{gsy23_coin}
Yeqi Gao, Zhao Song, and Junze Yin.
\newblock Gradientcoin: A peer-to-peer decentralized large language models.
\newblock {\em arXiv preprint arXiv:2308.10502}, 2023.

\bibitem[GSYZ23]{gsyz23_quantum}
Yeqi Gao, Zhao Song, Xin Yang, and Ruizhe Zhang.
\newblock Fast quantum algorithm for attention computation.
\newblock {\em arXiv preprint arXiv:2307.08045}, 2023.

\bibitem[HG15]{hg15}
Ben Hattenbach and Joshua Glucoft.
\newblock Patents in an era of infinite monkeys and artificial intelligence.
\newblock {\em Stan. Tech. L. Rev.}, 19:32, 2015.

\bibitem[HLSY21]{hls+21}
Baihe Huang, Xiaoxiao Li, Zhao Song, and Xin Yang.
\newblock Fl-ntk: A neural tangent kernel-based framework for federated
  learning analysis.
\newblock In {\em International Conference on Machine Learning}, pages
  4423--4434. PMLR, 2021.

\bibitem[Hri16]{hri16}
Kalin Hristov.
\newblock Artificial intelligence and the copyright dilemma.
\newblock {\em Idea}, 57:431, 2016.

\bibitem[HWC{\etalchar{+}}22]{hwc+20}
Kai Han, Yunhe Wang, Hanting Chen, Xinghao Chen, Jianyuan Guo, Zhenhua Liu,
  Yehui Tang, An~Xiao, Chunjing Xu, Yixing Xu, et~al.
\newblock A survey on vision transformer.
\newblock {\em IEEE transactions on pattern analysis and machine intelligence},
  45(1):87--110, 2022.

\bibitem[HXL{\etalchar{+}}22]{hxl+22}
Xuanli He, Qiongkai Xu, Lingjuan Lyu, Fangzhao Wu, and Chenguang Wang.
\newblock Protecting intellectual property of language generation apis with
  lexical watermark.
\newblock In {\em Proceedings of the AAAI Conference on Artificial
  Intelligence}, volume~36, pages 10758--10766, 2022.

\bibitem[HXZ{\etalchar{+}}22]{hxq+22}
Xuanli He, Qiongkai Xu, Yi~Zeng, Lingjuan Lyu, Fangzhao Wu, Jiwei Li, and Ruoxi
  Jia.
\newblock Cater: Intellectual property protection on text generation apis via
  conditional watermarks.
\newblock {\em Advances in Neural Information Processing Systems},
  35:5431--5445, 2022.

\bibitem[IJA{\etalchar{+}}23]{ija+23}
Oana Ignat, Zhijing Jin, Artem Abzaliev, Laura Biester, Santiago Castro, Naihao
  Deng, Xinyi Gao, Aylin Gunal, Jacky He, Ashkan Kazemi, et~al.
\newblock A phd student's perspective on research in nlp in the era of very
  large language models.
\newblock {\em arXiv preprint arXiv:2305.12544}, 2023.

\bibitem[JRL23]{jrl23}
Dongfu Jiang, Xiang Ren, and Bill~Yuchen Lin.
\newblock Llm-blender: Ensembling large language models with pairwise ranking
  and generative fusion.
\newblock {\em arXiv preprint arXiv:2306.02561}, 2023.

\bibitem[JT19]{jt19}
Ziwei Ji and Matus Telgarsky.
\newblock Polylogarithmic width suffices for gradient descent to achieve
  arbitrarily small test error with shallow relu networks.
\newblock {\em arXiv preprint arXiv:1909.12292}, 2019.

\bibitem[KGW{\etalchar{+}}23]{kgw+23}
John Kirchenbauer, Jonas Geiping, Yuxin Wen, Jonathan Katz, Ian Miers, and Tom
  Goldstein.
\newblock A watermark for large language models.
\newblock {\em arXiv preprint arXiv:2301.10226}, 2023.

\bibitem[KKL20]{kkl20}
Nikita Kitaev, {\L}ukasz Kaiser, and Anselm Levskaya.
\newblock Reformer: The efficient transformer.
\newblock {\em arXiv preprint arXiv:2001.04451}, 2020.

\bibitem[LL18]{ll18}
Yuanzhi Li and Yingyu Liang.
\newblock Learning overparameterized neural networks via stochastic gradient
  descent on structured data.
\newblock {\em Advances in neural information processing systems}, 31, 2018.

\bibitem[LLH{\etalchar{+}}23]{llh+23}
Hong Liu, Zhiyuan Li, David Hall, Percy Liang, and Tengyu Ma.
\newblock Sophia: A scalable stochastic second-order optimizer for language
  model pre-training.
\newblock {\em arXiv preprint arXiv:2305.14342}, 2023.

\bibitem[LSS{\etalchar{+}}20]{lss+20}
Jason~D Lee, Ruoqi Shen, Zhao Song, Mengdi Wang, et~al.
\newblock Generalized leverage score sampling for neural networks.
\newblock {\em Advances in Neural Information Processing Systems},
  33:10775--10787, 2020.

\bibitem[LSX{\etalchar{+}}23]{lsxyz23}
Shuai Li, Zhao Song, Yu~Xia, Tong Yu, and Tianyi Zhou.
\newblock The closeness of in-context learning and weight shifting for softmax
  regression, 2023.

\bibitem[LSZ23]{lsz23}
Zhihang Li, Zhao Song, and Tianyi Zhou.
\newblock Solving regularized exp, cosh and sinh regression problems.
\newblock {\em arXiv preprint arXiv:2303.15725}, 2023.

\bibitem[LWD{\etalchar{+}}23]{lwd+23}
Zichang Liu, Jue Wang, Tri Dao, Tianyi Zhou, Binhang Yuan, Zhao Song, Anshumali
  Shrivastava, Ce~Zhang, Yuandong Tian, Christopher Re, et~al.
\newblock Deja vu: Contextual sparsity for efficient llms at inference time.
\newblock In {\em International Conference on Machine Learning}, pages
  22137--22176. PMLR, 2023.

\bibitem[MGN{\etalchar{+}}23]{mgn+23}
Sadhika Malladi, Tianyu Gao, Eshaan Nichani, Alex Damian, Jason~D Lee, Danqi
  Chen, and Sanjeev Arora.
\newblock Fine-tuning language models with just forward passes.
\newblock {\em arXiv preprint arXiv:2305.17333}, 2023.

\bibitem[MOSW22]{mosw22}
Alexander Munteanu, Simon Omlor, Zhao Song, and David Woodruff.
\newblock Bounding the width of neural networks via coupled initialization a
  worst case analysis.
\newblock In {\em International Conference on Machine Learning}, pages
  16083--16122. PMLR, 2022.

\bibitem[NAB{\etalchar{+}}22]{nab+22}
Lorenzo Noci, Sotiris Anagnostidis, Luca Biggio, Antonio Orvieto, Sidak~Pal
  Singh, and Aurelien Lucchi.
\newblock Signal propagation in transformers: Theoretical perspectives and the
  role of rank collapse.
\newblock {\em Advances in Neural Information Processing Systems},
  35:27198--27211, 2022.

\bibitem[OS20]{os20}
Samet Oymak and Mahdi Soltanolkotabi.
\newblock Toward moderate overparameterization: Global convergence guarantees
  for training shallow neural networks.
\newblock {\em IEEE Journal on Selected Areas in Information Theory},
  1(1):84--105, 2020.

\bibitem[PMXA23]{pmx+23}
Abhishek Panigrahi, Sadhika Malladi, Mengzhou Xia, and Sanjeev Arora.
\newblock Trainable transformer in transformer.
\newblock {\em arXiv preprint arXiv:2307.01189}, 2023.

\bibitem[QSY23]{qsy23}
Lianke Qin, Zhao Song, and Yuanyuan Yang.
\newblock Efficient sgd neural network training via sublinear activated neuron
  identification.
\newblock {\em arXiv preprint arXiv:2307.06565}, 2023.

\bibitem[RGG{\etalchar{+}}20]{rgg+20}
Andreas R{\"u}ckl{\'e}, Gregor Geigle, Max Glockner, Tilman Beck, Jonas
  Pfeiffer, Nils Reimers, and Iryna Gurevych.
\newblock Adapterdrop: On the efficiency of adapters in transformers.
\newblock {\em arXiv preprint arXiv:2010.11918}, 2020.

\bibitem[RSM{\etalchar{+}}23]{rsm+23}
Rafael Rafailov, Archit Sharma, Eric Mitchell, Stefano Ermon, Christopher~D
  Manning, and Chelsea Finn.
\newblock Direct preference optimization: Your language model is secretly a
  reward model.
\newblock {\em arXiv preprint arXiv:2305.18290}, 2023.

\bibitem[Sag18]{sag18}
Matthew Sag.
\newblock The new legal landscape for text mining and machine learning.
\newblock {\em J. Copyright Soc'y USA}, 66:291, 2018.

\bibitem[SDFS20]{sdf+20}
Alexey Svyatkovskiy, Shao~Kun Deng, Shengyu Fu, and Neel Sundaresan.
\newblock Intellicode compose: Code generation using transformer.
\newblock In {\em Proceedings of the 28th ACM Joint Meeting on European
  Software Engineering Conference and Symposium on the Foundations of Software
  Engineering}, pages 1433--1443, 2020.

\bibitem[SHT23]{sht23}
Clayton Sanford, Daniel Hsu, and Matus Telgarsky.
\newblock Representational strengths and limitations of transformers.
\newblock {\em arXiv preprint arXiv:2306.02896}, 2023.

\bibitem[SY19]{sy19}
Zhao Song and Xin Yang.
\newblock Quadratic suffices for over-parametrization via matrix chernoff
  bound.
\newblock {\em arXiv preprint arXiv:1906.03593}, 2019.

\bibitem[SZS{\etalchar{+}}23]{szs+23}
Albert~Yu Sun, Eliott Zemour, Arushi Saxena, Udith Vaidyanathan, Eric Lin,
  Christian Lau, and Vaikkunth Mugunthan.
\newblock Does fine-tuning gpt-3 with the openai api leak
  personally-identifiable information?
\newblock {\em arXiv preprint arXiv:2307.16382}, 2023.

\bibitem[SZZ21]{szz21}
Zhao Song, Lichen Zhang, and Ruizhe Zhang.
\newblock Training multi-layer over-parametrized neural network in subquadratic
  time.
\newblock {\em arXiv preprint arXiv:2112.07628}, 2021.

\bibitem[TBM{\etalchar{+}}21]{tbm+20}
Yi~Tay, Dara Bahri, Donald Metzler, Da-Cheng Juan, Zhe Zhao, and Che Zheng.
\newblock Synthesizer: Rethinking self-attention for transformer models.
\newblock In {\em International conference on machine learning}, pages
  10183--10192. PMLR, 2021.

\bibitem[TDA{\etalchar{+}}20]{tda+20}
Yi~Tay, Mostafa Dehghani, Samira Abnar, Yikang Shen, Dara Bahri, Philip Pham,
  Jinfeng Rao, Liu Yang, Sebastian Ruder, and Donald Metzler.
\newblock Long range arena: A benchmark for efficient transformers.
\newblock {\em arXiv preprint arXiv:2011.04006}, 2020.

\bibitem[TLI{\etalchar{+}}23]{tli+23}
Hugo Touvron, Thibaut Lavril, Gautier Izacard, Xavier Martinet, Marie-Anne
  Lachaux, Timoth{\'e}e Lacroix, Baptiste Rozi{\`e}re, Naman Goyal, Eric
  Hambro, Faisal Azhar, et~al.
\newblock Llama: Open and efficient foundation language models.
\newblock {\em arXiv preprint arXiv:2302.13971}, 2023.

\bibitem[TMS{\etalchar{+}}23]{tms+23}
Hugo Touvron, Louis Martin, Kevin Stone, Peter Albert, Amjad Almahairi, Yasmine
  Babaei, Nikolay Bashlykov, Soumya Batra, Prajjwal Bhargava, Shruti Bhosale,
  et~al.
\newblock Llama 2: Open foundation and fine-tuned chat models.
\newblock {\em arXiv preprint arXiv:2307.09288}, 2023.

\bibitem[VKB23]{vkb23}
Nikhil Vyas, Sham Kakade, and Boaz Barak.
\newblock Provable copyright protection for generative models.
\newblock {\em arXiv preprint arXiv:2302.10870}, 2023.

\bibitem[VSP{\etalchar{+}}17]{vsp+17}
Ashish Vaswani, Noam Shazeer, Niki Parmar, Jakob Uszkoreit, Llion Jones,
  Aidan~N Gomez, {\L}ukasz Kaiser, and Illia Polosukhin.
\newblock Attention is all you need.
\newblock {\em Advances in neural information processing systems}, 30, 2017.

\bibitem[WFF{\etalchar{+}}23]{wff+23}
Junde Wu, Rao Fu, Huihui Fang, Yu~Zhang, and Yanwu Xu.
\newblock Medsegdiff-v2: Diffusion based medical image segmentation with
  transformer.
\newblock {\em arXiv preprint arXiv:2301.11798}, 2023.

\bibitem[WSC{\etalchar{+}}16]{wsc+16}
Yonghui Wu, Mike Schuster, Zhifeng Chen, Quoc~V Le, Mohammad Norouzi, Wolfgang
  Macherey, Maxim Krikun, Yuan Cao, Qin Gao, Klaus Macherey, et~al.
\newblock Google's neural machine translation system: Bridging the gap between
  human and machine translation.
\newblock {\em arXiv preprint arXiv:1609.08144}, 2016.

\bibitem[WWS{\etalchar{+}}22]{wws+22}
Jason Wei, Xuezhi Wang, Dale Schuurmans, Maarten Bosma, Fei Xia, Ed~Chi, Quoc~V
  Le, Denny Zhou, et~al.
\newblock Chain-of-thought prompting elicits reasoning in large language
  models.
\newblock {\em Advances in Neural Information Processing Systems},
  35:24824--24837, 2022.

\bibitem[WYW{\etalchar{+}}23]{wyw+23}
Junda Wu, Tong Yu, Rui Wang, Zhao Song, Ruiyi Zhang, Handong Zhao, Chaochao Lu,
  Shuai Li, and Ricardo Henao.
\newblock Infoprompt: Information-theoretic soft prompt tuning for natural
  language understanding.
\newblock {\em arXiv preprint arXiv:2306.04933}, 2023.

\bibitem[XZA{\etalchar{+}}23]{xza+23}
Zheng Xu, Yanxiang Zhang, Galen Andrew, Christopher~A Choquette-Choo, Peter
  Kairouz, H~Brendan McMahan, Jesse Rosenstock, and Yuanbo Zhang.
\newblock Federated learning of gboard language models with differential
  privacy.
\newblock {\em arXiv preprint arXiv:2305.18465}, 2023.

\bibitem[ZG19]{zg19}
Difan Zou and Quanquan Gu.
\newblock An improved analysis of training over-parameterized deep neural
  networks.
\newblock {\em Advances in neural information processing systems}, 32, 2019.

\bibitem[Zha22]{zha22}
Lichen Zhang.
\newblock {\em Speeding up optimizations via data structures: Faster search,
  sample and maintenance}.
\newblock PhD thesis, Master’s thesis, Carnegie Mellon University, 2022.

\bibitem[ZHL{\etalchar{+}}23]{zhl+23}
Eric Zelikman, Qian Huang, Percy Liang, Nick Haber, and Noah~D Goodman.
\newblock Just one byte (per gradient): A note on low-bandwidth decentralized
  language model finetuning using shared randomness.
\newblock {\em arXiv preprint arXiv:2306.10015}, 2023.

\bibitem[ZKV{\etalchar{+}}20]{zks+20}
Jingzhao Zhang, Sai~Praneeth Karimireddy, Andreas Veit, Seungyeon Kim, Sashank
  Reddi, Sanjiv Kumar, and Suvrit Sra.
\newblock Why are adaptive methods good for attention models?
\newblock {\em Advances in Neural Information Processing Systems},
  33:15383--15393, 2020.

\bibitem[ZMG19]{zmg19}
Guodong Zhang, James Martens, and Roger~B Grosse.
\newblock Fast convergence of natural gradient descent for over-parameterized
  neural networks.
\newblock {\em Advances in Neural Information Processing Systems}, 32, 2019.

\bibitem[ZPD{\etalchar{+}}20]{zpd+20}
Yi~Zhang, Orestis Plevrakis, Simon~S Du, Xingguo Li, Zhao Song, and Sanjeev
  Arora.
\newblock Over-parameterized adversarial training: An analysis overcoming the
  curse of dimensionality.
\newblock {\em Advances in Neural Information Processing Systems}, 33:679--688,
  2020.

\bibitem[ZPGA23]{zpg+23}
Haoyu Zhao, Abhishek Panigrahi, Rong Ge, and Sanjeev Arora.
\newblock Do transformers parse while predicting the masked word?
\newblock {\em arXiv preprint arXiv:2303.08117}, 2023.

\bibitem[ZRG{\etalchar{+}}22]{zrg+22}
Susan Zhang, Stephen Roller, Naman Goyal, Mikel Artetxe, Moya Chen, Shuohui
  Chen, Christopher Dewan, Mona Diab, Xian Li, Xi~Victoria Lin, et~al.
\newblock Opt: Open pre-trained transformer language models.
\newblock {\em arXiv preprint arXiv:2205.01068}, 2022.

\end{thebibliography}

\else

\fi



\end{document}